\documentclass[mnsc]{informs3}
\OneAndAHalfSpacedXI
\usepackage[utf8]{inputenc}
\usepackage{amsmath, amssymb, outlines, shortcuts, mathtools, booktabs, subfigure, caption,multirow,multicol,comment}
\usepackage[sort]{natbib}
 \bibpunct[, ]{(}{)}{,}{a}{}{,}%
 \def\bibfont{\small}%
 \def\bibsep{\smallskipamount}%
\usepackage{xcolor,colortbl}
\usepackage{nicefrac}
\usepackage{url, graphicx}
\usepackage[normalem]{ulem}
\usepackage{enumitem}
\newcommand{\norm}[1]{\left\lVert#1\right\rVert}
\newcommand{\pr}{\mathbb{P}}
\newcommand{\hpr}{\hat{\mathbb{P}}}
\newcommand{\expect}{\mathbb{E}}
\newcommand{\expectn}{\hat{\mathbb{E}}}
\newcommand{\expectnp}{\hat{\mathbb{E}}_{p}}
\newcommand{\expectna}{\hat{\mathbb{E}}_{a}}
\newcommand{\expectnk}{\hat{\mathbb{E}}_{k}}
\newcommand{\expectnpk}{\hat{\mathbb{E}}_{k, \op{pri}}}
\newcommand{\expectnak}{\hat{\mathbb{E}}_{k, \op{aux}}}

\usepackage{mathtools}
\usepackage{hyperref}
\usepackage[capitalise]{cleveref}
\Crefname{assumption}{Assumption}{Assumptions}
\Crefname{appsec}{Appendix}{Appendices}
\usepackage[Algorithm,ruled]{algorithm}
\usepackage[noend]{algpseudocode} 
\usepackage{amssymb}
\newcommand{\heta}{\hat{\eta}}

\newcommand{\ind}{\mathbb{I}}

\newcommand{\indhy}{\ind(\hY = \hy)}
\newcommand{\indy}{\ind(Y = y)}

\newcommand{\mw}{\mathcal{W}}
\newcommand{\tmw}{\tilde{\mathcal{W}}}

\newcommand{\tw}{\tilde{w}}

\newcommand{\wsuhyz}{w^*_{\alpha}(\hy, z)}

\newcommand{\wsuhyyz}{\tilde{w}^*_{\alpha}(\hy, y, z)}

\newcommand{\wuhyz}{w_{\alpha}(\hy, z)}
\newcommand{\bwuhyz}{w_{\alpha}(\hY, Z)}
\newcommand{\btwuhyyz}{\tw_{\alpha}(\hY, Y, Z)}
\newcommand{\wuhyyz}{\tilde{w}_{\alpha}(\hy, y, z)}
\newcommand{\bwuhyyz}{\tilde{w}_{\alpha}(\hY, Y, Z)}

\newcommand{\hY}{\hat{Y}}
\newcommand{\hy}{\hat{y}}

\newcommand{\puz}{\pr(A=\alpha \mid Z=z)}
\newcommand{\phyz}{\pr(\hY=\hy \mid Z=z)}
\newcommand{\phyyz}{\pr(\hY=\hy, Y=y \mid Z=z)}

\newcommand{\arange}{\alpha \in \mathcal{A}}
\newcommand{\yrange}{y \in \{0, 1\}}
\newcommand{\hyrange}{\hy \in \{0, 1\}}
\newcommand{\yhyrange}{\hy, y \in \{0, 1\}}
\newcommand{\zrange}{z \in \mathcal{Z}}

\newcommand{\DD}{{\op{DD}}}
\newcommand{\TPRD}{{\op{TPRD}}}
\newcommand{\TNRD}{{\op{TNRD}}}
\newcommand{\PPVD}{{\op{PPVD}}}
\newcommand{\NPVD}{{\op{NPVD}}}
\newcommand{\LTP}{{\op{LTP}}}
\newcommand{\Lip}{{\op{Lip}}}
\newcommand{\FH}{\op{FH}}

\def\edit{}
\def\bedit{}

\newcommand{\rangespace}{R}

\newcommand{\naux}{n_{\op{aux}}}
\newcommand{\npri}{n_{\op{pri}}}

\newcommand{\na}{\textrm{aux}}
\newcommand{\nm}{\textrm{pri}}

\newcommand{\pri}{\op{pri}}
\newcommand{\aux}{\op{aux}}

\newcommand{\etaa}{\eta_{\op{aux}}}
\newcommand{\etap}{\eta_{\op{pri}}}
\newcommand{\etapt}{\tilde\eta_{\op{pri}}}
\newcommand{\etak}{\hat{\eta}^{-k}}
\newcommand{\etaak}{\hat{\eta}^{-k}_{\op{aux}}}
\newcommand{\etapk}{\hat{\eta}^{-k}_{\op{pri}}}
\newcommand{\etaptk}{\hat{\tilde{\eta}}^{-k}_{\op{pri}}}
\newcommand{\etaah}{\hat{\eta}_{\op{aux}}}
\newcommand{\etaph}{\hat{\eta}_{\op{pri}}}

\newcommand{\hetat}{\hat{\tilde{\eta}}}

\newcommand{\hetaaz}{\hat\eta_{\aux}(\alpha, z)}
\newcommand{\hetahyz}{\hat\eta_{\pri}(\hy, z)}
\newcommand{\hetahyyz}{\hat{\tilde{\eta}}_{\pri}(\hy,y, z)}

\newcommand{\hE}{\hat{\mathbb{E}}}
\newcommand*{\defeq}{\mathrel{\vcenter{\baselineskip0.5ex \lineskiplimit0pt
			\hbox{\scriptsize.}\hbox{\scriptsize.}}}%
	=}
\newcounter{relctr} %
\everydisplay\expandafter{\the\everydisplay\setcounter{relctr}{0}} %

\newcommand\labelrel[2]{%
	\begingroup
	\refstepcounter{relctr}%
	\stackrel{\textnormal{(\alph{relctr})}}{\mathstrut{#1}}%
	\originallabel{#2}%
	\endgroup
}
\AtBeginDocument{\let\originallabel\label} %

\TheoremsNumberedThrough     %
\ECRepeatTheorems

\EquationsNumberedThrough    %
\MANUSCRIPTNO{MS-0000-0000.00}

\let\oldendproof\endproof
\renewcommand{\endproof}{\halmos\oldendproof}

\begin{document}

\RUNTITLE{Fairness Using Data Combination}
\TITLE{%
Assessing Algorithmic Fairness with Unobserved Protected Class Using Data Combination%
}
\RUNAUTHOR{Kallus, Mao, Zhou}
\ARTICLEAUTHORS{%
\AUTHOR{Nathan Kallus}
\AFF{Cornell University, \EMAIL{kallus@cornell.edu}}
\AUTHOR{Xiaojie Mao}
\AFF{Cornell University, \EMAIL{xm77@cornell.edu}}
\AUTHOR{Angela Zhou}
\AFF{Cornell University, \EMAIL{az434@cornell.edu}}
} %

\ABSTRACT{%
The increasing impact of algorithmic decisions on people's lives compels us to scrutinize their fairness and, in particular, the disparate impacts that ostensibly-color-blind algorithms can have on different groups. Examples include credit decisioning, hiring, advertising, criminal justice, personalized medicine, and targeted policymaking, where in some cases legislative or regulatory frameworks for fairness exist and define specific protected classes. In this paper we study a fundamental challenge to assessing disparate impacts in practice: protected class membership is often not observed in the data. This is particularly a problem in lending and healthcare. We consider the use of an auxiliary dataset, such as the US census, \edit{to construct models that predict the protected class from proxy variables, such as surname and geolocation.}
We show that \edit{even with such data}, a variety of common disparity measures are generally unidentifiable, providing a new perspective on the documented biases of popular proxy-based methods. \edit{We provide exact characterizations of the tightest-possible set of all possible true disparities that are consistent with the data (and possibly any assumptions)}. We further provide optimization-based algorithms for computing and visualizing these sets \edit{and statistical tools to assess sampling uncertainty. Together, these enable reliable and robust assessments of disparities} -- an important tool when disparity assessment can have far-reaching policy implications. We demonstrate this in two case studies with real data: mortgage lending and personalized medicine dosing.
}%

\HISTORY{First version: May 2019. This version: June 2020.}

\maketitle

\section{Introduction}

The spread of prescriptive analytics and algorithmic decision-making has given rise to urgent ethical and legal imperatives to avoid discrimination and guarantee fairness with respect to protected classes. 
In advertising, prescriptive algorithms target for maximal impact and revenue \citep{iyer2005targeting,goldfarb2011online}, but recent studies found gender-based discrimination in who receives ads for STEM careers \citep{lambrecht2019algorithmic} and other worrying disparities \citep{datta2015automated,sweeney2013discrimination}. 
In hiring, algorithms help employers efficiently screen applicants \citep{hiring}, but in some cases this can have unintended biases, e.g., against women and minorities \citep{dastin2018amazon}.
In criminal justice, algorithmic recidivism scores allow judges to assess risk \citep{monahan2016risk}, while recent studies have revealed systematic race-based disparities in error rates \citep{angwin2016machine,chouldechova2017fair}.
In healthcare, algorithms that allocate resources like care management have been shown to exhibit racial biases \citep{obermeyer2019dissecting} and personalized medicine algorithms can offer disparate benefits to different groups \citep{rajkomar2018ensuring,goodman2018machine}.
In lending, prescriptive algorithms optimize credit decisions using predicted default risks and their induced disparities are regulated by law \citep{occhandbook2010}, leading to legal cases against discriminatory lending \citep{CFPB:ally2013}.

For regulated decisions, there are two major legal theories of discrimination: 
\begin{itemize}
\item Disparate treatment \citep{Zimmer1996aa}: informally, intentionally treating an individual differently on the basis of membership in a protected class; and
\item Disparate impact \citep{Rutherglen:1987aa}: informally, adversely affecting members of one protected class more than another even if by an ostensibly neutral policy.
\end{itemize}
Thus, \edit{even} prescriptive algorithms that do not take race, gender, or other sensitive attributes as an input may \edit{often} satisfy equal treatment but may still induce disparate impact \citep{Kleinberg2017}.
Indeed, many of the disparities found above take the form of \emph{unintended disparate impact of ostensibly class-blind prescriptive algorithms}.
\edit{While our contextual discussion focuses on U.S. discrimination law and regulation, our methodology is a general one for assessing disparities with respect to protected class and may apply in many legal and regulatory contexts.\footnote{\edit{In the U.S., the Fair Housing Act (FHA) and Equal Credit Opportunity Act (ECOA) codify as protected attributes: age, race/ethnicity, disability, exercised rights under CCPA, familial status (household composition), gender identity, marital status (single or married), national origin, race, recipient of public assistance, religion, and/or sex.}}}

\edit{In consequential decision-making contexts such as hiring or lending, 
	assessing 
	disparities
	 is paramount for monitoring the potential harms of decision systems. Assessing
	 disparities induced by
	  a prescriptive algorithm involves evaluating the differences in the distributions of decision outcomes received by different groups, either marginally 
	  or conditional on some additional ground truth. 
We define precisely the disparity metrics of interest
in \cref{sec: disparity-measure} and discuss related work in \cref{sec: literature}.
While what size of disparity counts as unacceptable depends on the appropriate legal, ethical, and regulatory context, 
in any case, they must first be \emph{measured}.}

In this paper, we study a fundamental challenge to assessing the disparity induced by prescriptive algorithms in practice:
\begin{center}
{\bf \textit{protected class membership is often \underline{not} observed in the data}.}
\end{center}
There may be many reasons for this missingness in practice, both legal, operational, and behavioral.
In the US financial service industry, lenders are \emph{not} permitted to collect race and ethnicity information on applicants for non-mortgage products%
\footnote{The US Home Mortgage Disclosure Act (HMDA) authorizes lenders to collect such information for mortgage applicants and co-applicants.}
such as credit cards, auto loans, and student loans.
This considerably hinders auditing \edit{fairness} for non-mortgage loans, both by internal compliance officers and by regulators \citep{zhang2016assessing}.
Similarly, health plans and health care delivery entities lack race and ethnicity data on most of their enrollees and patients, as a consequence of high data-collection costs and people's reluctance to reveal their race information for fear of potential discrimination \citep{weissman2011advancing}. This data collection challenge makes monitoring of racial and ethnic differences in care impractical and impedes the progress of healthcare equity reforms \citep{gaffney2017affordable}.

To address this challenge, some methods heuristically use observed proxies to predict and impute unobserved protected class labels. The most (in)famous example is the Bayesian Improved Surname Geocoding (BISG) method.
BISG estimates conditional race membership probabilities given surname and geolocation (e.g., census tract, ZIP code, or county) using data from the US decennial census, and then imputes the race labels based on the estimated probabilities.
Since its invention \citep{elliott2008new, elliott2009using}, the BISG method has been widely used in assessing racial disparities in health care \citep[e.g., ][]{fremont2005use, nerenz2009race,weissman2011advancing, brown2016using},
as well in the US financial industry, where the Consumer Financial Protection Bureau (CFPB)
used BISG to support analysis leading to a \$98-million settlement against Ally Bank for harming minority borrowers for auto loans \citep{CFPBproxy2014,CFPB:ally2013}.

The validity of using proxies for the unobserved protected class for disparity assessment remains controversial, 
and relevant research is still limited. 
\edit{Although advanced proxy methods like BISG outperform previous proxy methods,
further research shows that it leads to biased disparity assessment \citep{Baines, zhang2016assessing}. In particular, \citet{chen2019fairness} analyzed the underlying mechanism for the \edit{statistical} bias of BISG's assessments due to the joint dependence among lending outcome, geolocation, and race.} However, a systematic understanding of the precise limitations of using proxy methods in disparity assessment in general, and possible remedies to the potential \edit{statistical} biases, is still lacking.\footnote{\edit{For clarity, we emphasize to the reader the difference between an algorithm's ``bias'' with respect to protected groups, e.g., as quantified by disparate impact, and the \emph{statistical bias} of assessments of such disparities. In this paper, ``bias'' only ever refers to the latter statistical bias and ``disparities'' to systematic differences in algorithmic outputs.}} 
Filling in this gap is an important and urgent need, especially given the wide use of proxy methods and the \edit{significant managerial and policy impacts} of disparity assessment in the settings where they are used, which motivates our current work. 

\edit{\textbf{Practical implications.} In this paper, we demonstrate that it is generally \emph{impossible} to identify impact disparities when only proxy information is available for protected class, and we instead study how to precisely and reliably characterize the \emph{range} of all possible disparities that are consistent with all available data, known as the \emph{partial identification set}. Since disparities are unidentifiable, any single point estimate thereof is fundamentally spurious, and any conclusion drawn from it is vulnerable to criticism. This is a grave and real concern in lending, healthcare, and other applications where disparate impact assessments can have far-reaching policy implications. 
In contrast, by conducting inference on the partial identification set from data, our proposed methods can support credible, principled conclusions about disparities. In particular, these quantify the fundamental ambiguity in disparities and the value of more informative proxies or assumptions, especially if our partial identification sets are large; or, if the sets are small, they provide a statistical test certifying the presence of disparities \textit{independent} of further untestable assumptions.}

\subsection{Contribution\edit{s}}
In this paper, we study the basic statistical identification limits for assessing disparities when protected class labels are unobserved and provide new optimization-based algorithms for computing the partial-identification bounds on said disparities, which can enable robust and reliable auditing of the disparate impact of prescriptive algorithms.

We highlight our primary contributions below: 
\begin{description}
\item[Problem formulation.] 
\edit{
	To facilitate a principled analysis of (partial) identifiability, we formulate disparity assessment with proxies as a \emph{data combination problem} with two datasets: 
}
\begin{description}
\item[-- a \textit{primary dataset}] with the decision outcomes, (potentially) true outcomes, and proxy variables, but where the protected class labels are \emph{missing};
and
\item[-- an \textit{auxiliary}] dataset with proxy variables and protected class labels, but without outcomes.
\end{description}
\item[Identification Conditions.] \edit{We prove tight necessary and sufficient conditions for the unidentifiability of disparity measures in this setting. In the absence of these (unrealistically strong) conditions, disparities are \emph{necessarily} unidentifiable from the two datasets. That is, the partial identification set of all disparity measure values consistent with the data-generating processes of the two datasets is not a singleton.}
\item[Characterizing and Computing the Partial Identification Set.] 
We exactly characterize the partial identification sets of a variety of disparity measures under data combination, that is, the smallest set containing all possible values that disparity measures may simultaneously take while still agreeing with the data. \edit{Our characterization is \emph{sharp} in that it is \emph{equal} to this set rather than merely containing it.}
\edit{We provide closed-form formulations of partial identification sets for binary comparisons.
And, we provide optimization algorithms to compute partial identification sets when we incorporate additional mild smoothness assumptions that reduce ambiguity or when we consider simultaneous comparisons across more than two protected classes. In the latter case, we compute the \emph{support function} of the partial identification set.}
\item[\edit{Estimation and Inference.}] \edit{We study the additional \emph{sampling uncertainty} of our proposals when given finite observations from each dataset. Specifically, we prove consistency guarantees when one plugs in estimates of probability and conditional probability models. To enable inference, i.e., constructing confidence intervals on top of the estimated partial identification intervals, we propose a approach based on debiased machine learning that is invariant to the estimation of certain conditional probability models.}
\item[Robust Auditing.] \edit{Together,} these tools facilitate robust and reliable fairness auditing. Since the sets we describe are \textit{sharp} in that they are the tightest-possible characterization of disparity given the data,
their size generally captures the amount of \textit{ambiguity}
that remains in evaluating disparity when the protected class is unobserved and only proxies are available.
When the observed data is very informative about the disparity measures, the set tends to be small and may still lead to meaningful conclusions regarding the sign and magnitudes of disparity, despite unidentifiability. 
In contrast, when the observed data is insufficient, the set tends to be large and gives a valuable warning about the risk of drawing conclusions from the fundamentally limited observed data.
\item[Empirical Analysis.] We apply our approach in two real case studies: evaluating the racial disparities (1) in mortgage lending decisions and (2) in personalized Warfarin dosing. We demonstrate how adding extra assumptions may decrease the size of partial identification sets of disparity measures, and illustrate 
how stronger proxies -- either for race or for outcomes -- can lead to smaller partial identification sets and more informative conclusions on disparities.  
\end{description}

\section{Problem Setup}\label{sec: set-up}
\begin{figure}
\centering
\begin{tabular}{ccccc}
\multicolumn{5}{c}{Primary dataset}\\
\toprule
$Z_s$ & $Z_g$ & \multirow{2}{*}{$\cdots$} & $\hY$ & $Y$ \\
Surname & ZIP code &  & Approval & Non-default \\\midrule
Jones & 94122 & \multirow{2}{*}{$\cdots$} & Y & N \\
\vdots&\vdots&&\vdots&\vdots\\\bottomrule
\end{tabular}\hfill
\begin{tabular}{ccccc}
\multicolumn{5}{c}{Auxiliary dataset}\\
\toprule
$Z_s$ & $Z_g$ & White & \multirow{2}{*}{$\cdots$} & API \\
Surname & ZIP code & \% &  & \% \\\midrule
Jones & 94122 & 47\% & \multirow{2}{*}{$\cdots$} & 31\% \\
\vdots&\vdots&\vdots&&\vdots\\\bottomrule
\end{tabular}
\caption{Illustration of the two observed datasets for assessing lending disparity with unobserved race labels.}
\label{fig: data-comb}
\end{figure}
We mainly consider four types of relevant variables:
\begin{description}
\item[Decision outcome, $\hY \in \{0, 1\}$,] is the prescription by either human decision makers or machine learning algorithms.
    For example, $\hY = 1$ represents approval of a loan application, which is often based on some prediction of default risk.
    We call $\hY = 1$ the \textit{positive} decision, even if is not favorable in terms of utility (e.g., high medicine dosage in \Cref{sec: warfarin}). 
\item[True outcome, $Y  \in \{0, 1\}$,] is a target variable that justifies an optimal decision. \edit{$\hY$ is often based on imperfect predictions of $Y$.}
    In the lending example (Section \ref{sec: hmda}), we denote $Y = 1$ for loan applicants who would not  default on loan payment if the loan application were approved. 
    $Y$ is not known to decision makers at the time of decision making. 
\item[Protected attribute, $A \in \mathcal{A}$,] is a categorical variable (e.g., race or gender). 
    Our convention is to let $A = a$ be a group understood to be generally \textit{advantaged} and $A = b$ \textit{disadvantaged}. 
\item[Proxy variables, $Z \in \mathcal{Z}$,] are a set of additional observed covariates. In proxy methods, these are used to predict $A$. In the BISG example (Section \ref{sec: hmda}), $Z$ stands for surname and geolocation. The proxy variables can be categorical, continuous, or mixed. 
\end{description}
In this paper, we mainly focus on binary outcomes (true outcome and decision outcome), but our results can be straightforwardly extended to multi-leveled outcomes. 

We formulate the problem of using proxy methods from a \textit{data combination} perspective. Specifically, we assume we have two datasets:
the \textit{main} dataset with observations of $(\hY,Y,Z)$, and
the \textit{auxiliary} dataset with observations of $(A,Z)$.
\edit{
\Cref{fig: data-comb} is an illustration of these two datasets in the example of BISG proxy method (\Cref{sec: hmda}).
\begin{assumption}\label{asn-samedist}
The primary and auxiliary datasets both consist of i.i.d. (independent and identically distribution) draws, each from the respective marginalization of a common joint distribution.
\end{assumption}
Therefore, the information from observing these two separate datasets can be characterized by  $\pr(\hY, Y, Z)$ and $\pr(A, Z)$ respectively, each being a marginalization of a common larger joint distribution $\pr(A,\hY,Y,Z)$.\footnote{\edit{\Cref{asn-samedist} can be relaxed by assuming instead that the distribution $\mathbb P_a$ of the auxiliary observations $(A, Z)$ satisfies $\mathbb P_a(A=\alpha\mid Z)=\mathbb P(A=\alpha\mid Z)$, with an \emph{arbitrary} distribution $\mathbb P_a(Z)$ of proxy variables. This relaxation does not change any of our results in \cref{sec-PI-identifiability,sec: PI-binary,sec: compute pi}, but it does change our estimators in \cref{sec-inference}, where we would need to account for this distributional shift in $Z$ across the datasets. We omit this straightforward extension for brevity.}}}
However, we cannot simply join these two datasets directly for many possible reasons. For example, no unique identifier for individuals (e.g., social security number) exists in both datasets.
Thus we \emph{cannot} learn the combined joint distribution $\pr(A,\hY,Y,Z)$ from these two separate, unconnected datasets.

\subsection{Disparity measures}\label{sec: disparity-measure}
In this paper, we focus on assessing the disparity in the decision $\hY$ with respect to the protected attribute $A$, as well as possibly with respect to true outcome labels $Y$. 
We illustrate our method with widely-used disparity measures that are a measure of class-conditional classification error, and, if we were given observations of true class labels, they could be computed from a $2 \times 2\times\abs{\mathcal A}$ within-class confusion matrix of the decision and true outcome. 

Specifically, we consider the following disparities:
\\\emph{Demographic Disparity}: 
$
\delta_\DD(a,b)=\pr(\hY = 1 \mid A = a)-\pr(\hY = 1 \mid A = b).
$
\\\emph{True Positive Rate Disparity}: 
$\delta_\TPRD(a,b)=\pr(\hY = 1 \mid A = a, Y=1)-\pr(\hY = 1 \mid A = b, Y=1).$
\\\emph{True Negative Rate Disparity}: 
$\delta_\TNRD(a,b)=\pr(\hY = 0 \mid A = a, Y=0)-\pr(\hY = 0 \mid A = b, Y=0).$
\\\emph{Positive Predictive Value Disparity}: 
$\delta_\PPVD(a,b)=\pr(Y = 1 \mid A = a, \hY=1)-\pr(Y = 1 \mid A = b, \hY=1).$
\\\emph{Negative Predictive Value Disparity}: 
$\delta_\NPVD(a,b)=\pr(Y = 0 \mid A = a, \hY=0)-\pr(Y = 0 \mid A = b, \hY=0).$

\edit{To illustrate,} we interpret these disparity measures using the running example of making lending decisions. DD measures the disparity in within-class average loan approval rate.\footnote{Strictly speaking, demographic disparity is not based on classification ``error'' but it can be also computed from the within-class confusion matrices.}
TPRD (respectively, TNRD) measures the disparity in the proportions of people who \textit{correctly} get approved (respectively, rejected) in loan applications between two classes, given their true non-default or default outcome. 
Compared to DD, TPRD and TNRD only measure the disparity that is unmediated by existing base disparities in true outcome $Y$ and is considered more relevant for classification settings
when concerned with disparities in allocation of a positive outcome in view of qualifying characteristics such as creditworthiness \citep{hardt2016equality}.
\edit{Such disparities can be interpreted as ``disparate opportunity'' to equally-qualified individuals from different groups.}
PPVD (respectively, NPVD) measures the disparity in the proportions of approved applicants who pay back their loan (respectively, rejected applicants who default) between two classes. Such disparities can be interpreted as ``disparate benefit of the doubt'' in an individual having the positive label.

We will present our results in terms of DD, TPRD, and TNRD. Indeed, by swapping the roles of $Y$ and $\hY$ in TPRD and TNRD, all our results can straightforwardly be extended to PPVD and NPVD, respectively.
Similarly, disparities based on false negative rate and false positive rate simply differ with TPRD and TNRD by a minus sign, i.e., are given by swapping $a$ and $b$. 
To streamline the presentation, we typically use $\alpha$, $z$, $\hy$, $y$ as generic values of the random variables $A$, $Z$, $\hY$, $Y$, respectively. We also use $a$ and $b$ as additional generic values for $A$, where $a$ is generally understood to be a majority or advantaged class label. We further define the outcome probabilities for protected class $\alpha$ as $\mu(\alpha)\coloneqq \pr(\hY=1\mid A=\alpha)$ and $\mu_{\hy y}(\alpha) \coloneqq \pr(\hY=\hy\mid A=\alpha,Y=y)$, so that $\delta_\DD(a,b)=\mu(a)-\mu(b)$, $\delta_\TPRD(a,b)=\mu_{11}(a)-\mu_{11}(b)$, and $\delta_\TNRD(a,b)=\mu_{00}(a)-\mu_{00}(b)$. Throughout this paper, we use $\expect$ to denote expectation with respect to the target distribution $\pr$.

\section{Related Literature}\label{sec: literature}

\paragraph{Proxy methods.} 
The validity of proxy methods for disparity assessment depends not only on the statistical estimation of, for example, $\pr(A = \alpha \mid Z=z)$,
but also the specific procedure with which this is combined with other information.
While BISG has been shown to outperform previous proxies (surname-only and geolocation-only analysis), these evaluations \citep{CFPBproxy2014,imai2016improving,dembosky2019indirect}
focus on classification accuracy, which is never perfect, and do not consider impact on downstream disparity assessment, mostly because this is usually unknowable. 
In contrast, \citet{Baines,zhang2016assessing} assessed disparity on a mortgage dataset, and found that using imputed race tends to overestimate the true disparity.
\citet{chen2019fairness} provided a full analysis of this bias and \edit{developed} sufficient conditions to determine its direction 
and found that disparity estimation methods using imputed race are very sensitive to arbitrary tuning parameters such as imputation threshold.
As we show in \cref{sec-PI-identifiability}, disparity is generally \textit{unidentifiable} from proxies when protected class is unobserved;
consequently, all previous point estimators are generally biased
unless very strong assumptions are satisfied. 

\paragraph{Algorithmic Fairness.} 
In this paper, we consider auditing two measures of fairness that have received considerable attention in the fair machine learning community: \textit{demographic (dis)parity} and \textit{classification (dis)parity}, which we outlined in \Cref{sec: disparity-measure}.
\edit{
Many other ``fairness metrics'' have been proposed to facilitate risk assessment for algorithmic decision making in different contexts \citep{narayanan2018translation,verma2018fairness}; for more comprehensive discussion, we refer to \cite{barocas-hardt-narayanan}. We emphasize that we focus on auditing, not adjusting, disparity measures. 
Whether observed disparities warrant adjustments depends on the legal, ethical, and regulatory context.\footnote{{For example,
	as fairness criteria, both demographic and classification parity have been criticized for their \emph{inframarginality}, i.e., they average 
	over individual risk far from the decision boundary 
	\citep{corbett2018measure}.
	However, inframarginality may be unavoidable when outcomes are binary. There may be no true individual ``risk,'' only the stratified frequencies of binary outcomes (default or recidivation) over strata defined by predictive features, which are in turn chosen by the decision maker.}}
} 
\paragraph{Partial Identification and Data Combination.}
There is an extensive literature on partial identification of unidentifiable parameters \citep[e.g.,][]{manski2003partial,beresteanu2011sharp}. There are many reasons parameters may be unidentifiable, including confounding \citep[e.g.,][]{kallus2018interval,kallus2018confounding}, missingness \citep[e.g.,][]{manski2005partial}, and multiple equilibria \citep[e.g.,][]{ciliberto2009market}. One prominent example is data combination, also termed the ``ecological inference problem,'' where joint distributions must be reconstructed from observation of marginal distributions \citep{schuessler1999ecological,jiang2018ecological,freedman1999ecological,wakefield2004ecological}. One key tool for studying this problem is the Fr\'echet-Hoeffding inequalities, which give sharp bounds on joint cumulative distributions and super-additive expectations given marginals \citep{cambanis1976inequalities,ridder2007econometrics,fan2014identifying}.
Such tools are also used in risk analysis in finance to assess risk without knowledge of copulas \citep{ruschendorf2013mathematical}. In contrast to much of the above work, we focus on assessing \emph{nonlinear} functionals of partially identified distributions, namely, true positive and negative rates, as well as on leveraging conditional information to integrate marginal information across proxy-value levels with possible smoothness constraints.

\section{Unidentifiability of Disparity Measures Under Data Combination}\label{sec-PI-identifiability}
In this section we study the fundamental limits of the two separate datasets to identify -- i.e., pinpoint -- the disparity measures of interest. 
We first introduce the concept of \textit{identification} \citep{lewbel2016identification}. 
\edit{We call a quantity of interest (either finite-dimensional or infinite-dimensional) \textit{identifiable} if it can be uniquely determined by (i.e., is a function of) the probability distribution function of the data. 
Conversely, it is \textit{unidentifiable} if multiple different values of this quantity all simultaneously agree with the distribution of observed data. This is motivated by the fact that, in the i.i.d. setting, the distribution of the data (equivalently, the distribution of any single data point) is the most we can hope to learn from any amount of observations, even infinitely many.}

\edit{
	The disparity measures of interest in \cref{sec: set-up} are all functions of the full joint distribution $\pr(A, \hY, Y, Z)$ and are clearly identifiable \textit{if} we observed the full data $(A, \hY, Y, Z)$.  
	We will show in \cref{sec: unidentifiability} that disparity measures are generally unidentifiable from two separate datasets, since the corresponding marginal distributions
	$\pr(\hY, Y, Z)$ and $\pr(A, Z)$ are
insufficient to uniquely determine the full joint distribution and, in particular, the disparity measures. 
}

\edit{
Analyzing the identifiability of disparity measures is crucial since
unidentifiability 
implies that
learning disparities based on the observed data alone is fundamentally ambiguous: it is \textit{impossible}, even with an infinite amount of observed data, to pin down the exact values of the disparity measures.
Consequently
 any point estimate is in some sense \emph{spurious}: biased and potentially sensitive to ad-hoc modeling specifications \citep{d2019multi}. In this case, generally one must be very cautious about drawing any substantive conclusions based on point estimates of disparity measures. 
}

\edit{Since identifiability and partial identification sets are properties of \emph{distributions} (i.e., are \emph{population} quantities), we focus for the time being on the consequences of fully knowing the marginals $\pr(\hY, Y, Z)$ and $\pr(A, Z)$, which can be learned from the two datasets given sufficient data. This captures the \emph{identification uncertainty} involved in disparity assessments using data combination. We revisit the assumption of full knowledge of marginals in \cref{sec-inference}, where we discuss how the partial identification sets can be \emph{estimated} from the data itself and how to construct \emph{confidence intervals} on these. This captures the \emph{sampling uncertainty} involved in only having finite datasets.}

\subsection{Unidentifiability of Disparities}\label{sec: unidentifiability}
Since the disparity measures are functions of the full joint distribution $\pr(A, \hY, Y, Z)$, to prove the unidentifiability of the disparity measures, we show that there generally exist multiple \textit{valid} full joint distributions that give rise to different disparities but at the same time agree with the marginal joint distributions $\pr(\hY, Y, Z)$ and $\pr(A, Z)$, which characterize the primary dataset and the auxiliary dataset, respectively. To formalize the validity of full joint distributions, we introduce the \textit{coupling} of two marginal distributions \citep{villani2008optimal}. Because outcomes and protected classes are discrete, we focus on couplings of discrete distributions.\footnote{Proxies can still be continuous, which we will leverage when we impose extra smoothness assumptions in \Cref{sec: PIsmooth}.}

\begin{definition}[Coupling Sets]\label{def: coupling}
	Given two discrete probability spaces $(\mathcal{S}, \sigma)$ and $(\mathcal{T}, \tau)$ (i.e., $\sigma(s)\geq0,\tau(t)\geq0,\sum_{s\in\mathcal S}\sigma(s)=1,\sum_{t\in\mathcal T}\tau(t)=1$), a distribution $\pi$ over $\mathcal S\times\mathcal T$ is a \emph{coupling} of $(\sigma, \tau)$ if the marginal distributions of $\pi$ coincide with $\sigma$, $\tau$. The set of all possible couplings is denoted by
	\begin{align}\label{eq: coupling}
		\Pi(\sigma, \tau) = \bigg\{\pi\in\R{\mathcal S\times\mathcal T}\ :\ 
						\sum_{t \in \mathcal{T}}\pi(s, t) = \sigma(s), 
						~ \sum_{s \in \mathcal{S}}\pi(s, t) = \tau(t), 
						~ 0 \le \pi(s, t) \le 1, 
						~ \forall s \in \mathcal{S}, t \in \mathcal{T}
						\bigg\}.
	\end{align}
\end{definition}
\edit{
	\cref{def: coupling} gives the set of all possible valid joint distributions that agree with given marginals. 
	It	states that any joint distribution is valid as long as it satisfies the \textit{Law of Total Probability} with respect to the fixed marginals.  
The classical Fre\'chet-Hoeffding inequality 
	provides bounds on the possible values of these joint distributions with knowledge of the fixed marginals \citep{cambanis1976inequalities, ridder2007econometrics, fan2014identifying}:
		this characterization informs our discussion of the size of partial identification sets in \Cref{sec: PI-size} and our 
		derivation of closed-form partial identification sets in \Cref{sec: PI-binary}.
}
\begin{proposition}[Fr\'echet-Hoeffding] \label{prop: FH-coupling}
	The coupling set is equivalently given by
		\begin{align}
			\Pi(\sigma, \tau) = 
			\bigg\{\pi\in\R{\mathcal S\times\mathcal T}\ :\  
			&\textstyle\sum_{t \in \mathcal{T}}\pi(s, t) = \sigma(s), 
                        ~ \sum_{s \in \mathcal{S}}\pi(s, t) = \tau(t), 
                        \nonumber \\ \label{eq: FH-coupling}
			~  &\max\{\sigma(s) + \tau(t) - 1, 0 \} \le \pi(s, t) \le \min\{\sigma(s), \tau(t)\}, 
			~ \forall s \in \mathcal{S}, t \in \mathcal{T}
			\bigg\}.
	\end{align}
\end{proposition}
\begin{figure}
\centering\setlength{\tabcolsep}{1.5em}%
\def\arraystretch{1.6}%
\definecolor{Gray}{gray}{0.85}%
\definecolor{white}{gray}{1}%
\begin{tabular}{c|>{\columncolor{Gray}}c>{\columncolor{Gray}}c|c}
\rowcolor{white}
        &   $A=a$                   &   $A=b$   &\\\hline
$\hY=0$ &$\pr(A=a,\hY=0\mid Z=z)$  & $\pr(A=b,\hY=0\mid Z=z)$ & $\pr(\hY=0\mid Z=z)$ \\
$\hY=1$ &$\pr(A=a,\hY=1\mid Z=z)$  & $\pr(A=b,\hY=1\mid Z=z)$ & $\pr(\hY=1\mid Z=z)$ \\\hline
        \rowcolor{white}& $\pr(A=a\mid Z=z)$ & $\pr(A=b\mid Z=z)$ & 1
\end{tabular}
\caption{Unidentifiability of joint distributions given marginal distributions. The gray region denotes unknown joint probabilities. Row and column sums are known. Even with  binary protected class and outcome, this leaves one degree of freedom in the unknowns, unless one of the marginals is degenerate.}
\label{fig:ecological-inference}
\end{figure}
\edit{
We let $\mathcal{P}_D$ denote the set of all valid full joint distributions of $(\hY, Y, A, Z)$ that agree with the marginal distributions $\pr(\hY, Y, Z)$ and $\pr(A, Z)$, as characterized by this definition of couplings:
\begin{align}\label{eq: P-D}
\mathcal{P}_D = \left\{\pr': \pr'(Z) = \pr(Z), ~~ \pr'(\hY, Y, A \mid Z) \in \Pi\left(\pr(\hY, Y \mid Z), \pr(A \mid Z)\right)\right\} 
\end{align}
The set $\mathcal{P}_D$ generally contains multiple elements,
since the joint dependence structure can be arbitrary so long as the marginals are compatible 
(characterized by either \cref{eq: coupling} or \cref{eq: FH-coupling}).
We illustrate this for $\Pi(\pr\fprns{A \mid Z}, \pr\fprns{\hY \mid Z})$ in \cref{fig:ecological-inference}.
With binary protected class and outcomes, marginal information provides only three independent constraints on four unknowns, so that the joint distribution cannot be uniquely determined. 
This also extends to $\Pi(\pr\fprns{A \mid Z}, \pr\fprns{\hY, Y \mid Z})$.}

\edit{%
We next show that in addition to the full joint distribution, the disparities, which are differences of nonlinear functionals of the full joint distribution, in particular cannot be uniquely identified. 
\begin{proposition}\label{prop: minimal-unid}
Let $\mathcal A=\{a,b\}$. Let any marginal distributions $\pr(\hY, Y, Z)$, $\pr(A, Z)$ be given. \\
(i) If there exists a set of $z$'s with positive probability such that $0 < \pr(\hY=\hy \mid Z=z) < 1$ and $0 < \pr(A=\alpha \mid Z=z) < 1$ for $\hyrange$ and $\arange$, then $\delta_\DD(a,b)$ is unidentifiable without further conditions.
That is, there exist two different joint distributions of $(A, \hY, Y, Z)$ that agree with these marginals but give rise to different values of $\delta_\DD(a,b)$. 
\\
(ii) If there exists a set of $z$'s with positive probability such that $0 < \pr(\hY=\hy,Y=y \mid Z=z) < 1$ and $0 < \pr(A=\alpha \mid Z=z) < 1$ for $\yhyrange$ and $\arange$, then both $\delta_\TPRD(a,b)$ and $\delta_\TNRD(a,b)$ are unidentifiable without further conditions.  
\end{proposition}
\Cref{prop: minimal-unid} shows that as long as the proxies $Z$ cannot perfectly predict the protected class $A$ or outcomes $\hY, Y$, 
then DD, TPRD, TNRD are unidentifiable from the observed data information alone. This holds for \emph{any} given pair of marginal distributions. We can also prove the same conclusion for PPVD, NPVD by exchanging $\hY$ and $Y$.
To prove \cref{prop: minimal-unid} we show that we can always construct different feasible couplings of the given marginals, i.e., different feasible elements in $\mathcal{P}_D$, that lead to different values of the disparities.
Since disparities are differences of nonlinear functionals of the coupling, we need to construct the couplings very carefully to achieve unambiguously different disparity values. 
See Appendix \ref{proofsec: prop: minimal-unid} for the proof. 
}

\subsection{Partial Identification Set of Disparities}\label{sec: PI-size}
\edit{
In the last section we showed that DD, TPRD, and TNRD (and symmetrically also PPVD and NPVD) are generally \emph{not} identifiable from the two separate datasets. Next, we will characterize exactly how identifiable or unidentifiable they are by characterizing the partial identification set of all disparity values that agree with the observed data, and possibly additional assumptions that reflect prior knowledge.
}

\edit{
Each disparity measure in \cref{sec: disparity-measure} can be viewed as a function of the true distribution of $(\hY, Y, A, Z)$, so we generically denote it as $\delta(a,b;\pr)$. The \textit{partial identification set} of this disparity measure of interest given observed data information (encoded by $\mathcal{P}_D$ defined in \cref{eq: P-D}) and extra assumptions (encoded by $\mathcal{P}_A$)\footnote{\edit{If no extra assumption is imposed, then $\mathcal{P}_A$ is the set of all joint distributions, so that $\mathcal{P}_D \cap \mathcal{P}_A = \mathcal{P}_D$.}} is defined as follows: 
	\begin{align}\label{eq: PI-set-abstract}
	\Delta(\mathcal{P}_D \cap \mathcal{P}_A) = \left\{\delta(a,b;\pr'): \pr' \in \mathcal{P}_D \cap \mathcal{P}_A\right\}.
	\end{align}
We will add subscripts such as $\vphantom{\Delta}_\DD$ or $\vphantom{\Delta}_\TPRD$ to indicate the set for a particular disparity measure.
The partial identification set in \cref{eq: PI-set-abstract} is the smallest set containing all possible values of the disparity measures that agree with both the observed data and possibly extra assumptions. 
	Each disparity value in this set is given by one valid full joint distribution that is compatible with the observed data and extra assumptions, and any disparity value outside this set is ruled out by either the observed data or the assumptions. 
	A natural question is when are these smallest possible sets also actually small. We next discuss different scenarios where the sets can be small or large.
}
\edit{
\paragraph{Informative Proxies.} 
If the proxies are very predictive, then the observed data alone may be informative enough to sufficiently pin down the the disparity measures. At the extreme, if proxies are perfectly predictive of either the outcomes or the protected class, then the partial identification sets collapse into singletons, i.e., the disparity measures are uniquely identified from the observed data. We formalize this in the following proposition. 
\begin{proposition}\label{prop: perfect-pred}
	Given marginal distributions $\pr(\hY, Y, Z)$, $\pr(A, Z)$, if assumptions of \Cref{prop: minimal-unid}(ii) are not satisfied, i.e., for almost all $z$, either $\pr(\hY=\hy, Y = y \mid Z = z) \in \{0, 1\}$ for $\yhyrange$, or $\pr(A=\alpha \mid Z = z) \in \{0, 1\}$ for $\arange$, then $\mathcal{P}_D$ is a singleton, and hence $\Delta(\mathcal{P}_D)$ for any disparity measure in \Cref{sec: set-up} is also a singleton.  
\end{proposition}
\proof{Proof.}
According to the Fr\'echet-Hoeffding inequality in \Cref{prop: FH-coupling}, any valid full joint distribution agreeing with the observed data, i.e., an element $\pr' \in \mathcal{P}_D$, has to satisfy that 
\begin{align}
&\pr'(A = \alpha, \hY = \hy, Y = y \mid Z) \le \min\{\pr(\hY=\hy, Y = y \mid Z), \pr(A=\alpha \mid Z)\}, \label{eq: FH-right}\\
&\pr'(A = \alpha, \hY = \hy, Y = y \mid Z) \ge \max\{\pr(\hY=\hy, Y = y \mid Z) + \pr(A=\alpha \mid Z)- 1, 0\}. \label{eq: FH-left}
\end{align}
Under the stated assumptions, the right-hand sides of \cref{eq: FH-right,eq: FH-left} are equal. Thus, the full joint distribution is uniquely determined by the marginals and $\mathcal{P}_D$ is a singleton. 
\endproof
This shows that the conditions of \Cref{prop: minimal-unid} are the tight necessary \emph{and} sufficient conditions for identifiability from marginals alone.
If proxies are not perfect but are very predictive, either of protected class, of outcomes, or of both, then the endpoints of Fr\'echet-Hoeffding inequality (i.e., right-hand sides of \cref{eq: FH-left,eq: FH-right}) are not exactly equal but they are still close. Consequently, the partial identification sets will be small. This is the case we observe in \cref{sec: warfarin} when using very informative genetic proxies for race. 
}
\edit{
	\paragraph{Strong Assumptions on the Joint Distributions.} 
For the case of DD, \cite{chen2019fairness} discussed a conditional independence assumption that admits an unbiased proxy-based estimator. 
	Actually, this assumption is sufficient for the identifiability of disparities more generally.   
	\begin{proposition}\label{prop: cond-ind}
	If we assume that $Y,\hY \independent A \mid Z$, i.e., 
	\begin{align*}
	\mathcal{P}_A = \left\{\pr': \pr'(\hY = \hy, Y = y, A = \alpha \mid Z = z) = \pr'(\hY = \hy, Y = y \mid Z = z)\pr'(A = \alpha \mid Z = z), \forall \alpha, \hy, y, z\right\},
	\end{align*}
	then $\mathcal{P}_D\cap \mathcal{P}_A$ is a singleton, and hence $\Delta(\mathcal{P}_D\cap \mathcal{P}_A)$ for any disparity measure is also a singleton.  
	\end{proposition}
	\proof{Proof.}
	Any $\pr'\in\mathcal{P}_D \cap \mathcal{P}_A$ satisfies for $\zrange, \yhyrange, \arange$, 
$
	\pr'(\hY = \hy, Y = y, A = \alpha \mid Z = z) = \pr(\hY = \hy, Y = y \mid Z = z)\pr(A = \alpha \mid Z = z).
$
	Since this is uniquely determined by the marginals, $\mathcal{P}_D \cap \mathcal{P}_A$ contains only a single element.
	\endproof
	Although the conditional independence assumption is indeed very informative, it may be too unrealistic in practice. 
	Indeed, the proxies $Z$ that can be observed on both datasets are usually low-dimensional (e.g., surname and geolocation), so they are unlikely to capture all dependence between the outcomes and the protected class. Therefore, although imposing strong assumptions like this may help identification, it can also result in misleading conclusions, considering that these assumptions are often wrong in reality. 
}
\edit{
\paragraph{Un- or weakly informative proxies and no or weak assumptions.}
If the observed datasets alone are not highly informative, and we are not willing to impose overly stringent assumptions on the unknown joint distribution, we generally end up with partial identification sets with nontrivial size. 
For example, in \Cref{sec: hmda}, we find that using geolocation and income as proxies result in quite large partial identification sets in a lending example. 
Imposing additional mild smoothness assumptions (\Cref{sec: PIsmooth}) narrowed down the set based on income proxies only slightly.
In this case, the size of partial identification sets exactly captures the ambiguity in learning disparity measures based on the observed data and imposed assumptions. 
Large sets are not meaningless: they serve as an important warning about drawing any conclusions from highly flawed data. 
And, even large sets may be informative of the presence of disparities when they are well-separated from zero.
}

\section{Closed-form Partial Identification Sets of Disparities for Binary Protected Class Attribute}\label{sec: PI-binary}
 \edit{
 In this section, we show that the partial identification set in \Cref{eq: PI-set-abstract} has closed-form solutions when we consider a binary protected class (i.e., $\mathcal{A} = \{a, b\}$) without imposing any additional assumption (i.e., $\mathcal{P}_A$ does not impose any constraint and $\mathcal{P}_D \cap \mathcal{P}_A = \mathcal{P}_D$).
 }

\edit{We first reformulate the partial identification set in \cref{eq: PI-set-abstract} for different disparity measures in terms of weighted representations that are more amenable to analysis.} For any functions $w_\alpha(\hy,z)$ and $\tw_\alpha(\hy,y,z)$, we define, respectively,
\edit{ 
\begin{align}
\mu(\alpha; w) &:= 
    \frac{\expect\big[\bwuhyz\; \hY\big]}
                    {\pr(A = \alpha)}, 
                \label{eq: group-mean-reformula} \\ 
  \mu_{\hy y}(\alpha; \tw) &:=      \frac{\expect\big[\btwuhyyz \indy\indhy\big]}
   {\expect\big[\btwuhyyz \indy\indhy + \btwuhyyz \indy\ind(\hY \ne \hy)\big]},      \label{eq: cond-group-mean-reformula}
\end{align}
Furthermore define $\wsuhyz, \wsuhyyz$ as \textit{the conditional probabilities of protected class given outcomes and proxies}:}
$$
   \wsuhyz \coloneqq 
   \pr(A = \alpha \mid \hY = \hy, Z = z)
   ,\quad\wsuhyyz \coloneqq 
   \pr(A = \alpha \mid \hY = \hy, Y = y, Z = z),
$$
\edit{such that DD, TPRD, TNRD satisfy $\delta_{\DD}(a, b) = \mu(a;w^*) - \mu(b;w^*)$, $\delta_{\TPRD}(a, b) = \mu_{11}(a;\tw^*) - \mu_{11}(b;\tw^*)$, and $\delta_{\TNRD}(a, b) = \mu_{00}(a;\tw^*) - \mu_{00}(b;\tw^*)$, respectively. 
}

\edit{
These weighted representations conveniently separate the identifiable and \emph{un}identifiable parts of the disparities. Indeed,} for any fixed functions $w,\tw$, both $\mu(\alpha; w)$ and $\mu_{\hy y}(\alpha; \tw)$ are \emph{identifiable} from just the marginal distribution $\pr(\hY,Y,Z)$ since every term is just an expectation over this distribution.
\edit{On the other hand, $w^*,\tw^*$, which depend upon the unidentifiable full joint distribution $\pr(A,\hY,Y,Z)$, are themselves unidentifiable and therefore render the disparities, which depend upon them, unidentifiable. 
Although the true $w^*,\tw^*$ are unidentifiable, we can construct the set of all possible values of these unknown conditional probabilities that agree with the observed data. For any set $\mathcal{P}$ of joint distributions, define:
\begin{align}
&\mw(\mathcal P) = \left\{w: \wuhyz = \pr'(A = \alpha \mid \hY = \hy, Z = z),\, \forall \hy, y, z,\, \pr' \in \mathcal{P} \right\} \label{eq: weight-pd1}\\
&\tmw(\mathcal P) = \left\{\tw: \wuhyyz = \pr'(A = \alpha \mid \hY = \hy, Y = y, Z = z),\, \forall \hy, y, z,\,  \pr' \in \mathcal{P} \right\} \label{eq: weight-pd2}
\end{align}
Then, we can characterize the partial identification sets of disparities simply by these sets of conditional probabilities: 
\begin{proposition}\label{prop: w-formulation} For any set $\mathcal P$ of joint distributions on $(A,\hY,Y,Z)$, we have
$
\Delta_\DD(\mathcal{P})=\braces{\mu(a;w)-\mu(b;w)\;:\:w\in\mw(\mathcal P)}$,
$\Delta_\TPRD(\mathcal{P})=\braces{\mu_{11}(a;\tw)-\mu_{11}(b;\tw)\;:\:\tw\in\tmw(\mathcal{P})}$, and
$\Delta_\TNRD(\mathcal{P})=\braces{\mu_{00}(a;\tw)-\mu_{00}(b;\tw)\;:\:\tw\in\tmw(\mathcal{P})}.
$
\end{proposition}
In particular, \cref{prop: w-formulation} holds for $\mathcal P_D$.
In the following proposition, we give explicit formulae for $\mw(\mathcal{P}_D),\,\tmw(\mathcal{P}_D)$ in terms of the Law of Total Probability (LTP) constraints as in \Cref{def: coupling}. 
\begin{proposition}\label{prop: sharp-set}
Given marginals $\pr(A\mid Z),\ \pr(\hY,Y,Z)$, we have
\begin{align*}
       \mw(\mathcal{P}_D) &= 
            \bigg\{w~~:~~ \begin{array}{l}
                \sum_{\hy \in \{0, 1\}} \wuhyz \phyz = \puz,\\ \sum_{\arange} \wuhyz = 1, 
                0 \le \wuhyz \le 1,  \text{ for any } \alpha, z, \hy\end{array}
            \bigg\} \\
        \tmw(\mathcal{P}_D) &=
            \bigg\{\tw~~:~~\begin{array}{l} 
                \sum_{\yhyrange} \wuhyyz \phyyz  = \puz, \\\sum_{\arange} \wuhyyz = 1, 
                0 \le \wuhyyz \le 1, \text{ for any } \alpha, z, \hy, y 
            \end{array}\bigg\}.
\end{align*}
\end{proposition}
}

Based on \Cref{prop: w-formulation,prop: sharp-set}, we can show that the partial identification sets of DD, TPRD, TNRD for binary protected class without imposing extra assumptions actually have closed-form solutions.
	\begin{proposition}[Closed-form set for DD]\label{prop: dd binary} Let
\begin{align*}
&\textstyle w^L_\alpha(\hy,z)=\max\braces{0,\ 1 + \frac{\puz - 1}{\phyz}},&&
 w^U_\alpha(\hy,z)= \min\braces{1,\ \frac{\puz}{\phyz}},
\end{align*}
Then
\begin{align}\label{eq: DD-binary}
\Delta_\DD(\mathcal{P}_D)
&=[\mu(a; w^L)-\mu(b; w^U),\ \mu(a; w^U)-\mu(b; w^L)].
\end{align}
\end{proposition}
\proof{Proof.}
\edit{Notice that $[\phyz w^L_\alpha(\hy,z),\ \phyz w^U_\alpha(\hy,z)]$ are exactly the endpoints of the Fr\'echet-Hoeffding inequalities in \cref{eq: FH-coupling} for the coupling set $\Pi(\pr(\hY \mid Z = z), \pr(A \mid Z = z))$. According to \Cref{prop: FH-coupling,prop: sharp-set}, the set $\mw(\mathcal{P}_D)$ has the following equivalent formulation:
	\begin{align}\label{eq: W-HF}
	\mw(\mathcal{P}_D) &= 
	\bigg\{w~~:~~ \begin{array}{l}
	\sum_{\hy \in \{0, 1\}} \wuhyz \phyz = \puz,\\ \sum_{\arange} \wuhyz = 1, 
	w^L_\alpha(\hy,z) \le \wuhyz \le w^U_\alpha(\hy,z),  \text{ for any } \alpha, z, \hy\end{array}
	\bigg\}.
	\end{align}
	Notice $\mw(\mathcal{P}_D)$ is compact and connected \edit{in $L_\infty$} and that the function $\mu(\alpha, w)$ is continuous in $w$ for $\alpha = a, b$. Thus, by \cref{prop: sharp-set}, the partial identification set is an interval:
	\[
	\Delta_\DD(\mathcal{P}_D) = \left[\min_{w \in \mw(\mathcal P_D)}\mu(a, w) - \mu(b, w),	\max_{w \in \mw(\mathcal P_D)}\mu(a, w) - \mu(b, w)\right]
	\]
	We derive the lower bound as an example, and the upper bound can be derived analogously. 
	According to \cref{eq: group-mean-reformula},
	\begin{align}\label{eq: dd-est-form}
	\mu(a, w) - \mu(b, w) = \frac{\expect\big[w_a(\hY,Z)\;\hY\big]}
	{\pr(A = a)} - \frac{\expect\big[w_b(\hY,Z)\;\hY\big]}
	{\pr(A = b)}.
	\end{align}
	Since $\mu(a, w) - \mu(b, w)$ is increasing in $w_a$ and decreasing in $w_b$,
	$
	\min_{w \in \mw(\mathcal{P}_D)}(\mu(a, w) - \mu(b, w))\geq
	\min_{w \in \mw(\mathcal{P}_D)}\mu(a, w) - \max_{w \in \mw(\mathcal{P}_D)}\mu(b, w) = \mu(a; w^L) - \mu(b; w^U).
	$
	Moreover, it is easy to verify that $w^\dagger = (w_{a}^L, w_{b}^U)$ satisfies the law of total probability constraints and is feasible in $\mathcal{P}_D$.
}\endproof
\edit{
The partial identification set given in \Cref{prop: dd binary}  concretely illustrates the general unidentifiability of demographic disparity under data combination: any element within the interval in \cref{eq: DD-binary} is a valid disparity value that agrees with the observed data information. 
In the unrealistically ideal case, if the proxy variables $Z$ are perfectly predictive of either $\hY$ or $A$, then we can verify that $w^L = w^U$, and the two interval endpoints in \Cref{eq: DD-binary} are equal.
}

\begin{proposition}[Closed-form sets of TPRD, TNRD]\label{prop: tprd binary} Let
\begin{align*}
 &\mu'_{\hy y}(\alpha; \tw,\tw')  
    := \frac{\expect\big[\bwuhyyz \indy\indhy\big]}
        { \expect\big[ \bwuhyyz \indy\indhy\big]+\expect\big[ \tw'_\alpha(\hY,Y,Z) \indy\ind(\hY \neq \hy)\big]},    \\
 &\tw^L_\alpha(\hy,y,z)=\max\braces{0,\ 1 + \frac{\puz - 1}{\phyyz}},\\
 &\tw^U_\alpha(\hy,y,z)= \min\braces{1,\ \frac{\puz}{\phyyz}}.
\end{align*}
Then
\begin{align}
\Delta_\TPRD(\mathcal{P}_D)
&=[\mu'_{11}(a; \tw^L,\tw^U)-\mu'_{11}(b; \tw^U,\tw^L),\ \mu'_{11}(a; \tw^U,\tw^L)-\mu'_{11}(b; \tw^L,\tw^U)], \label{eq: tprd-binary}\\
\Delta_\TNRD(\mathcal{P}_D)
&=[\mu'_{00}(a; \tw^L,\tw^U)-\mu'_{00}(b; \tw^U,\tw^L),\ \mu'_{00}(a; \tw^U,\tw^L)-\mu'_{00}(b; \tw^L,\tw^U)]. \label{eq: tnrd-binary}
\end{align}
\end{proposition}
\edit{
\Cref{prop: tprd binary} can be proved by following similar procedures in the proof of \Cref{prop: dd binary}. We again leverage a reformulation of $
\tmw(\mathcal{P}_D)
$ in terms of Fr\'echet-Hoeffding inequalities with $\tw^L$ and $\tw^U$ as extremal weights. 
Then $\mu'_{\hy y}(\alpha; \tw,\tw')$ is continuous in $(\tw,\tw')$ and it is increasing in $\tw$ but decreasing in $\tw',$\footnote{\edit{$\mu'_{\hy y}$ differs with $\mu_{\hy y}$ in \cref{eq: cond-group-mean-reformula} only in the two separate arguments $\tw$ and $\tw'$ to explicitly characterize the  monotonicity in two different directions, a property crucial for deriving the closed-form sets.}} 
 which would imply that the interval endpoints in \Cref{eq: tprd-binary,eq: tnrd-binary} indeed bracket the partial identification sets.
It remains to verify that the extremal weights are simultaneously feasible in 
$\tmw(\mathcal{P}_D)$ so that the interval endpoints are attained.
See \Cref{proofsec: prop: tprd binary} for details. 
Again, when the proxy variables $Z$ can predict either $(\hY, Y)$ or $A$ perfectly, we can easily verify that $\tw^L = \tw^U$, so the intervals in \Cref{eq: tprd-binary,eq: tnrd-binary} also collapse into singletons, but this is unrealistic.
}

\section{Extensions for General Partial Identification Sets}\label{sec: compute pi}

\edit{In this section, we discuss \textit{general} partial identification sets, 
allowing additional structural assumptions, such as \textit{smoothness restrictions}, and accommodating \textit{multiple-level protected class}.
}

\subsection{Additional Smoothness Assumptions}\label{sec: PIsmooth}
\edit{
	We first introduce smoothness restrictions to illustrate possible additional structural knowledge that can be used to restrict the partial identification sets. One might expect that, for two similar values $z,z'$, the two true joint distributions $\pr(A,Y\mid Z=z),\pr(A,Y\mid Z=z')$ are also similar (some limited amount of similarity is already implied by the Law of Total Probability when the given marginals are themselves smooth). There is no way to verify this from the separate datasets only, but such an assumption may be defensible based on domain knowledge and can help narrow down the possible values disparities may take. 
	We therefore further consider partial identification sets of disparities when we impose the following additional assumptions:
	\begin{align}
	&{\pr(A=\alpha\mid Y=y, Z=z)-\pr(A=\alpha\mid Y=y, Z=z')}
	\leq d(z,z')&&\forall \alpha,y,z,z'\label{eq: lipcond dd}\\
	&{\pr(A=\alpha\mid \hY=\hy,Y=y,Z=z)-\pr(A=\alpha\mid \hY=\hy,Y=y,Z=z')}
	\leq d(z,z')&&\forall \alpha,\hy,y,z,z'\label{eq: lipcond tprd}
	\end{align}
	where 
	$d(z,z')$ is a given metric. In particular, we encode the implicit Lipschitz constant by scaling the metric $d$ itself. We can then let $\mathcal{P}_\Lip$ be the set of all joints that satisfy \cref{eq: lipcond dd,eq: lipcond tprd}.}

\edit{
	\cref{eq: lipcond dd,eq: lipcond tprd} imply that the weight constraints $\mw(\mathcal{P}_D\cap\mathcal{P}_\Lip)$ and $\tmw(\mathcal{P}_D\cap\mathcal{P}_\Lip)$ corresponding to the Lipschitz assumption take the following forms respectively:
	\begin{align*}
	\mw(\mathcal{P}_D\cap\mathcal{P}_\Lip)&=\mw(\mathcal{P}_D)\cap\mw_\Lip,~\text{where}~
	\mw_\Lip:=\braces{w: w_\alpha(\hy,z)-w_\alpha(\hy,z')|\leq d(z,z')\ \forall z,z',\hy};\\
	\tmw(\mathcal{P}_D\cap\mathcal{P}_\Lip)&=\tmw(\mathcal{P}_D)\cap\tmw_\Lip,~\text{where}~
	\tmw_\Lip:=\braces{\tw:\tw_\alpha(\hy,y,z)-\tw_\alpha(\hy,y,z')|\leq d(z,z')\ \forall z,z',\hy,y}.
	\end{align*}
	Leveraging \cref{prop: w-formulation}, we can translate this to the partial identification sets for DD, TPRD, and TNRD when we assume \cref{eq: lipcond tprd,eq: lipcond dd}.
	In particular, \cref{prop: sharp-set}
	and the above
	provide an explicit form for these sets.
	To actually compute their endpoints we now need to solve an optimization problem. For generality, we consider this optimization problem in the context of a multiple-level protected class attribute, which we study next.
}

\subsection{Multiple-level Protected Class Attribute}\label{sec-generalpi-multiplelevel}
\edit{
We now consider the most general case and study the partial identification set of all \textit{simultaneously achievable} disparities for multiple groups, potentially imposing additional assumptions such as smoothness.
Specifically, 
letting $\mathcal A_0 \coloneqq \mathcal{A}\setminus \{a\}$ and $\delta(a,b;\pr)$ be any of the disparities defined in \cref{sec: disparity-measure},
we consider the \emph{multivariate} partial identification of all pairwise disparities:
\begin{align}\label{eq: PI-set-abstract-multi}
\Delta(\mathcal{P}_D \cap \mathcal{P}_A) = 
\left\{(\delta(a,b;\pr'))_{b\in\mathcal{A}_0}: \pr' \in \mathcal{P}_D \cap \mathcal{P}_A\right\}\subset\R{\abs{\mathcal A}-1}.
\end{align}
Note that for any $b,b'$, $\delta(b,b';\pr')=\delta(a,b';\pr')-\delta(a,b;\pr')$ so that the above set characterizes all simultaneously achievable pairwise disparities, regardless of the choice of $a$. We can also extend the above general approach to linear combinations of multiple disparity measures at the same time.}

\edit{Next, we note that with $\mw(\mathcal P),\tmw(\mathcal P)$ as defined in \cref{eq: weight-pd1,eq: weight-pd2}, we have the following generalization of \cref{prop: w-formulation}:
\begin{proposition}\label{prop: w-formulation-multi} For any set $\mathcal P$ of joint distributions on $(A,\hY,Y,Z)$, we have
$\Delta_\DD(\mathcal{P})=\braces{(\mu(a;w)-\mu(b;w))_{b\in\mathcal A_0}\;:\:w\in\mw(\mathcal P)}$,
$\Delta_\TPRD(\mathcal{P})=\braces{(\mu_{11}(a;\tw)-\mu_{11}(b;\tw))_{b\in\mathcal A_0}\;:\:\tw\in\tmw(\mathcal{P})}$, and
$\Delta_\TNRD(\mathcal{P})=\braces{(\mu_{00}(a;\tw)-\mu_{00}(b;\tw))_{b\in\mathcal A_0}\;:\:\tw\in\tmw(\mathcal{P})}$.
\end{proposition}}

\edit{
Since these sets are multivariate, they have more than just two ``endpoints.'' In particular, we characterize these sets by computing their \emph{support functions}. Given a set $\Theta\subseteq\R d$, its support function is given by
$h_{\Theta}(\rho)=\sup_{\theta\in\Theta}\rho^\top\theta$. Not only does the support function provide the maximal and minimal contrasts achieved over a set, it also exactly characterizes its convex hull \citep{rockafellar2015convex}. That is, 
$
{\op{Conv}\prns{\Theta}:=\{\sum_{j=1}^m\lambda_j\theta_j:m\in\mathbb N,\,\theta_j\in\Theta,\lambda_j\geq0,\sum_{j=1}^m\lambda_j=1\}=\braces{\theta:\rho^\top\theta\leq h_{\Theta}(\rho), ~ \ \forall\rho \text{ s.t. } \|\rho\| = 1}}
$.\footnote{\edit{Note that $\Delta_{\DD}(\mathcal P)$ is convex as long as $\mathcal W(\mathcal P)$ is convex, since $\mu(\alpha, w)$ is affine in $w$; both $\mathcal W(\mathcal P_D)$ and $\mathcal W(\mathcal P_D)\cap\mathcal W_\Lip$ are convex. On the other hand, $\Delta_{\TPRD}(\mathcal P),\Delta_{\TNRD}(\mathcal P)$ are generally not convex in the non-binary setting and taking their convex hull provides the smallest convex outer approximation to them.}}
In the following, we characterize the support functions. In \Cref{sec: inference-general}, we discuss their computation and estimation from data and how to use this to visualize the partial identification set.}
\paragraph*{Demographic Disparity.} 
\edit{We first consider the simpler case of demographic disparity.
}
\begin{proposition}\label{prop: dd lp}
Let $\mathcal{P}_A$ be given. Then
	\begin{align*}
	h_{\Delta_\DD(\mathcal{P}_D \cap \mathcal{P}_A)}(\rho)\ =\ 
	\max_{w\in \mw(\mathcal{P}_D) \cap \mw(\mathcal{P}_A)}\;
	\sum_{b\in\mathcal A_0}\rho_b\Biggl(
	\frac{\Efb{w_a(\hY,Z)\hY}}{\pr(A=a)}-
	\frac{\Efb{w_b(\hY,Z)\hY}}{\pr(A=b)}
	\Biggr).
	\end{align*}
\end{proposition}
\edit{
\Cref{prop: dd lp} follows immediately \edit{from  \cref{prop: w-formulation-multi,eq: dd-est-form}}. 
{When either $\mathcal{P}_A$ imposes no restrictions or
$\mathcal{P}_A=\mathcal P_\Lip$}, the above gives an 
infinite  linear program since both the law of total probability constraint $\mw(\mathcal{P}_D)$ and the Lipschitz constraint $\mw_{\Lip}$ are linear in $w$. 
}

\paragraph*{Classification Disparity.} We next consider the case of classification disparities. For a concise and clear exposition, we focus on the case of TPRD. \edit{Note that $\Delta_\TPRD(\mathcal{P}_D \cap \mathcal{P}_A)$ is generally a nonconvex set.} The case of TNRD can be symmetrically handled\edit{. 
 	\begin{proposition}\label{prop: tpr flp}
	Let $\mathcal{P}_A$ be given. Then
	\begin{align}
	h_{\Delta_\TPRD(\mathcal{P}_D \cap \mathcal{P}_A)}&(\rho)\ = \textstyle
	\max_{
		t\in\R{\mathcal A}:t \geq 1
	}
	\phi(\rho;t)
	, \label{eq: tpr flp parametrized}
	\\\notag\phi(\rho;t)
	= \max_{\tilde u}\ &\textstyle\sum_{b\in\mathcal A_0}\rho_b\prns{
		{\Efb{\tilde u_a(\hY,Y,Z)Y\hY}}-
		{\Efb{\tilde u_b(\hY,Y,Z)Y\hY}}}
\\\notag\text{s.t.}\ 
	& \textstyle\Efb{\tilde u_\alpha(\hY,Y,Z)Y}=1, \qquad \forall\arange, \\\notag
	& \textstyle\sum_{\arange} \frac{\tilde u_\alpha(\hy,y,z)}{t_\alpha} = 1, \qquad \forall\hyrange,\yrange,\zrange, \\\notag
	& \textstyle\sum_{\yhyrange} \tilde u_\alpha(\hy,y,z) \phyyz  = \puz t_\alpha, \quad \forall\arange,\zrange, \\\notag
	& \textstyle\tilde u_\alpha(\hy,y,z) \geq 0, \qquad\forall\arange,\hyrange,\yrange,\zrange, \\\notag
	& \textstyle(\tilde u_\alpha/t_\alpha)_{\alpha \in \mathcal{A}}\in\tmw(\mathcal{P}_A). 
	\end{align}
\end{proposition}}%
\edit{%
\Cref{prop: tpr flp} follows by applying a Charnes-Cooper transformation \citep{charnes1962programming} within each class to our characterization of the partial identification set in \cref{prop: w-formulation-multi} to deal with the linear-fractional terms in \cref{eq: cond-group-mean-reformula}.
See \cref{proofsec: prop: tpr flp} for the detailed proof. 
 	While \Cref{eq: tpr flp parametrized} is generally a non-convex optimization problem, the inner problem $\phi(\rho,t)$ is a linear program whenever $\tmw(\mathcal P_A)$ is the product of polyhedra over $\arange$, such as $\tmw(\mathcal P_A) = \tmw_{\op{Lip}}$.
}

\section{Implementation, Estimation, and Inference}\label{sec-inference}

\edit{In this section, we discuss how to implement our approach in practice in order to go from actual data to assessments of disparities. Specifically, in previous sections, we characterized the partial identification sets for disparity measures in terms of the two population distributions $\pr(A, Z),\ \pr(\hY,Y,Z)$: these sets are \textit{deterministic} population objects that reflect the \textit{intrinsic ambiguity} of disparities given only marginal information. In practice, we are given \emph{data} rather than marginal distributions. The question we address in this section is how to \emph{estimate} the partial identification sets from data. We further discuss the consistency of our estimates and inferential procedures for constructing confidence intervals that characterize the \emph{additional} {uncertainty} due to finite-sample variability.}

\edit{In this section, instead of assuming access to the population-level marginal distributions, we assume we are given finite-sample datasets.
Let $\npri$ and $\naux$ denote the sample size of the \textit{primary} and \textit{auxiliary} datasets, respectively.
Our combined dataset is
\[
\{(\hY_i, Y_i, Z_i)_{i = 1}^{\npri}, (A_i, Z_i)_{i = \npri + 1}^{n}\}, \text{ with total sample size } n = \npri + \naux,
\]
where the first $\npri$ units form the primary dataset and the latter $\naux$ units form the auxiliary dataset.
We suppose the data satisfies \cref{asn-samedist}
 and that observations in these two datasets are {independent}. 
We assume that as $n$ grows to infinity, the proportion of the primary dataset $r_n = \npri/n$ converges to a limiting proportion $r$, i.e., $r_n \to r$. 
Since the primary data are typically more expensive to acquire than the auxiliary data, we focus on the setting where the primary dataset is asymptotically of comparable or smaller size,  i.e., $0 \le r < 1$.\footnote{\edit{If instead the auxiliary dataset is of smaller size, then we can focus on estimators that converge at rate of $O(\sqrt{\naux})$. For example, in \cref{eq: asymp-dist-dd1,eq: asymp-dist-dd2} of \cref{thm: asymp-dist-dd}, we can use scaling factor of $\sqrt{\naux}$ instead of $\sqrt{\npri}$ to get similar asymptotic normality result. These two different scaling factors are asymptotically equivalent when $\naux \asymp \npri$, but may differ when $r=0$ or $r=1$. For brevity, we only allow the first. The latter can be handled symmetrically.
}} 
}

\edit{
According to \cref{sec: PI-binary,sec: compute pi}, the partial identification sets of disparity measures involve the conditional probabilities of the protected class $A$ and the outcomes $\hY, Y$ given proxies $Z$. We denote these conditional probabilities by the following shorthand notations:
\begin{align*}
	\eta_{\op{aux}}(\alpha, z) \coloneqq \pr(A = \alpha \mid Z = z), ~~ \eta_{\op{pri}}(\hy , z) \coloneqq \pr(\hY = \hy \mid Z = z), ~~ \tilde\eta_{\op{pri}}(\hy,y,z) \coloneqq \pr(\hY = \hy, Y = y \mid Z = z).
\end{align*}
Along with $\pr(Z)$, these specify the marginals $\pr(A, Z),\ \pr(\hY,Y,Z)$.
In practice, these conditional probabilities are usually unknown and need to be estimated from the primary and auxiliary datasets, respectively.
Since $\etaa,\etap,\etapt$ are discrete regression functions with features $Z$ (or, probabilistic classification models), they can each be learned using supervised learning on each of the datasets. For example, in \cref{sec: casestudies}, we use logistic and multinomial logistic regression. Other options include random forests or neural networks.
Since we are primarily interested in estimating the partial identification sets rather than these conditional probabilities, we refer to these conditional probabilities as \textit{nuisance parameters}, and estimators for them as \textit{nuisance estimators}.}
\edit{
\subsection{Estimation and Inference for the Case of Binary Protected Class Using Debiased Machine Learning}\label{sec: est-binary}
In \Cref{prop: dd binary,prop: tprd binary}, we prove that the partial identification sets of DD, TPRD, TNRD for binary protected class are intervals with closed-form endpoints. 
Therefore, estimating these two endpoints is enough to characterize the whole partial identification set. 
For simplicity, we only present the estimation and inference for the partial identification set of DD. The results for TPRD and TNRD are analogous but require more involved notation so we defer them to \cref{proofsec: tprd-asymp}.
}
\edit{
\paragraph{Reformulation of the estimand.}
According to \Cref{prop: dd binary}, estimating the partial identification sets of demographic disparity only requires estimating the bounds $\mu(a; w^L) - \mu(b; w^U)$ and  $\mu(a; w^U) - \mu(b; w^L)$.
In the following lemma, we consider a reformulation of $\mu(\alpha; w^L)$ and  $\mu(\alpha; w^U)$ that will be useful for constructing estimators for them.  
\begin{lemma}\label{lemma: est-dd-reform}
For $\mu(\alpha, \cdot)$ given in \cref{eq: group-mean-reformula}, and $w^L$ and $w^U$ given in \cref{prop: sharp-set},
\begin{align}\textstyle
\mu(\alpha, w^L) = \frac{1}{p_{\alpha}}\left\{\expect\left[\lambda^L_{\alpha}(Z; \eta)\right]  + \expect\left[\xi^L_{\alpha}( A, Z; \eta)\right] + \expect\left[\gamma^L_{\alpha}(\hY, Z; \eta)\right]\right\}, \label{eq: est-eq-dd-1}\\\textstyle
\mu(\alpha, w^U) = \frac{1}{p_{\alpha}}\left\{\expect\left[\lambda^U_{\alpha}(Z; \eta)\right] +  \expect\left[\xi^U_{\alpha}(A, Z; \eta)\right] + \expect\left[\gamma^U_{\alpha}(\hY, Z; \eta)\right]\right\}, \label{eq: est-eq-dd-2}
\end{align}
where $p_{\alpha} \coloneqq \pr(A = \alpha)$, $\eta = (\etap, \etaa)$, and  
\begin{align*}
&\lambda^L_{\alpha}(z; \eta) 
	= I^L_{\alpha}(z)\left(\etap(1, z) + \etaa(\alpha, z) - 1\right), ~ \xi^L_{\alpha}(A, z; \eta) 
	\coloneqq I^L_{\alpha}(z)(\ind(A = \alpha) - \etaa(\alpha, z)),  \\
&\gamma^L_{\alpha}(\hY, z; \eta) 
	\coloneqq I^L_{\alpha}(z)(\hY - \etap(1, z)), \text{ with } I^L_{\alpha}(z) \coloneqq \ind\left(\etap(1, z) + \etaa(\alpha, z) - 1 \ge 0 \right), \\
&\lambda^U_{\alpha}(z; \eta) 
	= I^U_{\alpha}(z)\left(\etap(1, z) - \etaa(\alpha, z)\right)+ \etaa(\alpha, z), ~  \xi^U_{\alpha}(A, z; \eta) \coloneqq (1 - I^U_{\alpha}(z))(\ind(A = \alpha) - \etaa(\alpha, z)),\\
&\gamma^U_{\alpha}(\hY, z; \eta) \coloneqq I^U_{\alpha}(z)(\hY - \etap(1, z)),  \text{ with } I^U_{\alpha}(z) \coloneqq \ind\left(\etap(1, z) - \etaa(\alpha, z) \le 0\right).
\end{align*}
\end{lemma}
}
\edit{
It is straightforward to verify that $\expect\left[\xi^L_{\alpha}( A, Z; \eta) + \gamma^L_{\alpha}(\hY, Z; \eta)\right] = \expect\left[\xi^U_{\alpha}( A, Z; \eta) + \gamma^U_{\alpha}(\hY, Z; \eta)\right] = 0$, so they are not necessary for characterizing $\mu(\alpha, w^L)$ and $\mu(\alpha, w^U)$. However, incorporating these augmentation terms is very useful for estimation: when we use estimated values of the nuisance parameters $\eta$ instead of the unknown true values, 
these augmentation terms effectively debias the final partial identification bound estimators so that estimation errors of $\eta$ only have negligible effect.
In fact, by leveraging a cross-fitting strategy \citep{chernozhukov2018double} to estimate $\eta$, we can prove that our final bound estimators are asymptotically equivalent to the infeasible estimators that we get by plugging in the \emph{true} $\eta$ into \cref{eq: est-eq-dd-1,eq: est-eq-dd-2} and taking empirical averages (\Cref{thm: asymp-dist-dd} below). 
In \cref{proofsec: failure-naive}, we illustrate that estimators without these augmentation terms generally have intractable asymptotic distributions. 
Similar debiasing approaches based on extra augmentation terms have been also used in causal inference and missing data literature \citep[e.g., ][]{scharfstein1999adjusting,chernozhukov2018double}. 
\begin{algorithm}[t!]
\caption{Estimation of $\Delta_\DD(\mathcal P_D)$ for binary-valued protected class attribute} \label{alg-dml}
\edit{\begin{algorithmic}[1]
\State Input: number of folds $K$, nuisance estimation procedures
\State Randomly partition the two datasets into $K$ disjoint even folds: $\mathcal{I}_{\op{pri}} = \{1, \dots, \npri\}=\mathcal{I}_{1, \op{pri}}\cup\cdots\cup\mathcal{I}_{K, \op{pri}}$, $\abs{\abs{\mathcal{I}_{k, \op{pri}}}-\npri/K}\leq1$, $\mathcal{I}_{\op{aux}} = \{\npri+1, \dots, n\}=\mathcal{I}_{1, \op{aux}}\cup\cdots\cup\mathcal{I}_{K, \op{aux}}$, $\abs{\abs{\mathcal{I}_{k, \op{aux}}}-\naux/K}\leq1$.
\State Set $\mathcal{I}_{k} = \mathcal{I}_{k, \op{pri}} \cup \mathcal{I}_{k, \op{aux}}$ and let $\expectnk, \expectnpk, \expectnak$ be the sample average over the $k^\text{th}$ fold in the combined, primary, and auxiliary datasets, respectively. For example, $\expectnk \lambda^L_{\alpha}(Z; \eta) = \frac{1}{\abs{\mathcal{I}_k}}\sum_{i \in \mathcal{I}_k} \lambda^L_{\alpha}(Z_i; \eta)$.
\State Set $\hat{p}_{\alpha} = \frac{1}{K}\sum_{k = 1}^K \expectnak\left[\ind(A = \alpha)\right]$.
\For{$k=1,\dots,K$}: 
\State Train $\etapk$ on $\{(\hY_i, Z_i):i \in \mathcal{I}_{\op{pri}}\setminus \mathcal{I}_{k, \op{pri}}\}$.
\State Train $\etaak$ on $\{(A_i, Z_i):i \in \mathcal{I}_{\op{aux}}\setminus \mathcal{I}_{k, \op{aux}}\}$.
\State Set $\etak =(\etapk, \etaak)$.
\EndFor
\For{$\arange$}: compute 
\begin{align}\textstyle
\hat{\mu}(\alpha, w^L) = \frac{1}{\hat{p}_{\alpha}K}\sum_{k = 1}^K\left\{\expectnk\left[\lambda^L_{\alpha}(Z; \heta^{-k})\right]  + \expectnak\left[\xi^L_{\alpha}( A, Z; \heta^{-k})\right] + \expectnpk\left[\gamma^L_{\alpha}(\hY, Z; \heta^{-k})\right]\right\} \label{eq: est-dd-1},\\\textstyle
\hat{\mu}(\alpha, w^U) = \frac{1}{\hat{p}_{\alpha}K}\sum_{k = 1}^K\left\{\expectnk\left[\lambda^U_{\alpha}(Z; \heta^{-k})\right] +  \expectnak\left[\xi^U_{\alpha}(A, Z; \heta^{-k})\right] + \expectnpk\left[\gamma^U_{\alpha}(\hY, Z; \heta^{-k})\right]\right\} \label{eq: est-dd-2}.
\end{align}
\EndFor
\State Return the estimated partial identification set
\begin{align}
\hat{\Delta}_\DD(\mathcal P_D) = \left[\hat{\mu}(a, w^L) - \hat{\mu}(b, w^U), ~~ \hat{\mu}(a, w^U) - \hat{\mu}(b, w^L)\right] \label{eq: est-dd-3}.
\end{align}
\end{algorithmic}}
\end{algorithm}
}
\edit{
\paragraph{The estimator.}
Our estimator for the partial identification set is given in \cref{alg-dml}.
Our estimates for $\hat{\mu}(\alpha, w^L)$ and $\hat{\mu}(\alpha, w^U)$ are based on \cref{eq: est-eq-dd-1,eq: est-eq-dd-2} and a cross-fitting strategy: the nuisance estimator $\heta^{-k}$ is only applied to data in the $k^\text{th}$ fold, i.e., data not used to train $\heta^{-k}$. This prevents the nuisance estimators from overfitting to the data where they are evaluated. 
}
\edit{
\paragraph{Inference.}
We next prove that the estimated endpoints in \cref{eq: est-dd-3} are asymptotically normal with closed-form asymptotic variance. This allows us to construct confidence intervals. It also shows that we are largely invariant to how one fits $\eta$ and that no conditions except for a slow convergence rate are needed, which is appealing when one uses machine learning methods for this task.
\begin{theorem}\label{thm: asymp-dist-dd}
Suppose that the nuisance estimators converge at the following rate:
\[
	\left|\etapk(1, Z) - \etap(1, Z)\right| = O_p(\kappa_{\npri, \hY}), ~~ \left|\etaak(\alpha, Z) - \etaa(\alpha, Z)\right| = O_p(\kappa_{\naux, A}),\quad \alpha = a, b, \ k = 1, \dots, K.
\]
Assume the following conditions: for $\alpha = a, b$, 
\begin{enumerate}[label=(\roman*)]
\item ${p}_{\alpha} > 0$; 
\item there exist positive constants $m_1, m_2, c_1, c_2$ such that for $\arange$ and any $p \geq 0$,
\[\pr\left(0 \leq \left|\etap(1, Z) + \etaa(\alpha, Z) - 1 \right| \le p\right) \le c_1 p^{m_1}, ~~~ \pr\left(0 \le \left|\etap(1, Z) - \etaa(\alpha, Z) \right| \le p\right) \le c_2 p^{m_2};
\]
\item $\max\{\kappa_{\naux, A}, \kappa_{\npri, \hY}\}= o(\npri^{-1/(2+2m_1)})$,  $\max\{\kappa_{\naux, A}, \kappa_{\npri, \hY}\}= o(\npri^{-1/(2+2m_2)})$;
\item 
$|r - r_n|\kappa_{\npri, \hY} = o(\npri^{-1/2})$, $|r - r_n|\kappa_{\naux, A} = o(\npri^{-1/2})$.
\end{enumerate}
Then, as $n \to \infty$, the lower bound and upper bound estimators for demographic disparity with binary protected class are asymptotically normal:
\begin{align}
\sqrt{\npri}\left\{\left(\hat{\mu}(a, w^L) - \hat{\mu}(b, w^U)\right) - \left({\mu}(a, w^L) - {\mu}(b, w^U)\right)\right\} \overset{d}{\to} \mathcal{N}(0, V_L) \label{eq: asymp-dist-dd1}\\
\sqrt{\npri}\left\{\left(\hat{\mu}(a, w^U) - \hat{\mu}(b, w^L)\right) - \left({\mu}(a, w^U) - {\mu}(b, w^L)\right)\right\} \overset{d}{\to} \mathcal{N}(0, V_U) \label{eq: asymp-dist-dd2}
\end{align}
\begin{align*}
\text{where}~~~V_L 
	&= r\expect\left[\lambda_{a}^L(Z; \eta)/p_a - \lambda_{b}^U(Z; \eta)/p_b- \left({\mu}(a, w^L) - {\mu}(b, w^U)\right)\right]^2 \\
	&\phantom{=}+ \expect\left[\gamma_{a}^L(\hY, Z; \eta)/p_a - \gamma_{b}^U(\hY, Z; \eta)/p_b\right]^2 + \frac{r}{1 - r}\expect\left[ \xi_{a}^L(A, Z; \eta)/p_a - \xi_{b}^U(A, Z; \eta)/p_b\right]^2, \\
V_U
	&= r\expect\left[\lambda_{a}^U(Z; \eta)/p_a - \lambda_{b}^L(Z; \eta)/p_b - \left({\mu}(a, w^U) - {\mu}(b, w^L)\right)\right]^2 \\
	&\phantom{=}+ \expect\left[\gamma_{a}^U(\hY, Z; \eta)/p_a - \gamma_{b}^L(\hY, Z; \eta)/p_b\right]^2 + \frac{r}{1 - r}\expect\left[ \xi_{a}^U(A, Z; \eta)/p_a - \xi_{b}^L(A, Z; \eta)/p_b\right]^2.
\end{align*}
\end{theorem}
Condition (i) is needed for the problem to be well-defined: both classes need to be present to compare them.
Condition (ii) is a margin condition \citep{audibert2007fast} that characterizes the probability mass near the non-differentiable boundary. In particular, for $p=0$, it implies that ${\etap}(1, Z) + {\etaa}(\alpha, Z) - 1 \ne 0$ and ${\etap}(1, Z) - {\etaa}(\alpha, Z) \ne 0$ almost surely, which is trivially satisfied if $Z$ includes continuous variables. This ensures that even though $w^L$ and $w^U$ depend on non-smooth $\max$ and $\min$ operators respectively, $\mu(\alpha, w^L)$ and $\mu(\alpha, w^U)$ are still smooth functionals of the conditional probabilities $\eta$. Otherwise, statistical inference for non-smooth functionals is a notoriously difficult nonregular problem, and it is well-known that no estimator with well-behaved asymptotic distribution exits in this case \citep[e.g., ][]{laber2014dynamic,hirano2012impossibility}. 
Similar regularity conditions also appear in other partial identification literature to circumvent non-smoothness \citep[e.g., ][]{kennedy2018sharp,bonvini2019sensitivity}. 
Conditions (iii) requires that our nuisance estimators are consistent but only requires a slow, non-parametric rate, i.e., slower than $\npri^{-1/2}$. 
For example, if $m_1, m_2 \ge 1$, then conditions (iii) is satisfied if $\kappa_{\naux, A} = o_p(\npri^{-1/4})$ and $\kappa_{\naux, \hY} = o_p(\npri^{-1/4})$. This slow rate together with no other assumptions on our nuisance estimators means that the theorem holds even when we use flexible machine learning models to estimate nuisances (e.g., random forest, gradient boosting tree, neural networks with many neurons relative to $\npri$, etc.). 
Lastly, Condition (iv) requires that the observed ratio of primary to auxiliary data, $r_n$, is sufficiently similar to the asymptotic ratio. It is trivially satisfied if $r_n=r$ or $r_n-r=O(\npri^{-1/2})$, such as would be the case if $\npri\sim\op{Binomial}(n,r)$.}

\edit{
In the proof (\cref{profsec: asymp-dist-dd}), we show that the asymptotic distributions in \cref{eq: asymp-dist-dd1,eq: asymp-dist-dd2} are actually the same as distributions of the infeasible oracle estimators where we use the true values of nuisances, $\eta$. In other words, using the estimated value $\hat{\eta}$ instead of the unknown true value $\eta$ does not inflate the variance of our estimates. 
This is possible mainly because of the augmented formulation we derive in \cref{eq: asymp-dist-dd1,eq: asymp-dist-dd2} (see \cref{proofsec: failure-naive}). 
}

\edit{
The closed-form asymptotic variances in \cref{thm: asymp-dist-dd} suggest the following variance estimators:
\begin{align}\label{eq: var-est}
\hat{V}_L 
	&\textstyle= \frac{r_n}{K}\sum_{k = 1}^K\expectnk\left[\lambda_{a}^L(Z; \hat{\eta}^{-k})/\hat{p}_a - \lambda_{b}^U(Z; \hat{\eta}^{-k})/\hat{p}_b- \left(\hat{\mu}(a, w^L) - \hat{\mu}(b, w^U)\right)\right]^2  \\\notag
	&\textstyle\phantom{=}+ \frac{1}{K}\sum_{k = 1}^K\expectnpk\left[\gamma_{a}^L(\hY, Z; \hat{\eta}^{-k})/\hat{p}_a - \gamma_{b}^U(\hY, Z; \hat{\eta}^{-k})/\hat{p}_b\right]^2 \\\notag
	&\textstyle\phantom{=}+ \frac{r_n}{1 - r_n}\frac{1}{K}\sum_{k = 1}^K\expectnak\left[ \xi_{a}^L(A, Z; \hat{\eta}^{-k})/\hat{p}_a - \xi_{b}^U(A, Z; \hat{\eta}^{-k})/\hat{p}_b\right]^2, 
\end{align}
and $\hat V_U$ is similarly defined by swapping $L$ and $U$ everywhere above.
}

\edit{
We further prove in the following theorem that the asymptotic variance estimators above are consistent, and they can be used to construct confidence intervals for the partial identification sets. 
\begin{theorem}\label{corollary: CI-dd}
Under the assumptions of \Cref{thm: asymp-dist-dd},  $\hat{V}_L,\hat{V}_U$ are consistent: as $\npri \to \infty$,
\[
\hat{V}_L  \overset{p}{\to} V_L, ~~~ \hat{V}_U \overset{p}{\to} V_U.
\]
Therefore, we can construct the following $(1 - \beta)\times 100\%$ confidence interval 
\[
	\operatorname{CI} = [\hat{\mu}(a, w^L) - \hat{\mu}(b, w^U) - \Phi^{-1}(1 - \beta/2){\hat{V}_L}^{1/2}/\npri^{1/2}, ~~ \hat{\mu}(a, w^U) - \hat{\mu}(b, w^L) + \Phi^{-1}(1 - \beta/2){\hat{V}_U}^{1/2}/\npri^{1/2}]
\]
where $\Phi^{-1}$ is the quantile function of standard normal distribution. This confidence interval asymptotically covers the  partial identification set of DD  with  probability at least $1 - \beta$: 
\begin{equation}\notag
	\liminf_{\npri \to \infty} \pr\left(\Delta_\DD(\mathcal P_D)\subseteq \operatorname{CI}\right) \ge 1 - \beta.
\end{equation}
\end{theorem}
In \cref{sec: casestudies} (\cref{fig:hmda-CI,fig: warfarin-CI}), we illustrate how to use these confidence intervals to test whether a given disparity value (or a range) is compatible with the observed data information and thus belongs to the corresponding partial identification set.}

\edit{Note that the confidence interval above is \textit{conservative} in that its asymptotic coverage may \emph{exceed} $1-\beta$. In \cref{sec: cal-CI}, we present a calibrated confidence interval with asymptotic coverage \textit{exactly} $1-\beta$, albeit having a more complicated form. }

\subsection{General Partial Identification Sets}\label{sec: inference-general}
\edit{
We next discuss finite-sample estimation of general partial identification sets given in \cref{sec: compute pi}. That is, we discuss how we obtain a representation of the partially identified sets $\Delta(\mathcal{P}_D \cap \mathcal{P}_A)$ when we consider a multiple-level protected class attribute, impose smoothness restrictions in $\mathcal{P}_A$, or both. We propose an estimator for the support function using a linear program and prove it is statistically consistent.\footnote{\edit{Unlike the case in \cref{sec: est-binary}, statistical inference (confidence intervals) for general multivariate sets characterized by estimated support functions is an active research area \citep{molinari2019econometrics} and generally computationally burdensome, so we leave it for further research and focus on the consistency of our support function estimates.}}
We then describe how to use these support function estimates to visualize $\op{Conv}(\Delta(\mathcal{P}_D \cap \mathcal{P}_A))$.
For this section, we 
employ a simpler plug-in estimator based on nuisance estimators constructed on the whole primary and auxiliary datasets, respectively.}

 \paragraph{Demographic Disparity.}
\edit{
We first introduce the support function estimator for the case of demographic disparity, $h_{\Delta_\DD(\mathcal{P}_D \cap \mathcal{P}_A)}(\rho)$. The estimator applies for the case of multiple-leveled protected attributes with any linearly-representable additional constraints $\mw(\mathcal{P}_A)$, such as none or $\mw_\Lip$. 
 Given nuisance estimators $\hat\eta_{\aux},\hat\eta_{\pri}$ and letting $\expectnp$ denote computing sample averages over the primary dataset, we define our estimator as the following linear program:
 \begin{align*}
 \hat h_{\Delta_\DD(\mathcal{P}_D \cap \mathcal{P}_A)}(\rho)
 \ = 
 \ \max_{w}\ &
 \sum_{b\in\mathcal A_0}\rho_b\prns{{
 		\frac{\expectnp{[w_a(\hY,Z)\hY]}}{\expectnp[\hat\eta_{\aux}(a,Z)]} }
 		-
 	\frac{\expectnp{[w_b(\hY,Z)\hY]}}{\expectnp[\hat\eta_{\aux}(b,Z)]}
 } \\ 
  \text{s.t.}\ 
&0 \le \wuhyz \le 1,  \; 
  \forall\arange, \hy\in\{0,1\}, y\in\{0,1\}, 
  z\in\{Z_i\}_{i = 1}^{n}
 \\
  &\textstyle\sum_{\hy \in \{0, 1\}} \wuhyz \hetahyz = \hetaaz, 
  \;\;
   \sum_{\arange} \wuhyz = 1,
  \;\; 
 w \in \mw(\mathcal P_A) .
 \end{align*}
 We next show that the estimator is consistent. 
 \begin{theorem}\label{thm-consistency-dd}
	Assume that: 
	\begin{enumerate}[label=(\roman*)]
		\item $\sup_{\hyrange,
			z \in \mathcal Z 
		}\left|\hetahyz - \eta_{\pri}(\hy,z) \right| = o_p(1)$ and $\sup_{\alpha \in \mathcal A, 	z \in \mathcal Z }
		\left|   \hat \eta_{\aux} (\alpha,z) - \eta_{\aux} (\alpha,z) \right| 
		= o_p(1), $	 \label{asn-conditional-prob-consistency-dd} 
		\item $Z$ has finite support, i.e., $|\mathcal Z|$ is finite. 
	\end{enumerate}
	Then, for any $\rho$,
	$$
	\hat h_{\Delta_\DD(\mathcal P_D \cap \mathcal P_A)}(\rho
	)-
	h_{\Delta_\DD(\mathcal P_D \cap \mathcal P_A)}(\rho)
	\overset{p}{\longrightarrow} 0.
	$$	
\end{theorem}
Proving \cref{thm-consistency-dd} uses a stability analysis due to \cite{robinson1975stability} to bound the deviation of a linear program under stochastic perturbations to coefficients of the constraint matrix that arise from the estimation errors of the nuisance functions $ \hetaaz, \hetahyz$. 
}
\edit{In \Cref{prop-nuisance-az} of the Appendix, we discuss how to additionally obtain the asymptotic distribution of $\hat h_{\Delta_\DD(\mathcal P_D \cap \mathcal P_A)}$, under the assumption of unique primal and dual solutions.}
\edit{
\paragraph{Classification disparity.}
We next handle the general case for TPRD (TNRD is handled symmetrically).
Estimating the support function of $\Delta_{\TPRD}$ introduces additional challenges since the optimization problem that defines it is generally nonconvex
(see \cref{prop: tpr flp}).
We instead leverage the fact that it is the maximum of linear programs if $\tmw(\mathcal P_A)$ is linearly representable. 
}
\begin{algorithm}[t!]
\caption{Estimation of $\op{Conv}(\Delta)$ from support function estimates} \label{alg-suppfn}
\begin{algorithmic}[1]
	\State 
	\edit{Input: Support function estimator $\hat h_{\Delta}(\rho)$, contrast sample size $N_\rho$
	}
	\State 
	\edit{Sample contrast vectors, $\rho_1, \dots, \rho_{N_{\rho}}$, uniformly from the $(\vert \mathcal A \vert-1)$-dimensional unit sphere.}
	\For{\edit{$j=1,\dots,N_{\rho}$}}: 
\State 
\edit{
Solve $\hat h_{\Delta}(\rho_j)$, record the maximizer $\hat\delta_j\in\Delta$ such that $\hat h_{\Delta}(\rho_j)=\hat\delta_j^T\rho_j$.
}
	\EndFor
	\State \edit{Return 
	$\hat\Delta_\text{inner}=\op{Conv}(\{\delta_1,\dots,\delta_{N_\rho}\})$, 
	$\hat\Delta_\text{outer}={		\{\delta  \in \mathbb R^{\mathcal A_0} \colon
		\delta^\top \rho_j \leqslant 
		\hat h_{\Delta}(\rho_j)\;
		\forall j =1,\dots,N_{\rho}  \}.}$
	}
\end{algorithmic}
\end{algorithm}

\edit{
 The estimator for the support function, which computes the sample-level subproblem $\hat\phi(\rho;t)$ for a collection of values of $t$, $\mathcal T \subseteq \mathbb R^{\vert \mathcal A \vert } $, and the nuisance estimators
	$\hetahyyz,\hetaaz$ is:
	\begin{align}
\hat h_{\Delta_\TPRD(\mathcal{P}_D \cap \mathcal{P}_A)}&(\rho;\mathcal T)\ = \textstyle
\max_{
	t\in\mathcal T
}
\hat\phi(\rho;t)
, \label{eq: tpr flp parametrized}
\\\notag
\hat\phi(\rho;t)
= \max_{\tilde u}\ &\textstyle
 \sum_{b\in\mathcal A_0}\rho_b\prns{{\expectnp{
			[\tilde u_a(\hY,Y,Z)Y\hY
			]}}-
	{\expectnp{[\tilde u_b(\hY,Y,Z)Y\hY]}}
}
\\\notag\text{s.t.}\ 
&
\textstyle
\forall\alpha, \hy, y,
z\in\{Z_i\}_{i = 1}^{n},~~
\expectnp[\tilde u_\alpha(\hY,Y,Z)Y ]=1, ~~ 
(\tilde u_\alpha/t_\alpha)_{\alpha \in \mathcal{A}}\in\tmw(\mathcal{P}_A), 
\\  ~~& \textstyle\sum_{\yhyrange} \tilde u_\alpha(\hy,y,z)
\hetahyyz
= t_\alpha
\hetaaz,
 \sum_{\arange} \frac{\tilde u_\alpha(\hy,y,z)}{t_\alpha} = 1,
\\
&\textstyle \tilde u_\alpha(\hy,y,z) \geq 0.
\end{align}
}%
\edit{
In the following theorem, we show that the proposed support function estimator is point-wise consistent if $Z$ has only finitely many values and the nuisance estimators are uniformly consistent. 
\begin{theorem}\label{thm-consistency}
Assume that:
	\begin{enumerate}[label=(\roman*)]
		\item $\sup_{\hyrange,\yrange
			z \in \mathcal Z 
		}\left|
\hetahyyz
	- \tilde\eta_{\pri}(\hy,y,z)\right| \overset{p}{\to} 0$
	 and $\sup_{\alpha \in \mathcal A, 	z \in \mathcal Z }
	\left|   
\hetaaz %
	- \eta_{\aux}(\alpha,z)
	  \right| 
	\overset{p}{\to} 0$.\label{asn-conditional-prob-consistency} 
\item 
	There exists a positive constant $\nu$ such that 
	$\pr(A=\alpha, Y=1) \geq \nu, \;\; \forall \alpha \in \mathcal A$.
	\label{asn-conditional-prob-support-overlap}
\item Let $\mathcal T^{-1}$ be an $\epsilon_{\npri}$-covering of 
$\mathcal T_0^{-1}\coloneqq\{\tau\in \mathbb R^{\vert \mathcal A \vert  }  \colon \sum_{\alpha \in \mathcal{A}}\tau_\alpha = \expectnp \left[Y\right]; \  \nu\leq \tau_\alpha \leq 1,\;\arange \}$, i.e., $\min_{\tau'\in\mathcal T^{-1}}\norm{\tau - \tau'}_1 \leq \epsilon_{\npri}$ for any $\tau\in\mathcal T_0^{-1}$. Let $\mathcal T$ be the componentwise inverse of $\mathcal T^{-1}$.
	\item $Z$ has finite support, i.e., $|\mathcal Z|$ is finite. 
	\item 
	$\epsilon_{\npri}\to0$ as $\npri\to\infty$. \label{asn-grid-vanish}
\end{enumerate}
Then, for any $\rho$,
	$$
	\hat h_{\Delta_\TPRD(\mathcal P_D \cap \mathcal P_A)}(\rho;\mathcal T
		)-
		h_{\Delta_\TPRD(\mathcal P_D \cap \mathcal P_A)}(\rho)
\overset{p}{\to} 0
	$$	
\end{theorem}
The proof of \cref{thm-consistency} is similar to that of \cref{thm-consistency-dd} but also shows that the optimization problem is stable under approximation errors from the discretization, $\mathcal T$.
 Condition~\ref{asn-conditional-prob-support-overlap} ensures that we may restrict attention to a compact range for $t$. 
 Condition~\ref{asn-grid-vanish} ensures consistency as we consider a sequence of finer $t$-grids.
} 
\paragraph{Estimating and visualizing the partial identification set.}
 \edit{The procedures above estimate the support function of the partial identification set. It remains to actually estimate the partial identification set itself. Given a support function estimator, \cref{alg-suppfn} provides a procedure to obtain inner and outer approximations to the set (up to vanishing estimation errors in the support function) by sampling the contrast directions, $\rho$. These inner and outer approximations are polyhedra given explicitly either as the convex hull of a given set of points or as the intersection of halfspaces, respectively.
 As the number of contrasts sampled increases, the sets become closer. Either set can be visualized using standard tools for plotting convex hulls and polyhedra. We recommend to use the outer approximation since (up to vanishing estimation errors in the support function) it is guaranteed to contain the true partial identification set, and this is the set we use in \cref{sec: casestudies}.
}

\section{Case Studies}\label{sec: casestudies}
In the subsequent sections we consider applying our results and methods in two different case studies: mortgage credit decisioning and personalized Warfarin dosing. 

\subsection{Mortgage Credit Decisioning}\label{sec: hmda}
\edit{
	We consider assessing demographic disparity -- the simplest measure (see \cref{sec: disparity-measure,sec: literature} for others) that is relevant for the context
	of mortgage credit decisioning \citep{zhang2016assessing,chen2019fairness}: here, it measures the discrepancy in marginal approval rates between different racial groups. For groups, we consider White, Black, and  Asian and Pacific Islander (API).
}
\edit{
\paragraph*{Dataset, Proxy Variables, and Nuisance Estimation.}  
We demonstrate the partial identification set of demographic disparity using the public HMDA (Home Mortgage Disclosure Act) data set for US mortgage market.  
 This dataset contains self-reported race labels, and it has been used in the literature to evaluate proxy methods for race \citep{Baines, zhang2016assessing, chen2019fairness}.\footnote{The dataset can be downloaded from \url{https://www.consumerfinance.gov/data-research/hmda/explore}. This dataset includes mortgage loan application records in the U.S., which include self-reported race/ethnicity, loan origination outcome, geolocation (state, county, and census tract), annual income, loan amount, among other variables.} 
 However, this dataset is anonymized and does not include surname information, so we could not evaluate the popular BISG method exactly; it also does not contain default outcomes, so we only study demographic disparity. 
}

\edit{
We use a random $0.1\%$ subsample containing $14903$ loan application records for White, Black, and API applicants with annual income no more than
$\$100$K during 2011-2012 as the primary dataset, and the full sample of all records in this population as the auxiliary dataset. 
This mimics the fact that in BISG the primary dataset typically only contains information of a subset of units in the auxiliary data (decennial census data). 
}
We denote $\hY = 1$ if a loan application was approved or originated, and $\hY = 0$ if it was denied. 

We consider three different set of proxy variables for race: only geolocation (county), only annual income, and both geolocation and annual income. 
The distribution of race/ethnicity by these proxies can both be estimated from public records.
U.S. census Summary File I \citep{us20102010} contains race distributions for different geolocation levels, and the Annual Population Survey \citep{CPS} contains race distributions for different income brackets.

We estimate the conditional probabilities of race and decision outcome directly on the auxiliary dataset. 
When only geolocation is used as the proxy variable, we use the within-county race proportions and average loan acceptance rate to estimate the conditional probabilities of race and loan acceptance respectively. 
When only income is used as the proxy variable, we fit a logistic regression to estimate the conditional probability of loan acceptance, and a multinomial logistic regression to estimate the conditional probabilities of races. 
When both income and geolocation are used, we fit the logistic and multinomial logistic regressions with respect to income within each county.

\begin{figure}\centering
\includegraphics[width=\textwidth]{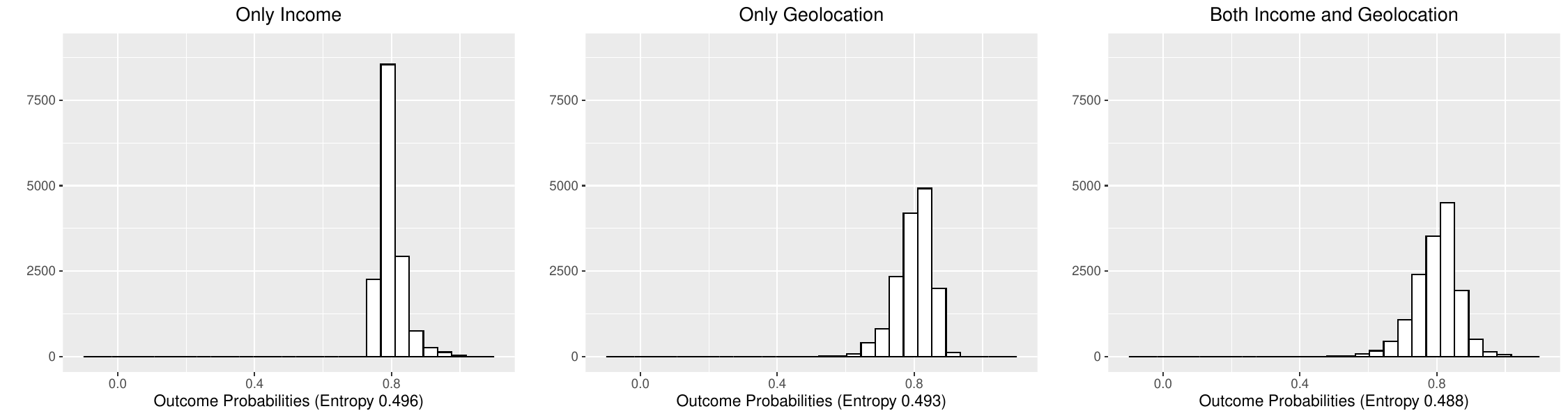}
\includegraphics[width=\textwidth]{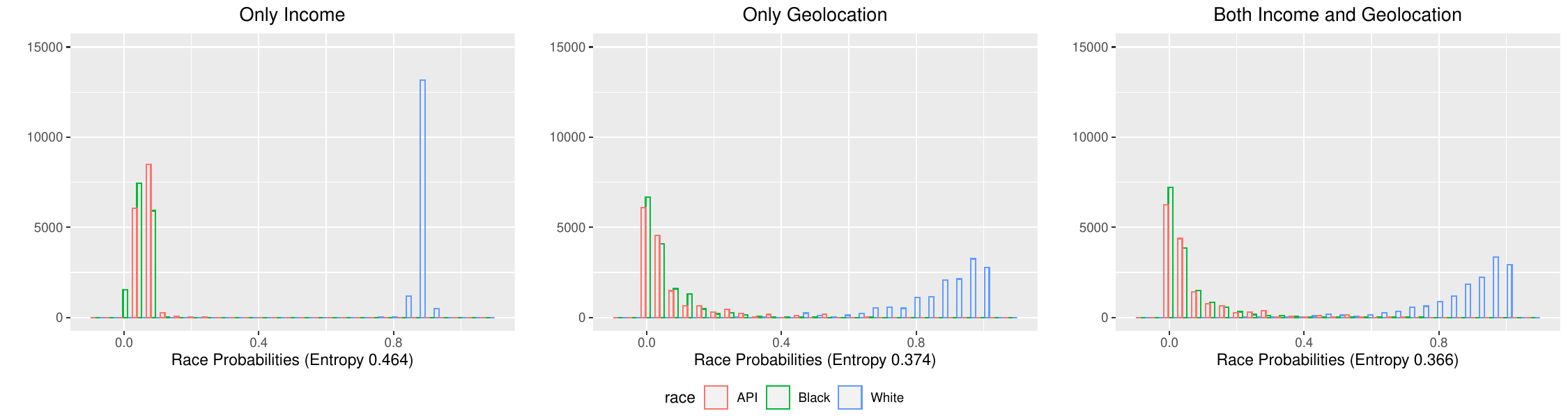}
\caption{Histograms of conditional probabilities of outcomes (upper row) and race (lower row) for different choices of proxies in the HMDA dataset, along with the resulting entropy.}
\label{fig:hmda-proxy-outcome-hists}
\end{figure}

Recall that the size of the partial identification set depends on the informativeness of the proxies about both protected class and outcomes (\cref{sec: PI-size}).
In \cref{fig:hmda-proxy-outcome-hists}, we show the histograms of the conditional probabilities for each race and, separately, for the positive outcome. We also report the (negative) entropy, which summarizes how predictive the proxies are. 
\edit{
For example, the entropy for race probabilities is $\expect[\sum_{\arange}\pr(A = \alpha \mid Z)\log \pr(A = \alpha \mid Z)]/|\mathcal{A}|$. 
Smaller entropy means that the race probabilities are more concentrated toward $0$ or $1$, which indicates more predictive proxies.}
We find that, in terms of outcome, all proxies are equally uninformative \edit{(the entropy without using any proxy is around $0.5$)}. In terms of protected class, we find geolocation more informative than income and that combining them adds very little. 
\begin{figure}\centering
\includegraphics[width=\textwidth]{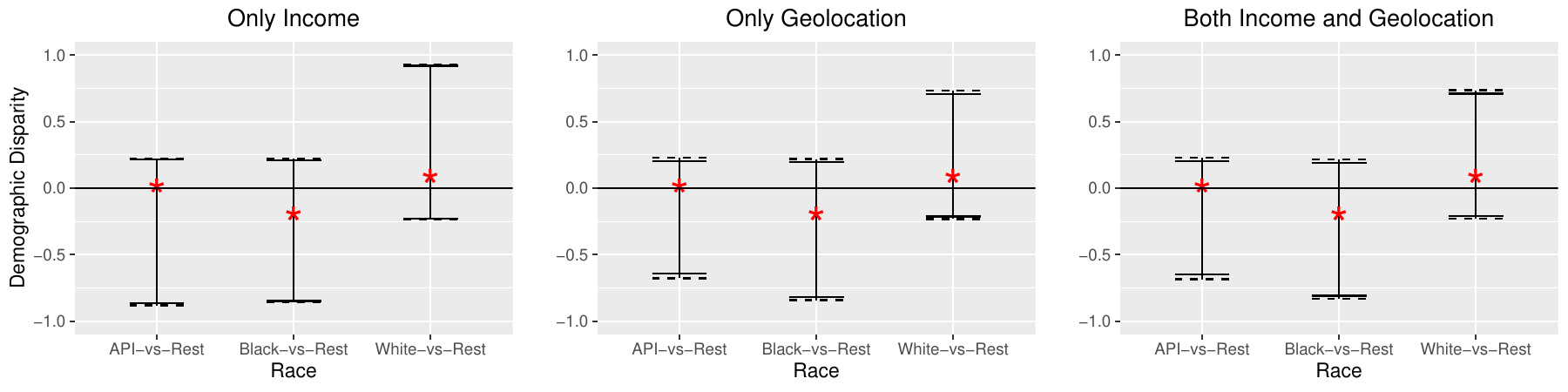}
\caption{Partial identification bounds of demographic disparity (\cref{prop: dd binary}) for different proxy variables in the HMDA dataset. Solid bars represent the estimates of the bounds, and dashed bars indicate $95\%$ confidence intervals. The true value based on self-reported race is shown as a red asterisk.}
\label{fig:hmda-CI}
\end{figure}
\edit{
\paragraph*{Binary comparisons.}
\Cref{fig:hmda-CI} demonstrates estimates of closed-form bounds of demographic disparities of one race versus the rest\footnote{For example, the White-vs-Rest disparity is the demographic disparity of $a$ as White and $b$ as either API or Black.} without any extra assumptions (\cref{prop: dd binary}), and also the associated confidence intervals. 
By recognizing that in case studies like the BISG proxy, the auxiliary dataset typically describes the whole population (e.g., the whole US population in decennial census data), we use an alternative estimator and confidence interval in \cref{sec: known-prob} that assumes the true conditional race probabilities (but not the conditional outcome probabilities) are exactly known from the auxiliary dataset.  
This figure also shows the true demographic disparity computed based on the self-reported race using the full data directly. 
We can observe that overall all estimated partial identification intervals are fairly wide, and all of them correctly contain the ground truth demographic disparity. 
Moreover, the finite-sample uncertainty of these estimates is quite small, and the confidence intervals show that at a $5\%$ significance level, we cannot reject zero as a valid disparity value according to the observed data information.}
	\begin{figure}%
		\centering
		\includegraphics[width=\textwidth]{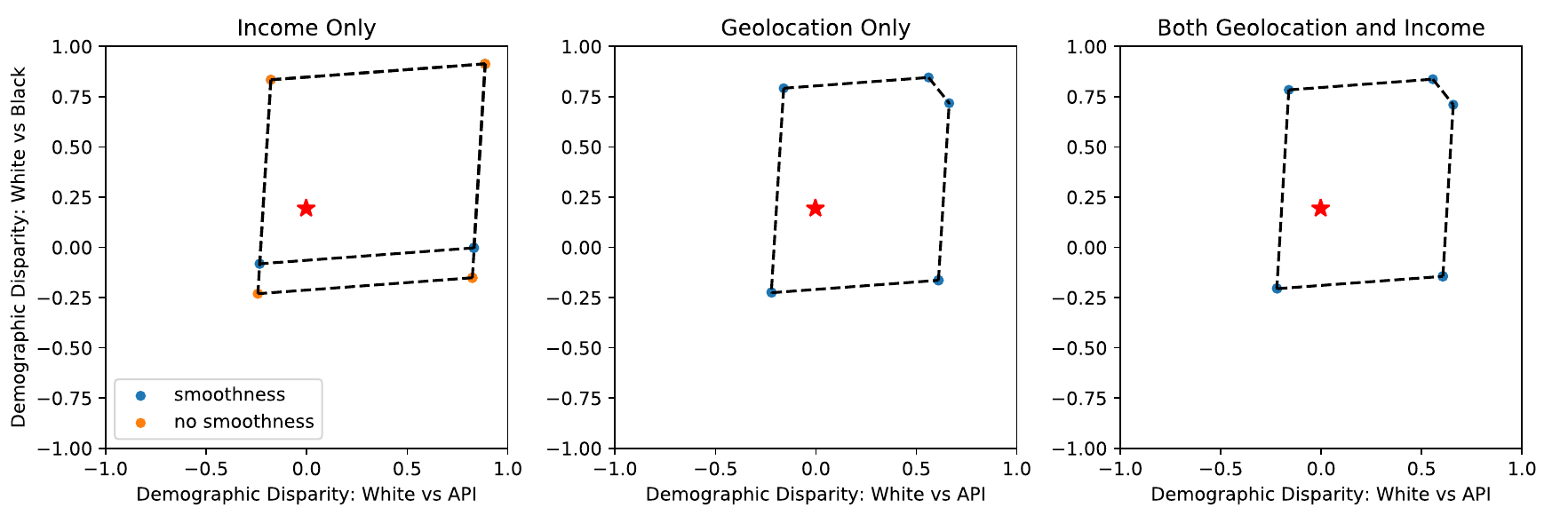}
		\caption{%
			The outer approximation of partial identification set for demographic disparity in loan approval rates in the HMDA dataset as determined by different proxies. Positive values correspond to disparity in favor of White. The true demographic disparity is shown as a red star.
		}
		\label{fig: pi-set-big}
	\end{figure}
\edit{\paragraph{Multiple-level protected class and extra smoothness assumption.}
\Cref{fig: pi-set-big} shows the estimated partial identification sets of the demographic disparities of White versus each other group. 
The sets are computed by the support function approach described in \cref{sec: inference-general}.
For the income-only proxy, we show the partial identification sets both without the smoothness constraint and with the smoothness constraint, where the Lipschitz constant is set as the minimal one such that the constraint set $\mw(\mathcal{P}_D) \cap \mw(\mathcal{P}_A)$ is still feasible.\footnote{Restricting the conditional joint distribution to be any smoother can in fact be refuted from the data via infeasibility.}  Smoothness constraints are implemented by enforcing the constraint of \cref{eq: lipcond dd} 
on the weight function, while the pairwise distance $d(z,z')$ can be computed efficiently for all observed values of the proxy variables.
}
The figure shows that using income as the only proxy, without additional smoothness constraints, seems quite weak in terms of identifying the demographic disparity. 
Income-only proxy without smoothness results in the largest partial identification set, and using income on top of geolocation barely shrinks the partial identification set relative to the set from using only the geolocation proxy.
Adding the smoothness constraint indeed shrinks the partial identification set of income-only proxy, and, given we are willing to assume smoothness, it shows that the White group either has a higher approval rate than the Black group or about roughly the same.
However, the magnitude of a positive White-vs-Black disparity, and the direction of White-vs-API disparity still remains very ambiguous.
\edit{
These observations are very likely to be valid not only for this sample but also for the whole population, given the small finite-sample uncertainty shown in \cref{fig:hmda-CI}. 
}
 
Overall the large size of all partial identification sets reflects the tremendous ambiguity in assessing lending disparities based on proxy variables like geolocation and income. 
Thus it is nearly \textit{impossible} to draw reliable conclusions about demographic disparity only according to the observed data. 
This conclusion is roughly in line with previous analyses of BISG \citep{chen2019fairness}, but provides a precise meaning to these limits.

\subsection{Personalized Warfarin Dosing}\label{sec: warfarin}
\begin{figure}\centering
\includegraphics[width=\textwidth]{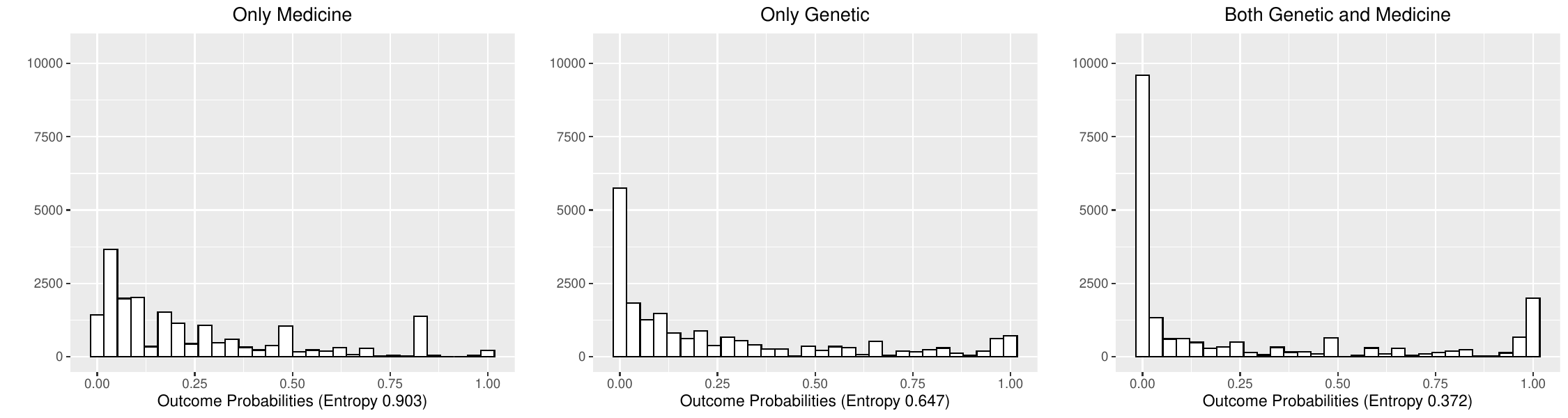}
\includegraphics[width=\textwidth]{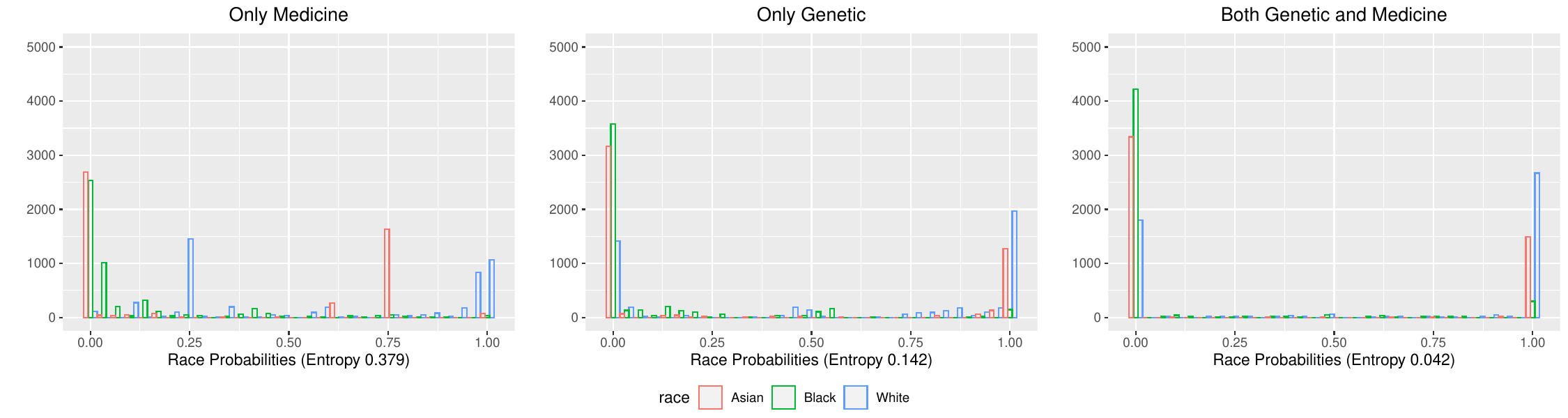}
\caption{Histograms of conditional probabilities of outcomes (upper row) and race (lower row) for different choices of proxies in the warfarin dataset, along with the resulting entropy.}
\label{fig: warfarin-proxy-outcome-hists}
\end{figure}

\paragraph{Background.} Warfarin is the most commonly used oral anticoagulant agent worldwide \citep{international2009estimation}.
Finding appropriate warfarin dosage is very challenging and important, since it can vary drastically among patients and incorrect dose can possibly lead to serious adverse outcomes. 
This challenge attracts considerable interest in designing personalized warfarin dosage algorithms, including linear regression \citep{international2009estimation}, LASSO \citep{bastani2015online}, and decision trees \citep{kallus2017recursive}.
However, it was shown that the personalized dosing algorithms may show disparate performance for different ethnic groups (see, e.g., Appendix 9 in \citealp{international2009estimation}). 

\paragraph*{Dataset, Proxy Variables, and Nuisance Estimation. } We use the PharmGKB dataset\footnote{The dataset can be downloaded from  \url{https://www.pharmgkb.org/downloads}.} of 5700 patients treated with warfarin. 
The data for each patient  includes demographics (sex, ethnicity, age, weight, height, and smoker), reason for treatment (e.g., atrial fibrillation), current medications, co-morbidities (e.g., diabetes), genetic factors (presence of genotype variants of CYP2C9 and VKORC1). 
All of these variables are categorical, and we treat missing value of each variable as a separate value. 
Moreover, this dataset contains the true patient-specific optimal warfarin doses determined by  physicians' adjustment over a few weeks. 
We focus on the subsample of 4891 White, Black, and Asian patients whose optimal warfarin doses are not missing.
We dichotomize the optimal doses into high dosage (more than $35$mg/week, denoted $Y = 1$), and low dosage (less than $35$mg/week, denoted $Y = 0$). 
To develop a personalized dosage algorithm, we follow \citet{international2009estimation} and fit a linear regression to predict the optimal dosage based on all other variables, and recommend high dosage if the predicted optimal dosage is  more than $35$mg/week ($\hY = 1$) and recommend low dosage ($\hY = 0$) otherwise.

We randomly split the dataset into two halves with one half as the primary dataset and the other as the auxiliary dataset, so that the independence of two datasets assumed in \cref{sec-inference} is satisfied.
Our goal is to evaluate the partial identification sets for true positive rate disparities
of this personalized dosage algorithm.
Positive disparities indicate that the personalized algorithm has higher chance to correctly recommend high dosage to one group than to another group.

We consider three sets of discrete proxy variables: only genetic factors, only current medications, and both genetic factors and current medications. 
Among the proxy variables, the genetic factors are particularly strong candidates since they are found to be highly predictive for the optimal  warfarin dosage \citep{international2009estimation}. At the same time, genotype variants of CYP2C9 and VKORC1 are known to also be highly correlated with race. For example, \citet{international2009estimation} even recommended imputing missing values of the genotypes based on race labels. 

The conditional probabilities of race, optimal dosage indicator $Y$, and recommended dosage indicator $\hY$ given these proxy variables can be easily estimated by corresponding sample averages within each level of the proxy variables. 
In \cref{fig: warfarin-proxy-outcome-hists} we display the histograms of the estimated conditional probabilities for both race and outcomes, for each proxy.
For outcomes, we show probabilities of all four combinations of true outcome and decision outcome. For race, we separate the probabilities by label.
We note that current medications and genetic factors together form a highly informative proxy, both for race and for outcomes.
\begin{figure}\centering
\includegraphics[width=\textwidth]{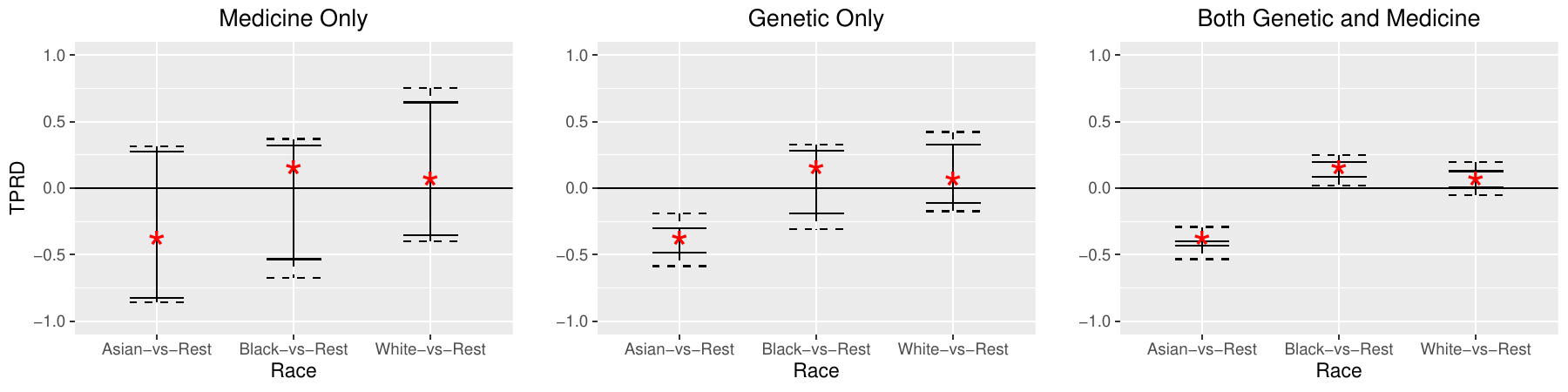}
\caption{Partial identification bounds of demographic disparity (\cref{prop: tprd binary}) for different proxy variables in the warfarin dosing. Solid bars represent the estimates of the bounds, and dashed bars indicate $95\%$ confidence intervals. The true value based on self-reported race is shown as a red asterisk.}
\label{fig: warfarin-CI}
\end{figure}
\edit{\paragraph*{Binary comparisons.} \Cref{fig: warfarin-CI} shows the estimates of closed-form bounds of TPRD for one race versus the rest without any extra assumptions. The bound estimators for TPRD and associated confidence intervals are similar to those for DD in \cref{sec: est-binary} (see \cref{proofsec: tprd-asymp} for details).
We first observe that using genetic factor proxies, whether in combination with current medication or not, provides clear evidence that the TPR disparity between Asian and other races is negative, in disfavor of Asians.
Although the directions of the Black-vs-Rest TPRD and White-vs-TPRD are unclear when using either genetic factor proxies or current medication proxies alone, combining these two set of proxies considerably narrows the bounds of these two disparities, and the Black-vs-Rest TPRD is positive at a 95\% confidence level. 
}
\begin{figure}
\includegraphics[width=0.32\textwidth]{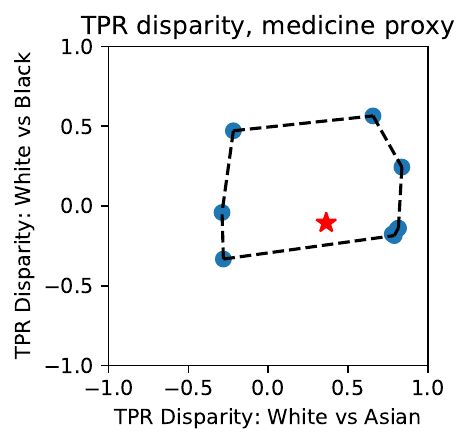}\;%
\includegraphics[width=0.32\textwidth]{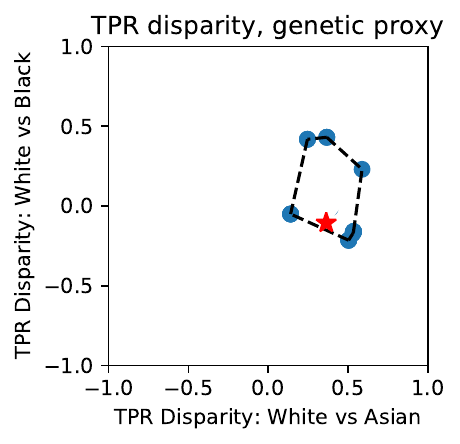}\;%
\includegraphics[width=0.32\textwidth]{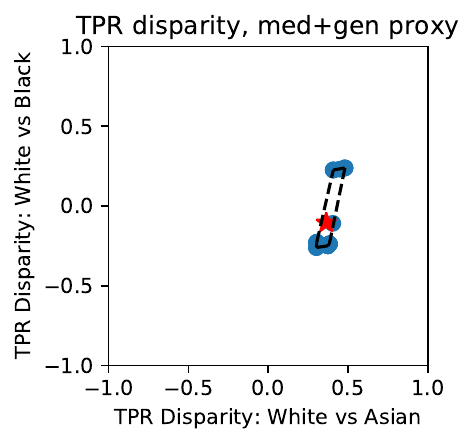}%
		\caption{The outer approximation of partial identification set for TPRD in warfarin dosing as determined by different proxies. The true disparity is shown as a red star.}
\label{fig:warfarin-suppfn}
\end{figure}
\edit{
\paragraph{Multiple-level protected class.}
\Cref{fig:warfarin-suppfn} shows the estimated partial identification sets of TPRD for White versus another group. 
The sets are computed by the support function approach described in \cref{sec: inference-general}.
We observe that using genetic factor proxies provides clear evidence that the TPRD between White and Asian is positive in favor of White. Further adding medication proxies provides a very clear sense of the significant magnitude of the TPRD between White and Asian, not just its direction.
However, in all cases, both the direction and magnitude of the disparity between White and Black is unclear.}

Overall our observations are consistent with the different quality of the proxies: while the genetic proxy is stronger than the medicine proxy, combining the proxies adds additional information that tightens the partially identified set. Studying the partially identified plots allows a practitioner to assess the value of additional information and, in some cases, the direction of disparities.

\section{Conclusion}

Assessing the fairness of algorithmic decisions is a fundamentally difficult task: it is now well-understood that even when algorithms do not take sensitive information as an input they can still be biased in various worrisome ways, but what counts as ``unfair'' can be very context-dependent. But any such adjudication and scrutiny must start from understanding how different groups are disparately impacted by such decisions. For example, disparate impact has been codified in US law and regulation as evidentiary basis for closer review and even sanction. We here studied a further complication: membership in protected groups is usually not even \edit{recorded} in the data, requiring the use of auxiliary data where such labels are present.
This limitation hinders both fair lending and healthcare reforms and it is important to address it.

We formulated this problem from the perspective of data combination and studied the fundamental limits of identification. This provided a new perspective on the commonplace usage of proxy models and a way to assess what can and cannot be learned from the data. The tools we developed allow one to compute exactly the tightest-possible bounds on disparity that could possibly be learned from the data. We believe this is an invaluable tool given that disparate impact assessments can have far-reaching policy implications.

Beyond the specific tools we presented here, we also hope our work will inspire other researchers to consider fundamental statistical ambiguities in the \edit{measurement} of fairness, beyond just the ambiguities between the different definitions. Given the sensitivity of such matters, truly understanding the limits of what cannot actually be measured, and what on the other hand can be said with certainty, is critical for any reliable assessment of the fairness of any decision-making algorithm.

\bibliographystyle{informs2014}
\setlength{\bibsep}{-3pt plus .3ex}
\bibliography{ecological}

\newpage
\begin{center}\Large
$ $\\
Online Appendix:\\
\vspace{8pt} Assessing the Fairness of Algorithmic Decisions with\vspace{8pt} Unobserved Protected Class Using Data Combination
\end{center}\vspace{8pt}

\begin{APPENDICES}
\crefalias{section}{appsec}
\crefalias{subsection}{appsec}
\crefalias{subsubsection}{appsec}

\edit{
\section{Inference for Partial Identification Sets with Binary Protected Class}\label{proofsec: TPRD-CI}
\subsection{Failure of Naive Plug-in Estimator}\label{proofsec: failure-naive}
In \cref{eq: est-eq-dd-1,eq: est-eq-dd-2}, we leverage extra augmentation terms $\expect\left[\xi^L_{\alpha}( A, Z; \eta)\right] + \expect\left[\gamma^L_{\alpha}(\hY, Z; \eta)\right]$ and $\expect\left[\xi^U_{\alpha}( A, Z; \eta)\right] + \expect\left[\gamma^U_{\alpha}(\hY, Z; \eta)\right]$ to ensure that the resulting bound estimators based on estimated values of $\eta$ have desirable asymptotic distribution. Now we demonstrate the problem of not adding these augmentation terms by using the estimation of $\mu(\alpha, w^L)$ as an example.   
}

\edit{
Consider the following estimator that only uses $\expect\left[\lambda^L_{\alpha}(Z; {\eta})\right]$:
\begin{align*}
\hat{\mu}(\alpha, w^L) 
=& \frac{1}{\hat{p}_{\alpha}K}\sum_{k = 1}^K\expectnk\left[\lambda^L_{\alpha}(Z; \hat{\eta}^{-k})\right] \\
=& \frac{1}{\hat{p}_{\alpha}K}\sum_{k = 1}^K\expectnk \left[\ind\left(\etapk(1, Z) + \etaak(\alpha, Z) - 1 \ge 0\right)\left(\etapk(1, Z) + \etaak(\alpha, Z) - 1\right)\right].
\end{align*}
Note that 
\begin{align}
&\sqrt{\npri}\left(\hat{\mu}(\alpha, w^L) - {\mu}(\alpha, w^L)\right)\nonumber \\
=& \frac{1}{{p}_{\alpha}K}\sum_{k = 1}^K\bigg\{\expectnk \left[\ind\left(\etapk(1, Z) + \etaak(\alpha, Z) - 1 \ge 0\right)\left(\etapk(1, Z) + \etaak(\alpha, Z) - 1\right)\right] \nonumber \\ 
&\qquad\qquad\qquad\qquad-  \expect\left[\ind\left(\etap(1, Z) + \etaa(\alpha, Z) - 1 \ge 0\right)\left(\etap(1, Z) + \etaa(\alpha, Z) - 1\right)\right]\bigg\} + o_p(1) \nonumber \\
=& \frac{1}{{p}_{\alpha}}\sqrt{\npri}\bigg\{\expectn\left[\ind\left(\etap(1, Z) + \etaa(\alpha, Z) - 1 \ge 0\right)\left(\etap(1, Z) + \etaa(\alpha, Z) - 1\right)\right] \nonumber  \\
&\qquad\qquad\qquad\qquad-   \expect\left[\ind\left(\etap(1, Z) + \etaa(\alpha, Z) - 1 \ge 0\right)\left(\etap(1, Z) + \etaa(\alpha, Z) - 1\right)\right]\bigg\} + o_p(1) \label{eq: mu-main-term} \\
+& \frac{1}{{p}_{\alpha}K}\sum_{k = 1}^K\sqrt{\npri}\bigg\{\expectnk \left[\ind\left(\etapk(1, Z) + \etaak(\alpha, Z) - 1 \ge 0\right)\left(\etapk(1, Z) + \etaak(\alpha, Z) - 1\right)\right] \nonumber \\
&\qquad\qquad\qquad\qquad-  \expectnk\left[\ind\left(\etap(1, Z) + \etaa(\alpha, Z) - 1 \ge 0\right)\left(\etap(1, Z) + \etaa(\alpha, Z) - 1\right)\right]\bigg\}. \label{eq: mu-error-term}
\end{align}
Here the main term (\ref{eq: mu-main-term}) is asymptotically normal according to Central Limit Theorem and Slutsky's theorem, and the asymptotic distribution of $\hat{\mu}(\alpha, w^L)$ also depends on the remainder term (\ref{eq: mu-error-term}). This remainder term can be decomposed as follows:
\begin{align}
(\ref{eq: mu-error-term})
	&= \frac{1}{{p}_{\alpha}K}\sum_{k = 1}^K\sqrt{\npri}\bigg\{\expectnk \big[\ind\left(\etapk(1, Z) + \etaak(\alpha, Z) - 1 \ge 0\right) \nonumber \\
	&\qquad\qquad\qquad\qquad \times \left(\etapk(1, Z) - \etap(1, Z) + \etaak(\alpha, Z) - \etaa(\alpha, Z)\right)\big]\bigg\} \label{eq: mu-error-term1}\\
	&-  \frac{1}{{p}_{\alpha}K}\sum_{k = 1}^K\sqrt{\npri}\bigg\{\expectnk\big[\left(\ind\left(\hat{\eta}^{-k}_{1}(Z) + \hat{\eta}^{-k}_{\alpha}(Z) - 1 \ge 0\right) - \ind\left(\etap(1, Z) + \etaa(\alpha, Z) - 1 \ge 0\right)\right) \nonumber \\
	&\qquad\qquad\qquad\qquad \times \left(\etap(1, Z) + \etaa(\alpha, Z) - 1\right)\big]\bigg\}. \label{eq: mu-error-term2}
\end{align}
By following the proof of \Cref{thm: asymp-dist-dd}, we can show that (\ref{eq: mu-error-term2})  is $o_p(1)$, and prove that under the conditions in \Cref{thm: asymp-dist-dd},
\begin{align*}
(\ref{eq: mu-error-term1}) 
	&= \frac{1}{{p}_{\alpha}K}\sum_{k = 1}^K \sqrt{n}_p\expect \big[\ind\left(\etapk(1, Z) + \etaak(\alpha, Z) - 1 \ge 0\right) \\
	&\qquad\qquad\qquad \times \left(\etapk(1, Z) - \etap(1, Z) + \etaak(\alpha, Z) - \etaa(\alpha, Z)\right) \mid \hat{\eta}^{-k}\big] + o_p(1)
\end{align*}
However, as $\npri \to \infty$, the display above diverges if $\etapk$ and $\etaak$,  are nonparametric estimators with convergence rates slower than $\npri^{-1/2}$. 
Consequently, the convergence rate of estimator $\hat{\mu}(\alpha, w^L)$ is also slower than $\npri^{-1/2}$. Moreover, even if strong parametric assumption is true such that $\etapk, \etaak$,  indeed converge to true values at rate $O(\npri^{-1/2})$, the asymptotic distribution of $(\ref{eq: mu-error-term1})$ is generally intractable if $Z$ is continuous. 
}

\edit{
In \cref{sec: est-binary}, we solve this problem by using extra augmentation terms. With these augmentation terms, the errors of estimating $\eta$ have only negligible impact on the downstream estimators for ${\mu}(\alpha, w^L), {\mu}(\alpha, w^U)$, so that only the main term  (\ref{eq: mu-main-term}) involving true $\eta$ matters. In particular, the asymptotic distributions of the final estimators are the same as those that use the true $\eta$ directly.
Consequently, the final estimators still converge at rate of $\npri^{-1/2}$, and they have well-beahved asymptotic distributions. 
}

\edit{
\subsection{Classification Disparity with Binary Protected Class}\label{proofsec: tprd-asymp}
In this section, we present the estimator and confidence interval for the closed-form partial identification sets of TPRD and TNRD given in \cref{prop: tprd binary}. 
}

\edit{
Note that the partial identification lower bound and upper bound both involve
\begin{align*}
\mu'_{\hy y}(\alpha; \tw,\tw') = \frac{\overline{w}_{\alpha}(\hy, y)}{\overline{w}_{\alpha}(\hy, y) + \overline{w}'_{\alpha}(1 - \hy, y)},
\end{align*}
where $\overline{w}_{\alpha}(\hy, y) = \expect\big[\bwuhyyz \indy\indhy\big]$ and $\overline{w}'_{\alpha}(\hy, y) = \expect\big[\tilde{w}'(\hY, Y, Z) \indy\indhy\big]$ with $\tilde{w}$ and $\tilde{w}'$ equal either  $\tilde{w}^L$ or $\tilde{w}^U$ defined in \cref{prop: tprd binary}.
}

\edit{
To estimate $\overline{w}^L_{\alpha}(\hy, y) = \expect\big[\tilde{w}_{\alpha}^L(\hY, Y, Z)\indy\indhy\big]$ and $\overline{w}^U_{\alpha}(\hy, y) = \expect\big[\tilde{w}_{\alpha}^U(\hY, Y, Z)\indy\indhy\big]$, we first consider the following reformulation:
\begin{align}
\overline{w}^L_{\alpha}(\hy, y) = \expect\left[\tilde{\lambda}^L_{\alpha, \hy y}(Z; \tilde\eta) \right] + \expect\left[\tilde{\xi}^L_{\alpha, \hy y}(A, Z; \tilde\eta)\right] + \expect\left[\tilde{\gamma}^L_{\alpha, \hy y}(\hY, Y, Z; \tilde\eta)\right],  \label{eq: class-disparity-1}\\
\overline{w}^U_{\alpha}(\hy, y) = \expect\left[\tilde{\lambda}^U_{\alpha, \hy y}(Z; \tilde\eta) \right] + \expect\left[\tilde{\xi}^U_{\alpha, \hy y}(A, Z; \tilde\eta)\right] + \expect\left[\tilde{\gamma}^U_{\alpha, \hy y}(\hY, Y, Z; \tilde\eta)\right], \label{eq: class-disparity-2}
\end{align}
where $\tilde\eta = (\etaa, \etapt)$, and  
\begin{align*}
&\tilde{\lambda}^L_{\alpha, \hy y}(z; \eta) 
	 = I^L_{\alpha, \hy y}(z)\left(\etapt(\hy, y, z) + \etaa(\alpha, z) - 1\right), ~~~ \tilde{\xi}^L_{\alpha, \hy y}(A, z; \eta) 
	\coloneqq I^L_{\alpha, \hy y}(z)(\ind(A = \alpha) - \etaa(\alpha, z)),  \\
&\tilde{\gamma}^L_{\alpha, \hy y}(\hY, Y, z; \eta) 
	\coloneqq I^L_{\alpha, \hy y}(z)(\ind(\hY = \hy, Y = y) - \etapt(\hy, y, z)), \text{ with } I^L_{\alpha, \hy y}(z) \coloneqq \ind\left(\etapt(\hy, y, z) + \etaa(\alpha, z) - 1 \ge 0 \right), \\
&\tilde{\lambda}^U_{\alpha, \hy y}(z; \eta) 
	= I^U_{\alpha, \hy y }(z)\left(\etapt(\hy, y, z) - \etaa(\alpha, z)\right)+ \etaa(\alpha, z), ~~~  \tilde{\xi}^U_{\alpha, \hy y}(A, z; \eta) \coloneqq (1 - I^U_{\alpha, \hy y}(z))(\ind(A = \alpha) - \etaa(\alpha, z)),\\
&\tilde{\gamma}^U_{\alpha, \hy y}(\hY, Y, z; \eta) \coloneqq I^U_{\alpha, \hy y}(z)(\ind(\hY = \hy, Y = y) - \etapt(\hy, y, z)),  \text{ with } I^U_{\alpha, \hy y}(z) \coloneqq \ind\left(\etapt(\hy, y, z) - \etaa(\alpha, z) \le 0\right).
\end{align*}
Based on \cref{eq: class-disparity-1,eq: class-disparity-2}, we propose the following estimators for $\overline{w}^L_{\alpha}(\hy, y)$ and $\overline{w}^U_{\alpha}(\hy, y)$ respectively. 
\begin{align*}
\hat{\overline{w}}^L_{\alpha}(\hy, y) 
	= \frac{1}{K}\sum_{k = 1}^K \left\{\expectnk\left[\tilde{\lambda}^L_{\alpha, \hy y}(Z; \hetat^{-k}) \right] + \expectnak\left[\tilde{\xi}^L_{\alpha, \hy y}(A, Z; \hetat^{-k})\right] + \expectnpk\left[\tilde{\gamma}^L_{\alpha, \hy y}(\hY, Y, Z; \hetat^{-k})\right]\right\},  \\
\hat{\overline{w}}^U_{\alpha}(\hy, y) 
	= \frac{1}{K}\sum_{k = 1}^K \left\{\expectnk\left[\tilde{\lambda}^U_{\alpha, \hy y}(Z; \hetat^{-k}) \right] + \expectnak\left[\tilde{\xi}^U_{\alpha, \hy y}(A, Z; \hetat^{-k})\right] + \expectnpk\left[\tilde{\gamma}^U_{\alpha, \hy y}(\hY, Y, Z; \hetat^{-k})\right]\right\},
\end{align*}
where $\hetat^{-k} = (\etaak, \etaptk)$ for $k = 1, \dots, K$ are  cross-fitting nuisance estimators analogous to $\heta^{-k} = (\etaak, \etapk)$ in \cref{alg-dml}.
The resulting plug-in estimators for the partial identification sets are 
\begin{align}
\hat{\Delta}_\TPRD(\tmw_\LTP)&=[\hat{\mu}'_{11}(a; \tw^L,\tw^U)-\hat{\mu}'_{11}(b; \tw^U,\tw^L),\ \hat{\mu}'_{11}(a; \tw^U,\tw^L)-\hat{\mu}'_{11}(b; \tw^L,\tw^U)], \label{eq: est-tprd}\\
\hat{\Delta}_\TNRD(\tmw_\LTP)&=[\hat{\mu}'_{00}(a; \tw^L,\tw^U)-\hat{\mu}'_{00}(b; \tw^U,\tw^L),\ \hat{\mu}'_{00}(a; \tw^U,\tw^L)-\hat{\mu}'_{00}(b; \tw^L,\tw^U)], \label{eq: est-tnrd}
\end{align}
where 
\begin{align*}
\hat{\mu}'_{\hy y}(\alpha; \tw^L,\tw^U)= \frac{\hat{\overline{w}}^L_{\alpha}(\hy, y)}{\hat{\overline{w}}^L_{\alpha}(\hy, y)  + \hat{\overline{w}}^U_{\alpha}(1 - \hy, y)}, ~~~ \hat{\mu}'_{\hy y}(\alpha; \tw^U,\tw^L)= \frac{\hat{\overline{w}}^U_{\alpha}(\hy, y)}{\hat{\overline{w}}^U_{\alpha}(\hy, y)  + \hat{\overline{w}}^L_{\alpha}(1 - \hy, y)}.
\end{align*}
}

\edit{
In the following theorem, we show that bound estimators in \cref{eq: est-tprd,eq: est-tnrd} are asymptotically normal with closed-form asymptotic variance. 
\begin{theorem}\label{thm: asymp-dist-tprd}
Suppose that the nuisance estimators converge at the following rate: for $k = 1, \dots, K$, $\yhyrange$, and $\alpha = a, b$,  
\[
	\left|\etaptk(\hy, y, Z) - \etapt(\hy, y, Z)\right| = O_p(\kappa_{\npri, \hY Y}), ~~ \left|\etaak(\alpha, Z) - \etaa(\alpha, Z)\right| = O_p(\kappa_{\naux, A}).
\]
Assume the following conditions: for $\alpha = a, b$ and $\yhyrange$
\begin{enumerate}[label=(\roman*)]
\item $\overline{w}_{\alpha}(\hy, y) > 0$;
\item there exists positive constants $m_1, m_2, c_1, c_2$ such that for any $p \ge 0$,
\[\pr\left(0 \le \left|\etapt(\hy, y, Z) + \etaa(\alpha, Z) - 1 \right| \le p\right) \le c_1 p^{m_1}, ~~~ \pr\left(0 \le \left|\etapt(\hy, y, Z) - \etaa(\alpha, Z) \right| \le p\right) \le c_2 p^{m_2};
\]
\item $\max\{\kappa_{\naux, A}, \kappa_{\npri, \hY Y}\}= o(\npri^{-1/(2+2m_1)})$,  $\max\{\kappa_{\naux, A}, \kappa_{\npri, \hY Y}\}= o(\npri^{-1/(2+2m_2)})$;
\item 
$|r - r_n|\kappa_{\npri, \hY Y} = o(\npri^{-1/2})$, $|r - r_n|\kappa_{\naux, A} = o(\npri^{-1/2})$.
\end{enumerate}
Then the upper bound and lower bound estimators $\hat{\mu}'_{\hy y}(a, \tilde{w}^L, \tilde{w}^U)  - \hat{\mu}'_{\hy y}(b, \tilde{w}^U, \tilde{w}^L)$ and $\hat{\mu}'_{\hy y}(a, \tilde{w}^U, \tilde{w}^L)  - \hat{\mu}'_{\hy y}(b, \tilde{w}^L, \tilde{w}^U)$ satisfy that as $\npri \to \infty$,
\begin{align*}
&\sqrt{\npri}\left[\hat{\mu}'_{\hy y}(a, \tilde{w}^L, \tilde{w}^U)  - \hat{\mu}'_{\hy y}(b, \tilde{w}^U, \tilde{w}^L) - \left({\mu}'_{\hy y}(a, \tilde{w}^L, \tilde{w}^U) - {\mu}'_{\hy y}(b, \tilde{w}^U, \tilde{w}^L\right)\right] \overset{d}{\to} \mathcal{N}(0, \tilde{V}_L(\hy, y)), \\
&\sqrt{\npri}\left[\hat{\mu}'_{\hy y}(a, \tilde{w}^U, \tilde{w}^L)  - \hat{\mu}'_{\hy y}(b, \tilde{w}^L, \tilde{w}^U) - \left({\mu}'_{\hy y}(a, \tilde{w}^U, \tilde{w}^L)  - {\mu}'_{\hy y}(b, \tilde{w}^L, \tilde{w}^U)\right)\right] \overset{d}{\to} \mathcal{N}(0, \tilde{V}_U(\hy, y)), 
\end{align*}
where 
\begin{align*}
\tilde{V}_L(\hy, y)
	&= r\expect\bigg\{\left[\frac{{\mu}'_{1 - \hy, y}(a; \tw^U,\tw^L)}{{\overline{w}}^L_{a}(\hy, y)  + {\overline{w}}^U_{a}(1 - \hy, y)}\tilde{\lambda}^L_{a, \hy y}(Z; \tilde\eta) - \frac{{\mu}'_{\hy y}(a; \tw^L,\tw^U)}{{\overline{w}}^L_{a}(\hy, y)  + {\overline{w}}^U_{a}(1 - \hy, y)}\tilde{\lambda}^U_{a, 1 - \hy, y}(Z; \tilde\eta)\right]\\
	&\qquad\qquad - \left[\frac{{\mu}'_{1 - \hy, y}(b; \tw^L,\tw^U)}{{\overline{w}}^U_{b}(\hy, y)  + {\overline{w}}^L_{b}(1 - \hy, y)}\tilde{\lambda}^U_{b, \hy y}(Z; \tilde\eta) - \frac{{\mu}'_{\hy y}(b; \tw^U,\tw^L)}{{\overline{w}}^U_{b}(\hy, y)  + {\overline{w}}^L_{b}(1 - \hy, y)}\tilde{\lambda}^L_{b, 1 - \hy, y}(Z; \tilde\eta)\right]\bigg\}^2 \\
	&+ \expect\bigg\{\frac{{\mu}'_{1 - \hy, y}(a; \tw^U,\tw^L)}{{\overline{w}}^L_{a}(\hy, y)  + {\overline{w}}^U_{a}(1 - \hy, y)}\tilde{\gamma}^L_{a, \hy y}(\hY, Y, Z; \tilde\eta) - \frac{{\mu}'_{\hy y}(a; \tw^L,\tw^U)}{{\overline{w}}^L_{a}(\hy, y)  + {\overline{w}}^U_{a}(1 - \hy, y)}\tilde{\gamma}^U_{a, 1 -\hy, y}(\hY, Y, Z; \tilde\eta)\\
	&\qquad\qquad - \left[\frac{{\mu}'_{1 - \hy, y}(b; \tw^L,\tw^U)}{{\overline{w}}^U_{b}(\hy, y)  + {\overline{w}}^L_{b}(1 - \hy, y)}\tilde{\gamma}^U_{b, \hy y}(\hY, Y, Z; \tilde\eta) - \frac{{\mu}'_{\hy y}(b; \tw^U,\tw^L)}{{\overline{w}}^U_{b}(\hy, y)  + {\overline{w}}^L_{b}(1 - \hy, y)}\tilde{\gamma}^L_{b, 1 -\hy, y}(\hY, Y, Z; \tilde\eta)\right]\bigg\}^2 \\ 
	&+ \frac{r}{1 - r}\expect\bigg\{\frac{{\mu}'_{1 - \hy, y}(a; \tw^U,\tw^L)}{{\overline{w}}^L_{a}(\hy, y)  + {\overline{w}}^U_{a}(1 - \hy, y)}\tilde{\xi}^L_{a, \hy y}(A, Z; \tilde\eta) - \frac{{\mu}'_{\hy y}(a; \tw^L,\tw^U)}{{\overline{w}}^L_{a}(\hy, y)  + {\overline{w}}^U_{a}(1 - \hy, y)}\tilde{\xi}^U_{a, 1 -\hy, y}(A, Z; \tilde\eta) \\
	&\qquad\qquad -\left[\frac{{\mu}'_{1 - \hy, y}(b; \tw^L,\tw^U)}{{\overline{w}}^U_{b}(\hy, y)  + {\overline{w}}^L_{b}(1 - \hy, y)}\tilde{\xi}^U_{b, \hy y}(A, Z; \tilde\eta) - \frac{{\mu}'_{\hy y}(b; \tw^U,\tw^L)}{{\overline{w}}^U_{b}(\hy, y)  + {\overline{w}}^L_{b}(1 - \hy, y)}\tilde{\xi}^L_{b, 1 -\hy, y}(A, Z; \tilde\eta)\right]\bigg\}^2. \\
\tilde{V}_U(\hy, y)
	&= r\expect\bigg\{\left[\frac{{\mu}'_{1 - \hy, y}(a; \tw^L,\tw^U)}{{\overline{w}}^U_{a}(\hy, y)  + {\overline{w}}^L_{a}(1 - \hy, y)}\tilde{\lambda}^U_{a, \hy y}(Z; \tilde\eta) - \frac{{\mu}'_{\hy y}(a; \tw^U,\tw^L)}{{\overline{w}}^U_{a}(\hy, y)  + {\overline{w}}^L_{a}(1 - \hy, y)}\tilde{\lambda}^L_{a, 1 - \hy, y}(Z; \tilde\eta)\right]\\
	&\qquad\qquad - \left[\frac{{\mu}'_{1 - \hy, y}(b; \tw^U,\tw^L)}{{\overline{w}}^L_{b}(\hy, y)  + {\overline{w}}^U_{b}(1 - \hy, y)}\tilde{\lambda}^L_{b, \hy y}(Z; \tilde\eta) - \frac{{\mu}'_{\hy y}(b; \tw^L,\tw^U)}{{\overline{w}}^L_{b}(\hy, y)  + {\overline{w}}^U_{b}(1 - \hy, y)}\tilde{\lambda}^U_{b, 1 - \hy, y}(Z; \tilde\eta)\right]\bigg\}^2 \\
	&+ \expect\bigg\{\frac{{\mu}'_{1 - \hy, y}(a; \tw^L,\tw^U)}{{\overline{w}}^U_{a}(\hy, y)  + {\overline{w}}^L_{a}(1 - \hy, y)}\tilde{\gamma}^U_{a, \hy y}(\hY, Y, Z; \tilde\eta) - \frac{{\mu}'_{\hy y}(a; \tw^U,\tw^L)}{{\overline{w}}^U_{a}(\hy, y)  + {\overline{w}}^L_{a}(1 - \hy, y)}\tilde{\gamma}^L_{a, 1 -\hy, y}(\hY, Y, Z; \tilde\eta)\\
	&\qquad\qquad - \left[\frac{{\mu}'_{1 - \hy, y}(b; \tw^U,\tw^L)}{{\overline{w}}^L_{b}(\hy, y)  + {\overline{w}}^U_{b}(1 - \hy, y)}\tilde{\gamma}^L_{b, \hy y}(\hY, Y, Z; \tilde\eta) - \frac{{\mu}'_{\hy y}(b; \tw^L,\tw^U)}{{\overline{w}}^L_{b}(\hy, y)  + {\overline{w}}^U_{b}(1 - \hy, y)}\tilde{\gamma}^U_{b, 1 -\hy, y}(\hY, Y, Z; \tilde\eta)\right]\bigg\}^2 \\ 
	&+ \frac{r}{1 - r}\expect\bigg\{\frac{{\mu}'_{1 - \hy, y}(a; \tw^L,\tw^U)}{{\overline{w}}^U_{a}(\hy, y)  + {\overline{w}}^L_{a}(1 - \hy, y)}\tilde{\xi}^U_{a, \hy y}(A, Z; \tilde\eta) - \frac{{\mu}'_{\hy y}(a; \tw^U,\tw^L)}{{\overline{w}}^U_{a}(\hy, y)  + {\overline{w}}^L_{a}(1 - \hy, y)}\tilde{\xi}^L_{a, 1 -\hy, y}(A, Z; \tilde\eta) \\
	&\qquad\qquad -\left[\frac{{\mu}'_{1 - \hy, y}(b; \tw^U,\tw^L)}{{\overline{w}}^L_{b}(\hy, y)  + {\overline{w}}^U_{b}(1 - \hy, y)}\tilde{\xi}^L_{b, \hy y}(A, Z; \tilde\eta) - \frac{{\mu}'_{\hy y}(b; \tw^L,\tw^U)}{{\overline{w}}^L_{b}(\hy, y)  + {\overline{w}}^U_{b}(1 - \hy, y)}\tilde{\xi}^U_{b, 1 -\hy, y}(A, Z; \tilde\eta)\right]\bigg\}^2.
\end{align*}
\end{theorem}
}

\edit{
Moreover, the closed-form asymptotic variances in \cref{thm: asymp-dist-tprd} suggest the following plug-in variance estimators:
\begin{align}\label{eq: var-est-tprd}
\hat{\tilde{V}}_L(\hy, y)
	&= \frac{r_n}{K}\sum_{k = 1}^K\expectnk\bigg\{\left[\frac{{\hat{\mu}}'_{1 - \hy, y}(a; \tw^U,\tw^L)}{{\hat{\overline{w}}}^L_{a}(\hy, y)  + {\hat{\overline{w}}}^U_{a}(1 - \hy, y)}\tilde{\lambda}^L_{a, \hy y}(Z; \hetat^{-k}) - \frac{{\hat{\mu}}'_{\hy y}(a; \tw^L,\tw^U)}{{\hat{\overline{w}}}^L_{a}(\hy, y)  + {\hat{\overline{w}}}^U_{a}(1 - \hy, y)}\tilde{\lambda}^U_{a, 1 - \hy, y}(Z; \hetat^{-k})\right] \nonumber \\
	&\qquad\qquad - \left[\frac{{\hat{\mu}}'_{1 - \hy, y}(b; \tw^L,\tw^U)}{{\hat{\overline{w}}}^U_{b}(\hy, y)  + {\hat{\overline{w}}}^L_{b}(1 - \hy, y)}\tilde{\lambda}^U_{b, \hy y}(Z; \hetat^{-k}) - \frac{{\hat{\mu}}'_{\hy y}(b; \tw^U,\tw^L)}{{\hat{\overline{w}}}^U_{b}(\hy, y)  + {\hat{\overline{w}}}^L_{b}(1 - \hy, y)}\tilde{\lambda}^L_{b, 1 - \hy, y}(Z; \hetat^{-k})\right]\bigg\}^2 \nonumber \\
	&+ \frac{1}{K}\sum_{k = 1}^K\expectnpk\bigg\{\frac{{\hat{\mu}}'_{1 - \hy, y}(a; \tw^U,\tw^L)}{{\hat{\overline{w}}}^L_{a}(\hy, y)  + {\hat{\overline{w}}}^U_{a}(1 - \hy, y)}\tilde{\gamma}^L_{a, \hy y}(\hY, Y, Z; \hetat^{-k}) - \frac{{\hat{\mu}}'_{\hy y}(a; \tw^L,\tw^U)}{{\hat{\overline{w}}}^L_{a}(\hy, y)  + {\hat{\overline{w}}}^U_{a}(1 - \hy, y)}\tilde{\gamma}^U_{a, 1 -\hy, y}(\hY, Y, Z; \hetat^{-k}) \nonumber \\
	&\qquad\qquad - \left[\frac{{\hat{\mu}}'_{1 - \hy, y}(b; \tw^L,\tw^U)}{{\hat{\overline{w}}}^U_{b}(\hy, y)  + {\hat{\overline{w}}}^L_{b}(1 - \hy, y)}\tilde{\gamma}^U_{b, \hy y}(\hY, Y, Z; \hetat^{-k}) - \frac{{\hat{\mu}}'_{\hy y}(b; \tw^U,\tw^L)}{{\hat{\overline{w}}}^U_{b}(\hy, y)  + {\hat{\overline{w}}}^L_{b}(1 - \hy, y)}\tilde{\gamma}^L_{b, 1 -\hy, y}(\hY, Y, Z; \hetat^{-k})\right]\bigg\}^2 \nonumber \\ 
	&+ \frac{r_n}{(1 - r_n)K}\sum_{k = 1}^K\expectnak\bigg\{\frac{{\hat{\mu}}'_{1 - \hy, y}(a; \tw^U,\tw^L)}{{\hat{\overline{w}}}^L_{a}(\hy, y)  + {\hat{\overline{w}}}^U_{a}(1 - \hy, y)}\tilde{\xi}^L_{a, \hy y}(A, Z; \hetat^{-k}) - \frac{{\hat{\mu}}'_{\hy y}(a; \tw^L,\tw^U)}{{\hat{\overline{w}}}^L_{a}(\hy, y)  + {\hat{\overline{w}}}^U_{a}(1 - \hy, y)}\tilde{\xi}^U_{a, 1 -\hy, y}(A, Z; \hetat^{-k}) \nonumber  \\
	&\qquad\qquad -\left[\frac{{\hat{\mu}}'_{1 - \hy, y}(b; \tw^L,\tw^U)}{{\hat{\overline{w}}}^U_{b}(\hy, y)  + {\hat{\overline{w}}}^L_{b}(1 - \hy, y)}\tilde{\xi}^U_{b, \hy y}(A, Z; \hetat^{-k}) - \frac{{\hat{\mu}}'_{\hy y}(b; \tw^U,\tw^L)}{{\hat{\overline{w}}}^U_{b}(\hy, y)  + {\hat{\overline{w}}}^L_{b}(1 - \hy, y)}\tilde{\xi}^L_{b, 1 -\hy, y}(A, Z; \hetat^{-k})\right]\bigg\}^2,  
\end{align}
and estimator $\hat{\tilde{V}}_U$ can be obtained by switching $L$ and $U$ above.
}

\edit{
We can show that these variance estimators are also consistent, and they can be used to construct confidence intervals for the partial identification sets of TPRD and TNRD. 
\begin{corollary}\label{corollary: CI-tprd}
Under the assumptions in \cref{thm: asymp-dist-tprd}, the asymptotic variance estimators are consistent: for $\yhyrange$, as $\npri \to \infty$,
\[
	\hat{\tilde{V}}_L(\hy, y) \overset{d}{\to} {\tilde{V}}_L(\hy, y), ~~~ 
	\hat{\tilde{V}}_U(\hy, y) \overset{d}{\to} {\tilde{V}}_U(\hy, y).
\]
We can construct the corresponding $(1 - \beta) \times 100\%$ confidence interval:
\begin{align*}
\operatorname{CI}(\hy, y) 
	&= \bigg[\hat{\mu}'_{\hy y}(a, \tilde{w}^L, \tilde{w}^U)  - \hat{\mu}'_{\hy y}(b, \tilde{w}^U, \tilde{w}^L) - \Phi^{-1}(1 - \beta/2){\hat{\tilde{V}}_L}^{1/2}(\hy, y)/\npri^{1/2}, \\
	&\qquad\qquad\qquad\qquad \hat{\mu}'_{\hy y}(a, \tilde{w}^U, \tilde{w}^L)  - \hat{\mu}'_{\hy y}(b, \tilde{w}^L, \tilde{w}^U) + \Phi^{-1}(1 - \beta/2){\hat{\tilde{V}}_U}^{1/2}(\hy, y)/\npri^{1/2}\bigg],
\end{align*}
where $\Phi^{-1}$ is the quantile function of standard normal distribution. This confidence interval satisfies the following: 
\begin{align*}
&\liminf_{\npri \to \infty} \pr\left(\Delta_\TPRD(\mathcal W_\LTP)\subseteq \operatorname{CI}(1, 1)\right) \ge 1 - \beta, ~~~ \liminf_{\npri \to \infty} \pr\left(\Delta_\TNRD(\mathcal W_\LTP)\subseteq \operatorname{CI}(0, 0)\right) \ge 1 - \beta.
\end{align*}
\end{corollary}
}

\edit{
Analogous results also hold for PPVD and NPVD by exchanging the role of $\hY$ and $Y$. 
}

\edit{
\subsection{Asymptotic Guarantee without Cross-fitting}
In \cref{sec: est-binary} and \cref{proofsec: tprd-asymp}, we derive the asymptotic distribution of bound estimators based on cross-fitting. This technique enforces independence between nuisance estimators and the data at which these nuisance estimators are evaluated. This enables us to show that the impact of nuisance estimation on the final asymptotic distribution is negligible, under only high-level rate conditions on the nuisance estimators. 
}

\edit{
In particular, in step II in the proof of \cref{thm: asymp-dist-dd}, we decompose the error of proposed partial identification bound estimators in \cref{sec: est-binary} into error terms $\mathcal{R}_1 \sim \mathcal{R}_7$. With the cross-fitting technique, these terms can be bounded by Markov inequality. Let us take $\mathcal{R}_4 - \mathcal{R}_5$ as an example: 
\begin{align*}
\mathcal{R}_4  - \mathcal{R}_5 
	&= (1 - r_n)\expectnak\left[\ind\left(\etapk(1, Z) + \etaak(\alpha, Z) - 1 \ge 0\right)\left(\etapk(1, Z) - \etap(1, Z)\right)\right] \\
	&\qquad\qquad\qquad\qquad - \frac{r_n(1 - r)}{r}\expectnpk\left[\ind\left(\etapk(1, Z) + \etaak(\alpha, Z) - 1 \ge 0\right)\left(\etapk(1, Z) - \etap(1, Z)\right)\right]
\end{align*}
We can show that conditionally on data not in the $k^\text{th}$ fold (so that $\etapk$ and $\etaak$ can be viewed as fixed), the expectation of $\mathcal{R}_4 - \mathcal{R}_5$ is $o_p(\npri^{-1/2})$ and the expectation of $(\mathcal{R}_4 - \mathcal{R}_5)^2$ is $o_p(1)$ under high-level convergence rate conditions for the nuisance estimators $\etapk$ and $\etaak$ given in \cref{thm: asymp-dist-dd}. Then by Markov inequality and \cref{lemma: Chernochukov}, we can show that $\mathcal{R}_4 -  \mathcal{R}_5= o_p(\npri^{-1/2})$ unconditionally. Analogously, we can bound other terms in $\mathcal{R}_1 \sim \mathcal{R}_7$ and prove that $\sum_{j = 1}^7\mathcal{R}_j = o_p(\npri^{-1/2})$. 
This means that using cross-fitting nuisance estimators rather than the true values of nuisance parameters do not affect the final asymptotic distribution of estimators given in \cref{sec: est-binary}.
Importantly, with cross-fitting, we can prove this without restricting the nuisance estimators beyond assuming the high-level convergence rate conditions in \cref{thm: asymp-dist-dd}. 
}

\edit{
In contrast, if we do not use cross-fitting, and instead we plug into \cref{eq: est-eq-dd-1,eq: est-eq-dd-2} the same nuisance estimator $\hat{\eta} = (\etaah, \etaph)$ trained on all data, then we can still decompose the error of resulting partial identification bound estimators into similar error terms, but we have to bound each error term in a different way. For example, the counterpart of $\mathcal{R}_4 - \mathcal{R}_5$ without cross-fitting is now 
\begin{align*}
\mathcal{R}_4(\hat{\eta}) - \mathcal{R}_5(\hat{\eta})  
	&= (1 - r_n)\expectna\left[\ind\left(\etaph(1, Z) + \etaah(\alpha, Z) - 1 \ge 0\right)\left(\etaph(1, Z) - \etap(1, Z)\right)\right] \\
	&\qquad\qquad\qquad\qquad - \frac{r_n(1 - r)}{r}\expectnp\left[\ind\left(\etaph(1, Z) + \etaah(\alpha, Z) - 1 \ge 0\right)\left(\etaph(1, Z) - \etap(1, Z)\right)\right].
\end{align*}
Here the nuisance estimator $\hat{\eta}$ can depend on the data where it is evaluated in an arbitrary way, so our previous analysis does no longer applies, and instead we need to use more involved empirical process emthods \citep[][Chap 19]{van2000asymptotic}. Suppose that the nuisance estimator $\hat{\eta}$ is picked from a function class $\mathcal{T}$ (either parametric class or nonparametric class). The empirical process approach requires us to prove the following uniform convergence:
\begin{align}\label{eq: EP-term}
\sup_{\hat{\eta} \in \mathcal{T}} ~ |\mathcal{R}_4(\hat{\eta}) - \mathcal{R}_5(\hat{\eta})| = o_p(\npri^{-1/2}). 
\end{align}
Besides similar convergence rate conditions given in \cref{thm: asymp-dist-dd}, (\ref{eq: EP-term}) typically requires $\mathcal{T}$ to be a Donsker class, i.e., a sufficiently simple function class whose metric entropy or bracketing entropy has limited growth rate \citep[][Chap 19]{van2000asymptotic}. 
For example, Donsker condition is satisfied for nuisance estimators based on smooth parametric class (e.g., the logistic regression used in \cref{sec: hmda}), finite-dimensional vector space (e.g., the sample frequency estimator for proxies with finitely many discrete levels in \cref{sec: warfarin,sec: hmda}), and more generally the VC class \citep[][Chap 19]{van2000asymptotic}. 
For these function classes, we should be able to verify (\ref{eq: EP-term}) and analogously bound all other error terms. 
As a result, the conclusion in \cref{thm: asymp-dist-dd} (and similarly \cref{thm: asymp-dist-tprd}) also holds for estimators without cross-fitting, if we additionally assume Donsker condition in nuisance estimation. 
Since our nuisance estimators in \cref{sec: hmda,sec: warfarin} are all simple enough to satisfy the Donsker condition, we do not need cross-fitting when estimating the closed-form bounds in \cref{sec: hmda,sec: warfarin}.
}

\edit{
Therefore, when not using cross-fitting, we need to assume the extra Donsker condition that modern machine learning estimators often violate. Also, rigorously establishing (\ref{eq: EP-term}) and bounding other error terms also involves complicated chaining arguments like those in \cite{athey2017efficient}. 
For the sake of both generality and simplicity, we focus on estimators based on cross-fitting in this paper. 
}

\edit{
\subsection{Calibrated Confidence Interval}\label{sec: cal-CI}
In \cref{thm: asymp-dist-dd} (and \cref{thm: asymp-dist-tprd}), we prove that the estimators for upper bound and lower bound of DD (and TPRD or TNRD) are asymptotically normal separately. 
By leveraging Bonferroni adjustment and consistent variance estimators, we then construct \textit{conservative} confidence intervals whose asymptotic coverage probability is \textit{no less than} the confidence level (\cref{corollary: CI-dd,corollary: CI-tprd}). 
In this section, we also incorporate the covariance between the upper bound and lower bound estimators to construct \textit{calibrated} confidence interval whose asymptotic covarage probability is \textit{exactly} the confidence level. 
}

\edit{
In the following corollary, we first strengthen \cref{thm: asymp-dist-dd} by showing that asymptotically the bound estimators for DD are actually jointly normal with closed-form covariance.   
\begin{corollary}\label{corollary: asymp-dist-dd-extended}
Under the assumptions in \cref{thm: asymp-dist-dd}, 
\begin{align*}
\sqrt{n}_p\begin{bmatrix}\hat{\mu}(a, w^L) - \hat{\mu}(b, w^U) - \left({\mu}(a, w^L) - {\mu}(b, w^U)\right) \\ \hat{\mu}(a, w^U) - \hat{\mu}(b, w^L) - \left({\mu}(a, w^U) - {\mu}(b, w^L)\right) \end{bmatrix} \overset{d}{\to} 
\mathcal{N}
\left(
\begin{bmatrix}
0 \\ 0
\end{bmatrix},
\begin{bmatrix}
V_L & \op{CV}_{LU} \\
\op{CV}_{LU} & V_U
\end{bmatrix}
\right), 
\end{align*}
where $V_L, V_U$ are given in \cref{thm: asymp-dist-dd}, and 
\begin{align*}
\op{CV}_{LU} &= r\expect\bigg\{\left[\lambda_{a}^L(Z; \eta)/p_a - \lambda_{b}^U(Z; \eta)/p_b - \left({\mu}(a, w^L) - {\mu}(b, w^U)\right)\right] \\
&\qquad\qquad\qquad\qquad\qquad \times \left[\lambda_{a}^L(Z; \eta)/p_a - \lambda_{b}^U(Z; \eta)/p_b - \left({\mu}(a, w^L) - {\mu}(b, w^U)\right)\right]\bigg\} \\
	&+ \frac{r}{1 - r}\expect\bigg\{\left[\xi_{a}^L(A, Z; \eta)/p_a - \xi_{b}^U(A, Z; \eta)/p_b\right]\left[\xi_{a}^U(A, Z; \eta)/p_a - \xi_{b}^L(A, Z; \eta)/p_b\right]\bigg\} \\
	&+ \expect\bigg\{\left[\gamma_{a}^L(A, Z; \eta)/p_a - \gamma_{b}^U(A, Z; \eta)/p_b\right]\left[\gamma_{a}^U(A, Z; \eta)/p_a - \gamma_{b}^L(A, Z; \eta)/p_b\right]\bigg\}.
\end{align*}
\end{corollary}
}

\edit{
As we show in \cref{corollary: CI-dd}, the variances $V_L, V_U$ can be consistently estimated by the plug-in estimators $\hat V_L, \hat V_U$ in \cref{eq: var-est}. Analogously, the covariance term can be consistently estimated by the following plug-in estimator:
\begin{align*}
\hat{\op{CV}}_{LU} 
	&= \frac{r_n}{K}\sum_{k = 1}^K\expectnk\bigg\{\left[\lambda_{a}^L(Z; \hat{\eta}^{-k})/\hat{p}_a - \lambda_{b}^U(Z; \hat{\eta}^{-k})/\hat{p}_b - \left(\hat{\mu}(a, w^L) - \hat{\mu}(b, w^U)\right)\right] \\
	&\qquad\qquad\qquad\qquad\qquad\qquad\qquad \times \left[\lambda_{a}^U(Z; \hat{\eta}^{-k})/\hat{p}_a - \lambda_{b}^L(Z; \hat{\eta}^{-k})/\hat{p}_b - \left(\hat{\mu}(a, w^U) - \hat{\mu}(b, w^L)\right)\right]\bigg\} \\
	&+ \frac{r_n}{(1 - r_n)K}\sum_{k = 1}^K\expectnak\bigg\{\left[\xi_{a}^L(A, Z; \hat{\eta}^{-k})/\hat{p}_a - \xi_{b}^U(A, Z; \hat{\eta}^{-k})/\hat{p}_b\right]\left[\xi_{a}^U(A, Z; \hat{\eta}^{-k})/\hat{p}_a - \xi_{b}^L(A, Z; \hat{\eta}^{-k})/\hat{p}_b\right]\bigg\} \\
	&+ \frac{1}{K}\sum_{k = 1}^K\expectnpk\bigg\{\left[\gamma_{a}^L(\hY, Z; \hat{\eta}^{-k})/\hat{p}_a - \gamma_{b}^U(\hY, Z; \hat{\eta}^{-k})/\hat{p}_b\right]\left[\gamma_{a}^U(\hY, Z; \hat{\eta}^{-k})/\hat{p}_a - \gamma_{b}^L(\hY, Z; \hat{\eta}^{-k})/\hat{p}_b\right]\bigg\}.
\end{align*}
}

\edit{
In the following corollary, we give the calibrated confidence interval for the partial identification set of DD, and prove that it is asymptotically valid. 
\begin{corollary}\label{corollary: cal-CI}
Consider the following $(1 - \beta) \times 100 \%$ confidence interval
\begin{align*}
CI' = &\bigg[\hat{\mu}(a, w^L) - \hat{\mu}(b, w^U) + \frac{\op{det}(\hat V^{1/2}) + \hat V_L - \hat{\op{CV}}_{LU}}{\sqrt{\hat V_L + \hat V_U + 2\op{det}(\hat V^{1/2})}}\frac{\Phi^{-1}(\sqrt{1 - \beta})}{\npri^{1/2}}, \\
&\qquad\qquad\qquad \hat{\mu}(a, w^U) - \hat{\mu}(b, w^L) + \frac{\op{det}(\hat V^{1/2}) + \hat V_U - \hat{\op{CV}}_{LU}}{\sqrt{\hat V_L + \hat V_U + 2\op{det}(\hat V^{1/2})}}\frac{\Phi^{-1}(\sqrt{1 - \beta})}{\npri^{1/2}}\bigg]
\end{align*}
where $\Phi^{-1}$ is the quantile function for standard normal distribution, and 
\begin{align*}
\hat{V} = 
\begin{bmatrix}
\hat{V}_L & \hat{\op{CV}}_{LU} \\
\hat{\op{CV}}_{LU} & \hat{V}_U
\end{bmatrix}.
\end{align*}
Under the assumptions in \cref{thm: asymp-dist-dd}, 
\begin{align*}
\pr(\Delta_{\DD}(\mathcal P_D) \subseteq CI') \to 1 - \beta, \text{ as } \npri \to \infty.
\end{align*}
\end{corollary}
}
\edit{
We can also prove a similar conclusions for TPRD and TNRD in the following two corollaries. 
\begin{corollary}\label{corollary: asymp-dist-tprd-extended}
Under the assumptions in \cref{thm: asymp-dist-tprd}, 
\begin{align*}
&~~~ \sqrt{n}_p\begin{bmatrix}\hat{\mu}'_{\hy y}(a, \tilde{w}^L, \tilde{w}^U)  - \hat{\mu}'_{\hy y}(b, \tilde{w}^U, \tilde{w}^L) - \left({\mu}'_{\hy y}(a, \tilde{w}^L, \tilde{w}^U) - {\mu}'_{\hy y}(b, \tilde{w}^U, \tilde{w}^L)\right) \\ 
\hat{\mu}'_{\hy y}(a, \tilde{w}^U, \tilde{w}^L)  - \hat{\mu}'_{\hy y}(b, \tilde{w}^L, \tilde{w}^U) - \left({\mu}'_{\hy y}(a, \tilde{w}^U, \tilde{w}^L) - {\mu}'_{\hy y}(b, \tilde{w}^L, \tilde{w}^U)\right)
 \end{bmatrix} \\
 &\overset{d}{\to} \mathcal{N}
\left(
\begin{bmatrix}
0 \\ 0
\end{bmatrix},
\begin{bmatrix}
\tilde{V}_L(\hy, y) & \tilde{\op{CV}}_{LU}(\hy, y) \\
\tilde{\op{CV}}_{LU}(\hy, y)  & \tilde{V}_U(\hy, y)
\end{bmatrix}
\right), 
\end{align*}
where $\tilde{V}_L(\hy, y)$ and $\tilde{V}_U(\hy, y)$ are given in \cref{thm: asymp-dist-tprd}, and 
\begin{align*}
&\qquad {\tilde{\op{CV}}}_{LU}(\hy, y) \\
&= {r}\expect
\bigg[\frac{{{\mu}}'_{1 - \hy, y}(a; \tw^U,\tw^L)}{{{\overline{w}}}^L_{a}(\hy, y)  + {{\overline{w}}}^U_{a}(1 - \hy, y)}\tilde{\lambda}^L_{a, \hy y}(Z; \tilde\eta) - \frac{{{\mu}}'_{\hy y}(a; \tw^L,\tw^U)}{{{\overline{w}}}^L_{a}(\hy, y)  + {{\overline{w}}}^U_{a}(1 - \hy, y)}\tilde{\lambda}^U_{a, 1 - \hy, y}(Z; \tilde\eta) \\
&\qquad\qquad\qquad - \frac{{{\mu}}'_{1 - \hy, y}(b; \tw^L,\tw^U)}{{{\overline{w}}}^U_{b}(\hy, y)  + {{\overline{w}}}^L_{b}(1 - \hy, y)}\tilde{\lambda}^U_{b, \hy y}(Z; \tilde\eta) + \frac{{{\mu}}'_{\hy y}(b; \tw^U,\tw^L)}{{{\overline{w}}}^U_{b}(\hy, y)  + {{\overline{w}}}^L_{b}(1 - \hy, y)}\tilde{\lambda}^L_{b, 1 - \hy, y}(Z; \tilde\eta)\bigg]\\
&\qquad\qquad \times \bigg[\frac{{{\mu}}'_{1 - \hy, y}(a; \tw^L,\tw^U)}{{{\overline{w}}}^U_{a}(\hy, y)  + {{\overline{w}}}^L_{a}(1 - \hy, y)}\tilde{\lambda}^U_{a, \hy y}(Z; \tilde\eta) - \frac{{{\mu}}'_{\hy y}(a; \tw^U,\tw^L)}{{{\overline{w}}}^U_{a}(\hy, y)  + {{\overline{w}}}^L_{a}(1 - \hy, y)}\tilde{\lambda}^L_{a, 1 - \hy, y}(Z; \tilde\eta) \\
&\qquad\qquad\qquad- \frac{{{\mu}}'_{1 - \hy, y}(b; \tw^U,\tw^L)}{{{\overline{w}}}^L_{b}(\hy, y)  + {{\overline{w}}}^U_{b}(1 - \hy, y)}\tilde{\lambda}^L_{b, \hy y}(Z; \tilde\eta) + \frac{{{\mu}}'_{\hy y}(b; \tw^L,\tw^U)}{{{\overline{w}}}^L_{b}(\hy, y)  + {{\overline{w}}}^U_{b}(1 - \hy, y)}\tilde{\lambda}^U_{b, 1 - \hy, y}(Z; \tilde\eta)\bigg] \\
&+\expect\bigg[\frac{{{\mu}}'_{1 - \hy, y}(a; \tw^U,\tw^L)}{{{\overline{w}}}^L_{a}(\hy, y)  + {{\overline{w}}}^U_{a}(1 - \hy, y)}\tilde{\gamma}^L_{a, \hy y}(\hY, Y, Z; \tilde\eta) - \frac{{{\mu}}'_{\hy y}(a; \tw^L,\tw^U)}{{{\overline{w}}}^L_{a}(\hy, y)  + {{\overline{w}}}^U_{a}(1 - \hy, y)}\tilde{\gamma}^U_{a, 1 -\hy, y}(\hY, Y, Z; \tilde\eta)  \\
&\qquad\qquad\quad\quad - \frac{{{\mu}}'_{1 - \hy, y}(b; \tw^L,\tw^U)}{{{\overline{w}}}^U_{b}(\hy, y)  + {{\overline{w}}}^L_{b}(1 - \hy, y)}\tilde{\gamma}^U_{b, \hy y}(\hY, Y, Z; \tilde\eta) + \frac{{{\mu}}'_{\hy y}(b; \tw^U,\tw^L)}{{{\overline{w}}}^U_{b}(\hy, y)  + {{\overline{w}}}^L_{b}(1 - \hy, y)}\tilde{\gamma}^L_{b, 1 -\hy, y}(\hY, Y, Z; \tilde\eta)\bigg] \\ 
&\qquad\qquad\quad \times  \bigg[\frac{{{\mu}}'_{1 - \hy, y}(a; \tw^L,\tw^U)}{{{\overline{w}}}^U_{a}(\hy, y)  + {{\overline{w}}}^L_{a}(1 - \hy, y)}\tilde{\gamma}^U_{a, \hy y}(\hY, Y, Z; \tilde\eta) - \frac{{{\mu}}'_{\hy y}(a; \tw^U,\tw^L)}{{{\overline{w}}}^U_{a}(\hy, y)  + {{\overline{w}}}^L_{a}(1 - \hy, y)}\tilde{\gamma}^L_{a, 1 -\hy, y}(\hY, Y, Z; \tilde\eta)  \\
&\qquad\qquad\quad\quad - \frac{{{\mu}}'_{1 - \hy, y}(b; \tw^U,\tw^L)}{{{\overline{w}}}^L_{b}(\hy, y)  + {{\overline{w}}}^U_{b}(1 - \hy, y)}\tilde{\gamma}^L_{b, \hy y}(\hY, Y, Z; \tilde\eta) + \frac{{{\mu}}'_{\hy y}(b; \tw^L,\tw^U)}{{{\overline{w}}}^L_{b}(\hy, y)  + {{\overline{w}}}^U_{b}(1 - \hy, y)}\tilde{\gamma}^U_{b, 1 -\hy, y}(\hY, Y, Z; \tilde\eta)\bigg] \\
&+ \frac{r}{(1 - r)}\expect\bigg[\frac{{{\mu}}'_{1 - \hy, y}(a; \tw^U,\tw^L)}{{{\overline{w}}}^L_{a}(\hy, y)  + {{\overline{w}}}^U_{a}(1 - \hy, y)}\tilde{\xi}^L_{a, \hy y}(A, Z; \tilde\eta) - \frac{{{\mu}}'_{\hy y}(a; \tw^L,\tw^U)}{{{\overline{w}}}^L_{a}(\hy, y)  + {{\overline{w}}}^U_{a}(1 - \hy, y)}\tilde{\xi}^U_{a, 1 -\hy, y}(A, Z; \tilde\eta)  \\
&\qquad\qquad\qquad\qquad -\frac{{{\mu}}'_{1 - \hy, y}(b; \tw^L,\tw^U)}{{{\overline{w}}}^U_{b}(\hy, y)  + {{\overline{w}}}^L_{b}(1 - \hy, y)}\tilde{\xi}^U_{b, \hy y}(A, Z; \tilde\eta) - \frac{{{\mu}}'_{\hy y}(b; \tw^U,\tw^L)}{{{\overline{w}}}^U_{b}(\hy, y)  + {{\overline{w}}}^L_{b}(1 - \hy, y)}\tilde{\xi}^L_{b, 1 -\hy, y}(A, Z; \tilde\eta)\bigg] \\
&\qquad\qquad\qquad \times \bigg[\frac{{{\mu}}'_{1 - \hy, y}(a; \tw^L,\tw^U)}{{{\overline{w}}}^U_{a}(\hy, y)  + {{\overline{w}}}^L_{a}(1 - \hy, y)}\tilde{\xi}^U_{a, \hy y}(A, Z; \tilde\eta) - \frac{{{\mu}}'_{\hy y}(a; \tw^U,\tw^L)}{{{\overline{w}}}^U_{a}(\hy, y)  + {{\overline{w}}}^L_{a}(1 - \hy, y)}\tilde{\xi}^L_{a, 1 -\hy, y}(A, Z; \tilde\eta)  \\
&\qquad\qquad\qquad\qquad -\frac{{{\mu}}'_{1 - \hy, y}(b; \tw^U,\tw^L)}{{{\overline{w}}}^L_{b}(\hy, y)  + {{\overline{w}}}^U_{b}(1 - \hy, y)}\tilde{\xi}^L_{b, \hy y}(A, Z; \tilde\eta) - \frac{{{\mu}}'_{\hy y}(b; \tw^L,\tw^U)}{{{\overline{w}}}^L_{b}(\hy, y)  + {{\overline{w}}}^U_{b}(1 - \hy, y)}\tilde{\xi}^U_{b, 1 -\hy, y}(A, Z; \tilde\eta)\bigg].
\end{align*}
\end{corollary}
}
\edit{
\begin{corollary}\label{corollary: cal-CI-tprd}
Consider the following $(1 - \beta) \times 100 \%$ confidence interval
\begin{align*}
CI'(\hy, y) = &\bigg[\hat{\mu}'_{\hy y}(a, \tilde{w}^L, \tilde{w}^U)  - \hat{\mu}'_{\hy y}(b, \tilde{w}^U, \tilde{w}^L) - \frac{\op{det}(\hat{\tilde{V}}^{1/2}(\hy, y)) + \hat{\tilde{V}}_L(\hy, y) - \hat{\tilde{\op{CV}}}_{LU}(\hy, y)}{\sqrt{\hat{\tilde{V}}_L(\hy, y) + \hat{\tilde{V}}_U(\hy, y) + 2\op{det}(\hat{\tilde{V}}^{1/2}(\hy, y))}}\frac{\Phi^{-1}(\sqrt{1 - \beta})}{\npri^{1/2}}, \\
&\qquad\qquad\qquad \hat{\mu}'_{\hy y}(a, \tilde{w}^U, \tilde{w}^L)  - \hat{\mu}'_{\hy y}(b, \tilde{w}^L, \tilde{w}^U) + \frac{\op{det}(\hat{\tilde{V}}^{1/2}(\hy, y)) + \hat{\tilde{V}}_U(\hy, y) - \hat{\tilde{\op{CV}}}_{LU}(\hy, y)}{\sqrt{\hat{\tilde{V}}_L(\hy, y) + \hat{\tilde{V}}_U(\hy, y) + 2\op{det}(\hat{\tilde{V}}^{1/2}(\hy, y))}}\frac{\Phi^{-1}(\sqrt{1 - \beta})}{\npri^{1/2}}\bigg],
\end{align*}
where $\Phi^{-1}$ is the quantile function for standard normal distribution, and 
\begin{align*}
\hat{\tilde{V}}(\hy, y) = 
\begin{bmatrix}
\hat{\tilde{V}}_L(\hy, y) & \hat{\tilde{\op{CV}}}_{LU}(\hy, y) \\
\hat{\tilde{\op{CV}}}_{LU}(\hy, y) & \hat{\tilde{V}}_U(\hy, y)
\end{bmatrix}
\end{align*}
with $\hat{\tilde{V}}_L(\hy, y), \hat{\tilde{V}}_U(\hy, y)$ are given in \cref{eq: var-est-tprd} and 
\begin{align*}
&\qquad \hat{\tilde{\op{CV}}}_{LU}(\hy, y) \\
&= \frac{r_n}{K}\sum_{k = 1}^K\expectnk
\bigg[\frac{{\hat{\mu}}'_{1 - \hy, y}(a; \tw^U,\tw^L)}{{\hat{\overline{w}}}^L_{a}(\hy, y)  + {\hat{\overline{w}}}^U_{a}(1 - \hy, y)}\tilde{\lambda}^L_{a, \hy y}(Z; \hetat^{-k}) - \frac{{\hat{\mu}}'_{\hy y}(a; \tw^L,\tw^U)}{{\hat{\overline{w}}}^L_{a}(\hy, y)  + {\hat{\overline{w}}}^U_{a}(1 - \hy, y)}\tilde{\lambda}^U_{a, 1 - \hy, y}(Z; \hetat^{-k}) \\
&\qquad\qquad\qquad - \frac{{\hat{\mu}}'_{1 - \hy, y}(b; \tw^L,\tw^U)}{{\hat{\overline{w}}}^U_{b}(\hy, y)  + {\hat{\overline{w}}}^L_{b}(1 - \hy, y)}\tilde{\lambda}^U_{b, \hy y}(Z; \hetat^{-k}) + \frac{{\hat{\mu}}'_{\hy y}(b; \tw^U,\tw^L)}{{\hat{\overline{w}}}^U_{b}(\hy, y)  + {\hat{\overline{w}}}^L_{b}(1 - \hy, y)}\tilde{\lambda}^L_{b, 1 - \hy, y}(Z; \hetat^{-k})\bigg]\\
&\qquad\qquad \times \bigg[\frac{{\hat{\mu}}'_{1 - \hy, y}(a; \tw^L,\tw^U)}{{\hat{\overline{w}}}^U_{a}(\hy, y)  + {\hat{\overline{w}}}^L_{a}(1 - \hy, y)}\tilde{\lambda}^U_{a, \hy y}(Z; \hetat^{-k}) - \frac{{\hat{\mu}}'_{\hy y}(a; \tw^U,\tw^L)}{{\hat{\overline{w}}}^U_{a}(\hy, y)  + {\hat{\overline{w}}}^L_{a}(1 - \hy, y)}\tilde{\lambda}^L_{a, 1 - \hy, y}(Z; \hetat^{-k}) \\
&\qquad\qquad\qquad- \frac{{\hat{\mu}}'_{1 - \hy, y}(b; \tw^U,\tw^L)}{{\hat{\overline{w}}}^L_{b}(\hy, y)  + {\hat{\overline{w}}}^U_{b}(1 - \hy, y)}\tilde{\lambda}^L_{b, \hy y}(Z; \hetat^{-k}) + \frac{{\hat{\mu}}'_{\hy y}(b; \tw^L,\tw^U)}{{\hat{\overline{w}}}^L_{b}(\hy, y)  + {\hat{\overline{w}}}^U_{b}(1 - \hy, y)}\tilde{\lambda}^U_{b, 1 - \hy, y}(Z; \hetat^{-k})\bigg] \\
&+\frac{1}{K}\sum_{k = 1}^K\expectnpk\bigg[\frac{{\hat{\mu}}'_{1 - \hy, y}(a; \tw^U,\tw^L)}{{\hat{\overline{w}}}^L_{a}(\hy, y)  + {\hat{\overline{w}}}^U_{a}(1 - \hy, y)}\tilde{\gamma}^L_{a, \hy y}(\hY, Y, Z; \hetat^{-k}) - \frac{{\hat{\mu}}'_{\hy y}(a; \tw^L,\tw^U)}{{\hat{\overline{w}}}^L_{a}(\hy, y)  + {\hat{\overline{w}}}^U_{a}(1 - \hy, y)}\tilde{\gamma}^U_{a, 1 -\hy, y}(\hY, Y, Z; \hetat^{-k})  \\
&\qquad\qquad\quad\quad - \frac{{\hat{\mu}}'_{1 - \hy, y}(b; \tw^L,\tw^U)}{{\hat{\overline{w}}}^U_{b}(\hy, y)  + {\hat{\overline{w}}}^L_{b}(1 - \hy, y)}\tilde{\gamma}^U_{b, \hy y}(\hY, Y, Z; \hetat^{-k}) + \frac{{\hat{\mu}}'_{\hy y}(b; \tw^U,\tw^L)}{{\hat{\overline{w}}}^U_{b}(\hy, y)  + {\hat{\overline{w}}}^L_{b}(1 - \hy, y)}\tilde{\gamma}^L_{b, 1 -\hy, y}(\hY, Y, Z; \hetat^{-k})\bigg] \\ 
&\qquad\qquad\quad \times  \bigg[\frac{{\hat{\mu}}'_{1 - \hy, y}(a; \tw^L,\tw^U)}{{\hat{\overline{w}}}^U_{a}(\hy, y)  + {\hat{\overline{w}}}^L_{a}(1 - \hy, y)}\tilde{\gamma}^U_{a, \hy y}(\hY, Y, Z; \hetat^{-k}) - \frac{{\hat{\mu}}'_{\hy y}(a; \tw^U,\tw^L)}{{\hat{\overline{w}}}^U_{a}(\hy, y)  + {\hat{\overline{w}}}^L_{a}(1 - \hy, y)}\tilde{\gamma}^L_{a, 1 -\hy, y}(\hY, Y, Z; \hetat^{-k})  \\
&\qquad\qquad\quad\quad - \frac{{\hat{\mu}}'_{1 - \hy, y}(b; \tw^U,\tw^L)}{{\hat{\overline{w}}}^L_{b}(\hy, y)  + {\hat{\overline{w}}}^U_{b}(1 - \hy, y)}\tilde{\gamma}^L_{b, \hy y}(\hY, Y, Z; \hetat^{-k}) + \frac{{\hat{\mu}}'_{\hy y}(b; \tw^L,\tw^U)}{{\hat{\overline{w}}}^L_{b}(\hy, y)  + {\hat{\overline{w}}}^U_{b}(1 - \hy, y)}\tilde{\gamma}^U_{b, 1 -\hy, y}(\hY, Y, Z; \hetat^{-k})\bigg] \\
&+ \frac{r_n}{(1 - r_n)K}\sum_{k = 1}^K\expectnak\bigg[\frac{{\hat{\mu}}'_{1 - \hy, y}(a; \tw^U,\tw^L)}{{\hat{\overline{w}}}^L_{a}(\hy, y)  + {\hat{\overline{w}}}^U_{a}(1 - \hy, y)}\tilde{\xi}^L_{a, \hy y}(A, Z; \hetat^{-k}) - \frac{{\hat{\mu}}'_{\hy y}(a; \tw^L,\tw^U)}{{\hat{\overline{w}}}^L_{a}(\hy, y)  + {\hat{\overline{w}}}^U_{a}(1 - \hy, y)}\tilde{\xi}^U_{a, 1 -\hy, y}(A, Z; \hetat^{-k})  \\
&\qquad\qquad\qquad\qquad -\frac{{\hat{\mu}}'_{1 - \hy, y}(b; \tw^L,\tw^U)}{{\hat{\overline{w}}}^U_{b}(\hy, y)  + {\hat{\overline{w}}}^L_{b}(1 - \hy, y)}\tilde{\xi}^U_{b, \hy y}(A, Z; \hetat^{-k}) - \frac{{\hat{\mu}}'_{\hy y}(b; \tw^U,\tw^L)}{{\hat{\overline{w}}}^U_{b}(\hy, y)  + {\hat{\overline{w}}}^L_{b}(1 - \hy, y)}\tilde{\xi}^L_{b, 1 -\hy, y}(A, Z; \hetat^{-k})\bigg] \\
&\qquad\qquad\qquad \times \bigg[\frac{{\hat{\mu}}'_{1 - \hy, y}(a; \tw^L,\tw^U)}{{\hat{\overline{w}}}^U_{a}(\hy, y)  + {\hat{\overline{w}}}^L_{a}(1 - \hy, y)}\tilde{\xi}^U_{a, \hy y}(A, Z; \hetat^{-k}) - \frac{{\hat{\mu}}'_{\hy y}(a; \tw^U,\tw^L)}{{\hat{\overline{w}}}^U_{a}(\hy, y)  + {\hat{\overline{w}}}^L_{a}(1 - \hy, y)}\tilde{\xi}^L_{a, 1 -\hy, y}(A, Z; \hetat^{-k})  \\
&\qquad\qquad\qquad\qquad -\frac{{\hat{\mu}}'_{1 - \hy, y}(b; \tw^U,\tw^L)}{{\hat{\overline{w}}}^L_{b}(\hy, y)  + {\hat{\overline{w}}}^U_{b}(1 - \hy, y)}\tilde{\xi}^L_{b, \hy y}(A, Z; \hetat^{-k}) - \frac{{\hat{\mu}}'_{\hy y}(b; \tw^L,\tw^U)}{{\hat{\overline{w}}}^L_{b}(\hy, y)  + {\hat{\overline{w}}}^U_{b}(1 - \hy, y)}\tilde{\xi}^U_{b, 1 -\hy, y}(A, Z; \hetat^{-k})\bigg].
\end{align*}
Under the assumptions in  \cref{thm: asymp-dist-tprd}, as $\npri \to \infty$,
\begin{align*}
\pr(\Delta_{\TPRD}(\mathcal P_D) \subseteq CI'(1, 1)) \to 1 - \beta, ~~ \pr(\Delta_{\TNRD}(\mathcal P_D) \subseteq CI'(0, 0)) \to 1 - \beta.
\end{align*}
\end{corollary}
}

\edit{
\subsection{Inference with Known Conditional Probabilities of Protected Class}\label{sec: known-prob}
All previous estimation and inference results assume two independent datasets. However, this assumption excludes datasets that possibly share some  overlapping units. For example, the decennial census data in BISG characterize the whole population, and thus should also include units in the primary dataset. 
To model this situation, we assume that the decennial census data  reveal the population distribution and do not have any finite-sample variability. In other words, we assume that the population conditional probability of protected class (race) given proxies (geolocation and surname) can be viewed as known.
The problem of combining datasets with general and unknown overlapping structure is beyond the scope of this paper.
}

\edit{
To formalize this working assumption, we suppose that we only observe individual-level data in the primary dataset $\{(\hY_i, Y_i, Z_i)\}_{i = 1}^{\npri}$, and we additionally know $\etaa(\alpha, z) = \pr(A = \alpha \mid Z = z)$ for $\alpha = a, b$ and $\zrange$ (e.g., from the decennial census data).
}

\edit{
We then construct the following estimator for demographic disparity:
\begin{align*}
\hat{\mu}(\alpha, w^L) = \frac{\expectnp\left[\ind\left(\etaph(1,Z) + \etaa(\alpha,Z) -1  \ge 0 \right)\left(\hY + \etaa(\alpha,Z) - 1\right)\right]}{\expectnp \left[\etaa(\alpha,Z)\right]}, \\
\hat{\mu}(\alpha, w^U) =  \frac{\expectnp\left[\ind\left(\etaph(1,Z) - \etaa(\alpha,Z)  \le 0 \right)\left(\hY - \etaa(\alpha,Z)\right) + \etaa(\alpha,Z)\right]}{\expectnp \left[\etaa(\alpha,Z)\right]}, 
\end{align*}
where $\etaph$ is an estimator for $\etap$ based on the entire primary dataset, and $\expectnp$ is the sample average operator on the primary dataset. 
}

\edit{
The corresponding partial identification set estimator is again 
\begin{align*}
\left[\hat{\mu}(a, w^L) - \hat{\mu}(b, w^U), ~~ \hat{\mu}(a, w^U) - \hat{\mu}(b, w^L)\right].
\end{align*}
We derive the asymptotic distribution of the upper bound and lower bound estimators in the following theorem. 
\begin{theorem}\label{thm: tprd-dd-known-dist}
Assume the following conditions:
\begin{enumerate}
\item $\left|\etaph(1, Z) - \etap(1, Z)\right| = O_p(\kappa_{\npri, \hY})$;
\item ${p}_{\alpha} > 0$ for $\arange$;
\item there exists positive constants $m_1, m_2, c_1, c_2$ such that for any $p \ge 0$,
\[\pr\left(0 \le \left|\etap(1, Z) + \etaa(\alpha, Z) - 1 \right| \le p\right) \le c_1 p^{m_1}, ~~~ \pr\left(0 \le \left|\etap(1, Z) - \etaa(\alpha, Z) \right| \le p\right) \le c_2 p^{m_2};
\]
\item $\max\{\kappa_{\naux, A}, \kappa_{\npri, \hY Y}\}= o(\npri^{-1/(2+2m_1)})$,  $\max\{\kappa_{\naux, A}, \kappa_{\npri, \hY Y}\}= o(\npri^{-1/(2+2m_2)})$.
\end{enumerate}
Then as $\npri \to \infty$, the lower bound and upper bound estimators for demographic disparity with binary protected class are asymptotically normal:
\begin{align*}
\sqrt{\npri}\left[\left(\hat{\mu}(a, w^L) - \hat{\mu}(b, w^U)\right) - \left({\mu}(a, w^L) - {\mu}(b, w^U)\right)\right] \overset{d}{\to} \mathcal{N}(0, V_L) \\
\sqrt{\npri}\left[\left(\hat{\mu}(a, w^U) - \hat{\mu}(b, w^L)\right) - \left({\mu}(a, w^U) - {\mu}(b, w^L)\right)\right] \overset{d}{\to} \mathcal{N}(0, V_U) 
\end{align*}
where 
\begin{align*}
V_L 
	=& \expect\bigg[\ind\left(\etap(1, Z) + \etaa(a, Z) -1  \ge 0 \right)\left(\hY + \etaa(a, Z) - 1\right)/\expect\left[\etaa(a, Z)\right] \\
-&\qquad\qquad \left[\ind\left(\etap(1, Z) - \etaa(b, Z)  \le 0 \right)\left(\hY - \etaa(b, Z)\right) + \etaa(b, Z)\right]/\expect\left[\etaa(b, Z)\right] - \left({\mu}(a, w^L)  - {\mu}(b, w^U)\right)\bigg]^2. \\
V_U 
	=& \expect\bigg[\left[\ind\left(\etap(1, Z) - \etaa(a, Z)  \le 0 \right)\left(\hY - \etaa(a, Z) \right) + \eta_{\alpha}(Z)\right]/\expect\left[\etaa(a, Z)\right] \\
	-&\qquad\qquad\qquad \ind\left(\etap(1, Z) + \etaa(b, Z) - 1  \ge 0 \right)\left(\hY + \etaa(b, Z) - 1\right)/\expect\left[\etaa(b, Z)\right] - \left({\mu}(a, w^U)  - {\mu}(b, w^L)\right)\bigg]^2. \\
\end{align*}
\end{theorem}
The closed-form variances in \cref{thm: tprd-dd-known-dist} suggest the following plug-in variance estimators:
\begin{align*}
\hat{V}_L  =& \expect\bigg[\ind\left(\etaph(1, Z) + \etaa(a, Z) -1  \ge 0 \right)\left(\hY + \etaa(a, Z) - 1\right)/\expectnp\left[\etaa(a, Z)\right] \\
&\qquad\qquad-\left[\ind\left(\etaph(1, Z) - \etaa(b, Z)  \le 0 \right)\left(\hY - \etaa(b, Z)\right) + \etaa(b, Z)\right]/\expectnp\left[\etaa(b, Z)\right] - \left({\mu}(a, w^L)  - {\mu}(b, w^U)\right)\bigg]^2. \\
\hat{V}_U 
	=& \expect\bigg[\left[\ind\left(\etaph(1, Z) - \etaa(a, Z)  \le 0 \right)\left(\hY - \etaa(a, Z) \right) + \eta_{\alpha}(Z)\right]/\expectnp\left[\etaa(a, Z)\right] \\
	&\qquad\qquad\qquad-\ind\left(\etaph(1, Z) + \etaa(b, Z) - 1  \ge 0 \right)\left(\hY + \etaa(b, Z) - 1\right)/\expectnp\left[\etaa(b, Z)\right] - \left({\mu}(a, w^U)  - {\mu}(b, w^L)\right)\bigg]^2. \\
\end{align*}
We can analogously prove that these variance estimators are consistent under the conditions in \cref{thm: tprd-dd-known-dist},  and they can be used to construct confidence intervals shown in \cref{corollary: CI-dd}.
}

\section{Omitted Proofs}
\subsection{Proof of \cref{prop: minimal-unid}}\label{proofsec: prop: minimal-unid}
\edit{
\proof{Proof.}
For binary protected group, for $\yhyrange$
    \begin{align}
        \pr(A = b, \hY = \hy \mid Z = z) &= \pr(\hY = \hy \mid Z = z)  - \pr(A = a, \hY = \hy \mid Z = z)  \label{eq: ltp-app-1} \\
        \pr(A = b, \hY = \hy, Y = y \mid Z = z) &= \pr(\hY = \hy, Y = y \mid Z = z)  - \pr(A = a, \hY = \hy, Y = y \mid Z = z) \label{eq: ltp-app-2}
    \end{align}
\textbf{Demographic disparity.} We first illustrate the unidentifiability of demographic disparity.  \cref{eq: ltp-app-1}implies that we can reformulate the demographic disparity as follows:
\begin{align}
\delta_\DD(a, b) 
    &= \frac{\int \pr(A = a, \hY = 1 \mid Z = z)d\pr(z)}{\pr(A = a)} -  \frac{\int \pr(A = b, \hY = 1 \mid Z = z)d\pr(z)}{\pr(A = b)} \nonumber \\
    &= \frac{\int \pr(A = a, \hY = 1 \mid Z = z)d\pr(z)}{\pr(A = a)} -  \frac{\int \big(\pr( \hY = 1 \mid Z = z) - \pr(A = b, \hY = 1 \mid Z = z)\big)d\pr(z)}{\pr(A = b)} \nonumber \\
    &= \bigg(\frac{1}{\pr(A = a)} + \frac{1}{\pr(A = b)}\bigg)\int \pr(A = a, \hY = 1 \mid Z = z)d\pr(z) - \frac{\pr(\hY = 1)}{\pr(A = b)}. \label{eq: DD-new-form}
\end{align}
This formulation means that $\delta_\DD(a, b)$ is a bijective map of $\int \pr(A = a, \hY = 1 \mid Z = z)d\pr(z)$. We will construct two valid distributions $\tilde{\pr}_1$ and $\tilde{\pr}_2$ such that $\int \tilde{\pr}_1(A = a, \hY = 1 \mid Z = z)d\pr(z) \ne \int \tilde{\pr}_2(A = a, \hY = 1 \mid Z = z)d\pr(z)$. As a result, $\delta_\DD(a, b)$ induced by these two distributions are different. 
}

\edit{
Given the asserted assumption in \Cref{prop: minimal-unid}(i), without loss of generality, we can assume that there exists a set $\mathcal{Z}_0$ with $\pr(Z \in \mathcal{Z}_0) > 0$ such that for $z \in \mathcal{Z}_0$, $0< \pr(A = a \mid Z = z) < 1$ and $0 < \pr(\hY = 1 \mid z) < 1$. 
Under this condition, the Fr\'echet Hoeffding inequality endpoints for $\pr(A = a, \hY = \hy \mid Z = z)$ satisfy that for any $z \in \mathcal{Z}_0$,
\begin{align}\label{eq: dd-fh-app}
    \max\bigg\{\pr(A = a \mid Z = z) + \pr(\hY = 1 \mid Z = z) - 1, 0\bigg\}  < \min\bigg\{\pr(A = a \mid Z = z), \pr(\hY = 1 \mid Z = z)\bigg\}
\end{align}
It is straightforward to verify that 
\begin{align*}
\pr(A = a \mid Z = z)\pr(\hY = 1 \mid Z = z) &< \min\bigg\{\pr(A = a \mid Z = z), \pr(\hY = 1 \mid Z = z)\bigg\}, \\
\pr(A = a \mid Z = z)\pr(\hY = 1 \mid Z = z) &> \max\bigg\{\pr(A = a \mid Z = z) + \pr(\hY = 1 \mid Z = z) - 1, 0\bigg\}.
\end{align*}
Assuming $\mathcal Z_0^c \ne \emptyset$: for $z \in \mathcal Z_0^c$, either $\pr(A = a \mid Z = z) \in \{0, 1\}$ or $\pr(\hY = 1 \mid Z = z) \in \{0, 1\}$. In particular, if at least one of them is $0$, then 
\begin{align*}
    &\max\bigg\{\pr(A = a \mid Z = z) + \pr(\hY = 1 \mid Z = z) - 1, 0\bigg\}  = \min\bigg\{\pr(A = a \mid Z = z), \pr(\hY = 1 \mid Z = z)\bigg\} \\
    =&  \pr(A = a \mid Z = z)\pr(\hY = 1 \mid Z = z) = 0.
\end{align*}
If both of them are $1$, we can also verify that 
\begin{align*}
    &\max\bigg\{\pr(A = a \mid Z = z) + \pr(\hY = 1 \mid Z = z) - 1, 0\bigg\}  = \min\bigg\{\pr(A = a \mid Z = z), \pr(\hY = 1 \mid Z = z)\bigg\} \\
    =&  \pr(A = a \mid Z = z)\pr(\hY = 1 \mid Z = z).
\end{align*}
All of these show that $\tilde{\pr}_1$ that satisfies conditional independence is always a valid full joint distribution that agrees with the given marginals. Namely, $\tilde{\pr}_1$ satisfies that for any $z \in \mathcal{Z}$,
\[
	\tilde{\pr}_1(A = a, \hY = 1 \mid Z = z) = \pr(A = a \mid Z = z)\pr(\hY = 1 \mid Z = z).
\]
Moreover, there exists a constant $\epsilon > 0$ and a subset $\tilde{\mathcal{Z}}_0 \subseteq \mathcal{Z}_0$ such that $\pr(Z \in \tilde{\mathcal{Z}}_0) > 0$ and for any $z \in \tilde{\mathcal{Z}}_0$
\begin{align}\label{eq: exist-eps}
\min\bigg\{\pr(A = a \mid Z = z), \pr(\hY = 1 \mid Z = z)\bigg\} - \pr(A = a \mid Z = z)\pr(\hY = 1 \mid Z = z) > \epsilon.
\end{align}
This is trivially true if $|\mathcal{Z}_0|$ is finite. If $|\mathcal{Z}_0|$ is infinite, and this is not true, then for any constant $\epsilon' > 0$ and any subset $\mathcal{Z}'_0 \subseteq \mathcal{Z}_0$ with positive measure, for any $z \in \mathcal{Z}_0'$, we must have 
\[
	 0 < \min\bigg\{\pr(A = a \mid Z = z), \pr(\hY = 1 \mid Z = z)\bigg\} - \pr(A = a \mid Z = z)\pr(\hY = 1 \mid Z = z) \le \epsilon'.
\]
We can simply take $\mathcal{Z}'_0 = \mathcal{Z}_0$, and send $\epsilon' \to 0$, then the inequality above shows clear contradiction. Therefore, \cref{eq: exist-eps} has to be true for a certain constant $\epsilon > 0$ and a subset $\tilde{\mathcal{Z}}_0 \subseteq \mathcal{Z}_0$ such that $\pr(Z \in \tilde{\mathcal{Z}}_0) > 0$. 
}

\edit{
For this $\epsilon > 0$, we can also find a valid joint distribution  $\tilde{\pr}_2$ that is comptaible with the given marginals and satisfies the following condition: for $\arange, \hyrange$,
\begin{align*}
\tilde{\pr}_2(A = \alpha, \hY = \hy \mid Z = z) 
	&= \tilde{\pr}_1(A = \alpha, \hY = \hy \mid Z = z)~~ \text{ for } ~~z \in \tilde{\mathcal{Z}}_0^c,  \\
\tilde{\pr}_2(A = a, \hY = 1 \mid Z = z) 
	&= \tilde{\pr}_1(A = a, \hY = 1 \mid Z = z)  + \epsilon ~~ \text{ for } ~~ z \in \tilde{\mathcal{Z}}_0,
\end{align*}
as long as we choose  other components $\tilde{\pr}_1(A = b, \hY = 1 \mid Z = z), \tilde{\pr}_1(A = b, \hY = 0 \mid Z = z), \tilde{\pr}_1(A = a, \hY = 0 \mid Z = z)$ accordingly to satisfy the law of total probability.
}

\edit{
Obviously 
\begin{align*}
\int \tilde{\pr}_2(A =a, \hY = 1 \mid Z = z)d\pr(z)  
&= \int_{\tilde{\mathcal{Z}}_0} \tilde{\pr}_2(A =a, \hY = 1 \mid Z = z)d\pr(z)  + \int_{\tilde{\mathcal{Z}}_0^c} \tilde{\pr}_2(A =a, \hY = 1 \mid Z = z)d\pr(z) \\
&= \int_{\tilde{\mathcal{Z}}_0} \tilde{\pr}_1(A =a, \hY = 1 \mid Z = z)d\pr(z)  + \int_{\tilde{\mathcal{Z}}_0^c} \tilde{\pr}_1(A =a, \hY = 1 \mid Z = z)d\pr(z)  + \epsilon \\
& > \int \tilde{\pr}_1(A =a, \hY = 1 \mid Z = z)d\pr(z).
\end{align*}
\textbf{True positive rate disparity.} Now we prove the unidentifiability for true positive rate disparity, and the conclusion for true negative rate disparity can be proved analogously. We start with a reformulation of $\delta_\TPRD(a, b)$:
\begin{align*}
\delta_\TPRD(a, b)  
&= \frac{\int \pr(A = a, \hY = 1, Y = 1 \mid Z = z)d\pr(z)}{\int \pr(A = a, \hY = 1, Y = 1 \mid Z = z)d\pr(z) + \int \pr(A = a, \hY = 0, Y = 1 \mid Z = z)d\pr(z)} \\
&- \frac{\int \pr(A = b, \hY = 1, Y = 1 \mid Z = z)d\pr(z)}{\int \pr(A = b, \hY = 1, Y = 1 \mid Z = z)d\pr(z) + \int \pr(A = b, \hY = 0, Y = 1 \mid Z = z)d\pr(z)}  \\
&= \frac{\int \pr(A = a, \hY = 1, Y = 1 \mid Z = z)d\pr(z)}{\int \pr(A = a, \hY = 1, Y = 1 \mid Z = z)d\pr(z) + \int \pr(A = a, \hY = 0, Y = 1 \mid Z = z)d\pr(z)} \\
&- \frac{\pr(\hY = 1, Y = 1) - \int \pr(A = a, \hY = 1, Y = 1 \mid Z = z)d\pr(z)}{\pr(Y = 1) - \int \pr(A = a, \hY = 1, Y = 1 \mid Z = z)d\pr(z) - \int \pr(A = a, \hY = 0, Y = 1 \mid Z = z)d\pr(z)},
\end{align*} 
}

\edit{
Here $\delta_\TPRD(a, b)$  depends on both $\int \pr(A = a, \hY = 0, Y = 1 \mid Z = z)d\pr(z)$ and $\int \pr(A = a, \hY = 1, Y = 1 \mid Z = z)d\pr(z)$. However, if we can fix $\int \pr(A = a, \hY = 0, Y = 1 \mid Z = z)d\pr(z) + \int \pr(A = a, \hY = 1, Y = 1 \mid Z = z)d\pr(z)$ while varying $\int \pr(A = a, \hY = 1, Y = 1 \mid Z = z)d\pr(z)$, then $\delta_\TPRD(a, b)$ is also a bijective map of $\int \pr(A = a, \hY = 1, Y = 1 \mid Z = z)d\pr(z)$. We will construct two valid distributions $\tilde{\pr}_1$ and $\tilde{\pr}_2$ such that 
\begin{align*}
    &\qquad\int \tilde{\pr}_1(A = a, \hY = 0, Y = 1 \mid Z = z)d\pr(z) + \int \tilde{\pr}_1(A = a, \hY = 1, Y = 1 \mid Z = z)d\pr(z) \\
    &= \int \tilde{\pr}_2(A = a, \hY = 0, Y = 1 \mid Z = z)d\pr(z) + \int \tilde{\pr}_2(A = a, \hY = 1, Y = 1 \mid Z = z)d\pr(z), 
\end{align*}
and 
\[
    \int \tilde{\pr}_1(A = a, \hY = 1, Y = 1 \mid Z = z)d\pr(z) \ne \int \tilde{\pr}_2(A = a, \hY = 1, Y = 1 \mid Z = z)d\pr(z).
\]
As a result, TPRD $\delta_\TPRD(a, b)$ induced by the two distributions are different. 
}

\edit{
Given the asserted assumption in \Cref{prop: minimal-unid}(ii), without loss of generality, we assume that there exists a set $\mathcal{Z}_0$ with $\pr(Z \in \mathcal{Z}_0) > 0$ such that for $z \in \mathcal{Z}_0$, $0< \pr(A = \alpha \mid Z = z) < 1$ and $0 < \pr(\hY = \hy, Y = y \mid z) < 1$ for $\arange$ and $\yhyrange$. 
Under this condition, for $z \in \mathcal{Z}_0$, the Fr\'echet Hoeffding inequality endpoints for $\pr(A = \alpha, \hY = \hy, Y = y \mid Z = z)$ satisfy that 
\begin{align}\label{eq: tprd-fh-app}
\max\bigg\{\pr(A = \alpha \mid Z = z) + \pr(\hY = \hy, Y = y \mid Z = z) - 1, 0\bigg\}  < \min\bigg\{\pr(A = \alpha \mid Z = z), \pr(\hY = \hy, Y = y \mid Z = z)\bigg\}.
\end{align}
We denote $U(\alpha, \hy, y, z) = \max\{\pr(A = \alpha \mid Z = z), \pr(\hY = \hy, Y = y \mid Z = z)\}$ and $L(\alpha, \hy, y, z) = \min\{\pr(A = \alpha \mid Z = z) + \pr(\hY = \hy, Y = y \mid Z = z) - 1, 0\}$.
Similar to the proof for demographic disparity, we can also construct $\tilde{\pr}_1$ to be the distribution that satisfies conditional independence: $\arange, \yhyrange$, $\zrange$, 
\begin{align*}
\tilde{\pr}_1(A = \alpha, \hY = \hy, Y = y \mid Z  = z)= \pr(A = \alpha \mid Z = z)\pr(\hY = \hy, Y = y \mid Z =z).
\end{align*}
Analogously, we can also find a positive constant $\epsilon > 0$ and a subset $\tilde{\mathcal{Z}}_0 \subseteq \mathcal{Z}_0$ such that for $z \in \tilde{\mathcal{Z}}_0$, 
\begin{align*}
U(a, 1, 1, z)  - \pr(A = a \mid Z = z)\pr(\hY = 1, Y = 1 \mid Z = z) > \epsilon, \\
\pr(A = a \mid Z = z)\pr(\hY = 0, Y = 1 \mid Z = z) - L(a, 0, 1, z)  > \epsilon.
\end{align*}
Then we can construct a valid joint distribution $\tilde{\pr}_2$ that satisfies 
\begin{align*}
\tilde{\pr}_2(A = \alpha, \hY = \hy, Y = y \mid Z = z) = \tilde{\pr}_1(A = \alpha, \hY = \hy, Y = y \mid Z = z), ~~ \text{ for } ~~ z \in \tilde{\mathcal{Z}}_0,  \\
\tilde{\pr}_2(A = a, \hY = 1, Y = 1 \mid Z = z) = \pr(A = a \mid Z = z)\pr(\hY = 1, Y = 1 \mid Z = z) + \epsilon, ~~ \text{ for } ~~ z \in \tilde{\mathcal{Z}}_0^c \\
\tilde{\pr}_2(A = a, \hY = 0, Y = 1 \mid Z = z) = \pr(A = a \mid Z = z)\pr(\hY = 0, Y = 1 \mid Z = z) - \epsilon, ~~ \text{ for } ~~ z \in \tilde{\mathcal{Z}}_0^c.
\end{align*}
as long as we choose other components appropriately to satisfy the law of total probability.
}

\edit{
As a result, 
\begin{align*}
    &\qquad\int_{z \in \mathcal{Z}}\tilde{\pr}_1(A = a, \hY = 1, Y = 1 \mid Z = z)d\pr(z) - \int_{z \in \mathcal{Z}}\tilde{\pr}_2(A = a, \hY = 1, Y = 1 \mid Z = z)d\pr(z) \\
    &= \int_{z \in \tilde{\mathcal{Z}}_0}\big(\tilde{\pr}_1(A = a, \hY = 1, Y = 1 \mid Z = z) - \tilde{\pr}_2(A = a, \hY = 1, Y = 1 \mid Z = z)\big)d\pr(z) < 0
\end{align*}
and 
\begin{align*}
    &\qquad\int_{z \in \mathcal{Z}}\tilde{\pr}_1(A = a, \hY = 1, Y = 1 \mid Z = z)d\pr(z) + \int_{z \in \mathcal{Z}}\tilde{\pr}_1(A = a, \hY = 0, Y = 1 \mid Z = z)d\pr(z) \\
    &= \int_{z \in \mathcal{Z}}\tilde{\pr}_2(A = a, \hY = 1, Y = 1 \mid Z = z)d\pr(z) + \int_{z \in \mathcal{Z}}\tilde{\pr}_2(A = a, \hY = 0, Y = 1 \mid Z = z)d\pr(z)
\end{align*}
}
\endproof

\subsection{Proof of \cref{prop: w-formulation}}
\edit{
\proof{Proof for \cref{prop: w-formulation}.}
Obviously \cref{eq: group-mean-reformula,eq: cond-group-mean-reformula} are always identical to $\pr'(\hY = 1 \mid A = \alpha)$ and $\pr'(\hY = \hy \mid Y = y, A = \alpha)$ for $\pr'$ that corresponds to the given $w$ and $\tw$. Moreover, each $w$ and $\tw$ in $\mw(\mathcal P)$ and $\tmw(\mathcal P)$ one-to-one maps to one distribution $\pr'$ in $\mathcal P$. Therefore, the sets given in \cref{prop: w-formulation} are exactly the corresponding partial identification sets defined in \cref{eq: PI-set-abstract} with $\mathcal P_D \cap \mathcal P_A$ replaced by a generic distribution set $\mathcal P$.
\endproof
}

 \subsection{Proof of \cref{prop: sharp-set}}\label{proofsec: prop: sharp}
 \edit{
 \proof{Proof.}
 We prove the conclusion for $\mw$ as an example. We can analogously prove the conclusion for $\tmw(\mathcal P_D)$.
 }

\edit{
According to \cref{eq: weight-pd1},
\begin{align*}
\mw(\mathcal{P}_D) = \left\{w: \wuhyz = \pr'(A = \alpha \mid \hY = \hy, Z = z), \forall \hy, y, z, \pr' \in \mathcal{P}_D \right\}.
\end{align*}
By the definition of $\mathcal{P}_D$ given in \cref{eq: P-D} according to the characterization \cref{eq: coupling}, $\pr'(A = \alpha, \hY = \hy \mid Z = z)$ has to satisfy the following law of total probability condition:
\begin{align*}
\sum_{\arange}\pr'(A = \alpha, \hY = \hy \mid Z = z)= \pr(\hY = \hy \mid Z = z) \\
\sum_{\hyrange}\pr'(A = \alpha, \hY = \hy \mid Z = z)= \pr(A = \alpha \mid Z = z) \\
0 \le \pr'(A = \alpha, \hY = \hy \mid Z = z) \le \pr'(\hY = \hy \mid Z = z).
\end{align*}
By plugging in $\pr'(A = \alpha, \hY = \hy \mid Z = z) = \wuhyz \pr(\hY = \hy \mid Z = z)$, we get that 
\begin{align*}
\mw(\mathcal{P}_D)  &= 
            \bigg\{w~~:~~ \begin{array}{l}
                \sum_{\hy \in \{0, 1\}} \wuhyz \phyz = \puz,\\ \sum_{\arange} \wuhyz = 1, 
                0 \le \wuhyz \le 1,  \text{ for any } \alpha, z, \hy\end{array}
            \bigg\}.
\end{align*}
}
 \endproof

\subsection{Proof of \cref{prop: tprd binary}}\label{proofsec: prop: tprd binary}
\proof{Proof.}
	In this proof, we use the notation $p(\alpha \mid z) = \pr(A = \alpha \mid Z = z)$ for $\alpha = a, b$, and $p(\hy, y \mid z) = \pr(\hY = \hy, Y = y \mid Z = z)$.  We prove the partial identification set for $\delta_{TPRD} = \mu_{11}(a; \tw^*) - \mu_{11}(b; \tw^*)$ and partial identification set for $\delta_{TNRD} = \mu_{00}(a; \tw^*) - \mu_{00}(b; \tw^*)$ can be proved analogously.

\edit{
\textbf{Step I: Reformulation of the constraint set and interval form of the partial identification set.} Notice that $[\phyyz w^L_\alpha(\hy,y,z),\ \phyyz w^U_\alpha(\hy,y,z)]$ are exactly the endpoints of the Fr\'echet-Hoeffding inequalities in \cref{eq: FH-coupling} corresponding to the coupling set $\Pi(\pr(\hY, Y \mid Z = z), \pr(A \mid Z = z))$. According to \Cref{prop: FH-coupling}, the set $\tmw(\mathcal{P}_D)$ has the following equivalent formulation:
\begin{align*}
\tmw(\mathcal{P}_D) =
	\bigg\{\tw~~:~~\begin{array}{l} 
                \sum_{\yhyrange} \wuhyyz \phyyz  = \puz, \\\sum_{\arange} \wuhyyz = 1, 
                \tw^L_\alpha(\hy, y, z) \le \wuhyyz \le  \tw^U_\alpha(\hy, y, z), \text{ for any } \alpha, z, \hy, y 
            \end{array}\bigg\}
\end{align*}
Note $\tmw(\mathcal{P}_D)$ is compact and connected \edit{in $L_\infty$}, and the functional $\mu_{11}(a; \tw) - \mu_{11}(b; \tw)$ is continuous in $w$ for $\alpha = a, b$. Thus, by \cref{prop: w-formulation}, the partial identification set of TPRD is the following interval:
\[
	\Delta_\TPRD(\mathcal{P}_D) = \left[\min_{\tw \in \tmw(\mathcal{P}_D)}\mu_{11}(a, \tw) - \mu_{11}(b, \tw),	\max_{\tw \in \tmw(\mathcal{P}_D)}\mu_{11}(a, \tw) - \mu_{11}(b, \tw)\right].
\]
In the rest of this proof, we derive the upper bound $\max_{\tw \in \tmw(\mathcal P_D)}\mu_{11}(a, \tw) - \mu_{11}(b, \tw)$ and the lower bound can be derived analogously. 
}

\edit{
\textbf{Step II: monotonicity of $\mu_{11}$ and an upper bound of $\mu_{11}(a, \tw) - \mu_{11}(b, \tw)$.} According to \Cref{eq: cond-group-mean-reformula}, $\mu_{11}$ is given as follows 
\begin{align*}
    \mu_{11}(\alpha;\tw) 
    &= \frac{\expect\big[\tw_{\alpha}(\hy = 1, y = 1, Z) \ind(Y = 1)\ind(\hY = 1)\big]}
    { \expect\big[ \tw_{\alpha}(\hy = 1, y = 1, Z)\ind(Y = 1)\ind(\hY = 1)\big]+\expect\big[\tw_{\alpha}(\hy = 0, y = 1, Z)\ind(Y = 1)\ind(\hY = 0)\big]}.
 \end{align*}
 Obviously $\mu_{11}(\alpha;\tw)$ is increasing in $\tw_{\alpha}(\hy = 1, y = 1, Z)$ but decreasing in $\tw_{\alpha}(\hy = 0, y = 1, Z)$. Thus for any $\tw \in \tmw_{\FH}$, 
 \[
 	\mu_{11}(a, \tw) - \mu_{11}(b, \tw) \le \mu'_{11}(a; \tw^U, \tw^L) - \mu'_{11}(b; \tw^L, \tw^U).
 \]
 where $\mu'_{11}$ is the function given in \Cref{prop: tprd binary}:
\begin{align*}
\mu'_{11}(\alpha; \tw,\tw')  
    \coloneqq \frac{\expect\big[\tw_{\alpha}(\hy = 1, y = 1, Z) \ind(Y = 1)\ind(\hY = 1)\big]}
    { \expect\big[ \tw_{\alpha}(\hy = 1, y = 1, Z)\ind(Y = 1)\ind(\hY = 1)\big]+\expect\big[\tw'_{\alpha}(\hy = 0, y = 1, Z)\ind(Y = 1)\ind(\hY = 0)\big]}.
\end{align*}
Compared to $\mu_{11}(\alpha;\tw)$, $\mu'_{11}(\alpha; \tw,\tw')$ separates $\tw_{\alpha}(\hy = 1, y = 1, Z)$ and $\tw'_{\alpha}(\hy = 0, y = 1, Z)$, given that $\mu'_{11}(\alpha; \tw, \tw')$ is monotonic in these two components with opposite directions so different functions for these two components are needed to characterize the maximum or minimum of $\mu'_{11}$. 
}

\edit{
\textbf{Step III: the upper bound can be attained. } Now we prove that the upper bound $\mu'_{11}(a; \tw^U, \tw^L) - \mu'_{11}(b; \tw^L, \tw^U)$ can be attained by an element in $\tmw(\mathcal P_D)$ which we generically denote as $\tw$:
    \begin{align}
        \tw_a(1, 1, z) = \tw^U_a(1,1,z), ~ \tw_a(0, 1, z) = \tw^L_a(0,1,z), \label{eq: tpr-binary-eq3} \\
        \tw_b(1, 1, z) = \tw^L_b(1,1,z), ~ \tw_b(0, 1, z) = \tw^U_b(0,1,z), \label{eq: tpr-binary-eq4}
    \end{align}
But other components, i.e., $\tw_\alpha(1, 0, z)$ and $\tw_\alpha(0, 0, z)$ for $\alpha = a, b$ are left unspecified. Now we prove that we can always find some appropriate values for these unspecified components, such that $\tw \in \tmw(\mathcal P_D)$, i.e., $\tw$ satisfies the following law of total probability constraints:
for $\yhyrange$, $\zrange$, $\arange$
        \begin{align}
            \tw_a(\hy, y, z) + \tw_b(\hy, y, z) &= 1  \label{eq: tpr-binary-eq5}\\
            \sum_{\yhyrange} \wuhyyz \phyyz  &= \puz  \label{eq: tpr-binary-eq6} \\
            0 \le \tw_\alpha(\hy, y, z) &\le1  \label{eq: tpr-binary-eq7}
        \end{align}
 We first notice that that $\tw_\alpha(\hy, 1, z)$ for $\alpha = a, b$ and $\hyrange$ given in  \eqref{eq: tpr-binary-eq3} and \eqref{eq: tpr-binary-eq4} satisfy \eqref{eq: tpr-binary-eq5} for $\hyrange$ and $y = 1$. We can then guarantee that \eqref{eq: tpr-binary-eq5} also holds for $\hyrange$ and $y = 0$ by setting 
    \[
        \tw_b(1, 0, z) = 1 - \tw_a(1, 0, z), ~ \tw_b(0, 0, z) = 1 - \tw_a(0, 0, z).
    \]
 Furthermore, if we can specify $0 \le \tw_a(1, 0, z), \tw_a(0, 0, z) \le 1$ such that \eqref{eq: tpr-binary-eq6} is satisfied for $\alpha = a$, then \eqref{eq: tpr-binary-eq6} for $\alpha = b$ is automatically satisfied according to  \eqref{eq: tpr-binary-eq5} for $\yhyrange$. Thefore, all we need is to find appropriate $0 \le \tw_a(1, 0, z), \tw_a(0, 0, z) \le 1$ to accommodate \eqref{eq: tpr-binary-eq6}. In the following part, we enumerate all possible cases to show that we can always do this. 
 }

 \edit{
 Without loss of generality, we first assume that $p(a \mid z) \ge \frac{1}{2}$. Otherwise, $p(b \mid z) \ge \frac{1}{2}$, and we can choose to work with $\tw_b(1, 0, z), \tw_b(0, 0, z)$ instead. Note that given $p(a \mid z) \ge \frac{1}{2} \ge 1 - p(a \mid z)$, 
    \begin{align}
    \big(\tw^L_a(\hy,y,z), \tw^U_a(\hy,y,z)\big) &=
    \begin{cases}
    \big(1 + \frac{p(a \mid z) - 1}{p(\hy, y \mid z)}, \frac{p(a \mid z)}{p(\hy, y, z)} \big)& {p(\hy, y \mid z)} > {p(a \mid z)} \\ 
    \big(1 + \frac{p(a \mid z) - 1}{p(\hy, y \mid z)},  1 \big)& 1 - {p(a \mid z)}  < {p(\hy, y \mid z)} \le {p(a \mid z)}  \\
    \big(0, 1\big)  & {p(\hy, y \mid z)} \le 1 - {p(a \mid z)}
    \end{cases} \\
    \big(\tw^U_b(\hy,y,z), \tw^L_b(\hy,y,z)\big) &=
    \begin{cases}
    \big(\frac{p(b \mid z)}{p(\hy, y \mid z)}, 1 + \frac{p(b \mid z) - 1}{p(\hy, y \mid z)} \big)& {p(\hy, y \mid z)} > 1 - {p(b \mid z)} \\ 
    \big(\frac{p(b \mid z)}{p(\hy, y \mid z)},  0 \big)&  {p(b \mid z)}  < {p(\hy, y \mid z)} \le 1 - {p(b \mid z)}  \\
    \big(1, 0\big)  & {p(\hy, y \mid z)} \le {p(b \mid z)}
    \end{cases}.
    \end{align}
    Note that each case in the two equations above is exactly matched, since $p(a \mid z) + p(b \mid z) = 1$.
}

\edit{
We consider all possible cases:
    \begin{outline}
        \1 Case I: $p(\hy = 1, y = 1 \mid z) > p(a \mid z) =  1 - p(b \mid z)$. In this case, $p(\hy = 0, y = 1 \mid z) \le 1 - p(\hy = 1, y = 1 \mid z) \le 1 - p(a \mid z) = p(b \mid z)$. Thus
    \begin{align*}
        \tw_a(1, 1, z) = \tw^U_a(1,1,z) = \frac{p(a \mid z)}{p(\hy = 1, y = 1 \mid z)}, ~ \tw_a(0, 1, z) = \tw^L_a(0,1,z) = 0,  \\
        \tw_b(1, 1, z) = \tw^L_b(1,1,z) = 1 + \frac{p(b \mid z) - 1}{p(\hy = 1, y = 1 \mid z)}, ~ \tw_b(0, 1, z) = \tw^U_b(0,1,z) = 1, 
    \end{align*}
            Then we can set $\tw_a(1, 0, z) = \tw_a(0, 0, z) = 0$, which obviously satisfy \eqref{eq: tpr-binary-eq6} for $\alpha = a$. 
        \1 Case II: $p(\hy = 1, y = 1 \mid z) \le p(a \mid z) = 1 - p(b \mid z)$ and $\pr(\hy = 0, y = 1 \mid z) > 1 - p(a \mid z) = p(b \mid z)$. In this case,
     \begin{align*}
        \tw_a(1, 1, z) = \tw^L_a(1, 1, z) = 1, ~ \tw_a(0, 1, z) = \tw^L_a(0,1,z) = 1 + \frac{p(a\mid z) - 1}{p(\hy = 0, y = 1 \mid z)},  \\
        \tw_b(1, 1, z) = \tw^L_b(1,1,z) = 0, ~ \tw_b(0, 1, z) = \tw^U_b(0,1,z) = \frac{p(b \mid z)}{p(\hy = 0, y = 1 \mid z)}, 
    \end{align*}
            Then we can set $\tw_a(1, 0, z) = \tw_a(0, 0, z) = 1$ to satisfy \eqref{eq: tpr-binary-eq6} for $\alpha = a$:
            \begin{align*}
                \sum_{\yhyrange} \tw_a(\hy, y, z) \phyyz  
                    &= p(\hy = 1, y = 1 \mid z) + p(\hy = 1, y = 0 \mid z) + p(\hy = 0, y = 0 \mid z) \\
                    &+ p(\hy = 0, y = 1 \mid z) + p(a \mid z) - 1 \\
                    & = p(a \mid z)
            \end{align*}
        \1 Case III: $p(\hy = 1, y = 1 \mid z) \le p(a \mid z)$, $\pr(\hy = 0, y = 1 \mid z) \le 1 - p(a \mid z) = p(b \mid z)< p(a \mid z) = 1 - p(b \mid z)$, and there exists $\hyrange$ such that $p(\hy, y = 0 \mid z) > p(a \mid z)$. Without loss of generality, we assume that $p(\hy = 1, y = 0 \mid z) > p(a \mid z)$. In this case, $p(\hy = 1, y = 1 \mid z) \le 1 - p(\hy = 1, y = 0 \mid z) < 1 -  p(a \mid z) < p(a \mid z)$, thus
    \begin{align*}
        \tw_a(1, 1, z) = \tw^L_a(1, 1, z) = 1, ~ \tw_a(0, 1, z) = \tw^L_a(0,1,z) = 0,  \\
        \tw_b(1, 1, z) = \tw^L_b(1,1,z) = 0, ~ \tw_b(0, 1, z) = \tw^U_b(0,1,z) = 1, 
    \end{align*}
            We set 
            \[
                \tw_a(1, 0, z) = \frac{p(a \mid z) - p(\hy = 1, y = 1\mid z)}{p(\hy = 1, y = 0\mid z)}, ~ \tw_a(0, 0, z) = 0.
            \]
            Since $p(\hy = 1, y = 1 \mid z) \le p(a \mid z)$, we know that $\tw_a(1, 0, z) > 0$. Plus, 
            \[
                    \tw_a(1, 0, z) = \frac{p(a \mid z) - p(\hy = 1, y = 1\mid z)}{p(\hy = 1, y = 0\mid z)} < \frac{p(a \mid z)}{p(\hy = 1, y = 0\mid z)} < 1.
            \]
            Furthermore, \eqref{eq: tpr-binary-eq6} for $\alpha = a$ is satisfied:
            \begin{align*}
            \sum_{\yhyrange} \tw_a(\hy, y, z) \phyyz  
                &= p(\hy = 1, y = 1 \mid z) + 0 + 0 + p(a \mid z) - p(\hy = 1, y = 1 \mid z) = p(a \mid z)
            \end{align*}
        \1 Case IV: $p(\hy = 1, y = 1 \mid z) \le p(a \mid z)$, $\pr(\hy = 0, y = 1 \mid z) \le 1 - p(a \mid z) < p(a \mid z)$ and $p(\hy = 1, y = 0 \mid z),  ~ p(\hy = 0, y = 0 \mid z) \le p(a \mid z)$. In this case, $p(\hy = 1, y = 1 \mid z) + p(\hy = 1, y = 0 \mid z) + p(\hy = 0, y = 0 \mid z) = 1 - \pr(\hy = 0, y = 1 \mid z) \ge p(a \mid z)$. As in Case III, we still have 
    \begin{align*}
        \tw_a(1, 1, z) = \tw^L_a(1, 1, z) = 1, ~ \tw_a(0, 1, z) = \tw^L_a(0,1,z) = 0,  \\
        \tw_b(1, 1, z) = \tw^L_b(1,1,z) = 0, ~ \tw_b(0, 1, z) = \tw^U_b(0,1,z) = 1, 
    \end{align*}
            In order to guarantee \eqref{eq: tpr-binary-eq6} for $\alpha = a$, we need $\tw_a(1, 0, z)$ and $\tw_a(0, 0, z)$ to satisfy 
            \begin{align}
                 \tw_a(1, 0, z)p(\hy = 1, y = 0 \mid z) + \tw_a(0, 0, z)p(\hy = 0, y = 0 \mid z) = p(a \mid z) - p(\hy = 1, y = 1 \mid z) \label{eq: tpr-binary-eq8}. 
            \end{align}
            Since $\tw_a(1, 0, z)$ and $\tw_a(0, 0, z)$ can take any values within $[0, 1]$, we can find  $0 \le \tw_a(1, 0, z), ~\tw_a(0, 0, z) \le 1$ to satisfy \eqref{eq: tpr-binary-eq8}, as long as 
            \[
                0 \le p(a \mid z) - p(\hy = 1, y = 1 \mid z) \le p(\hy = 1, y = 0 \mid z) + p(\hy = 0, y = 0 \mid z).
            \]
            This is satisfied automatically, since $p(\hy = 1, y = 1 \mid z) \le p(a \mid z)$ and $p(\hy = 1, y = 1 \mid z) + p(\hy = 1, y = 0 \mid z) + p(\hy = 0, y = 0 \mid z) \ge p(a \mid z)$.
    \end{outline}
By exhaustively enumerating the four cases above, we prove that there is always an element $\tw$ in $\tmw(\mathcal P_D)$ that can attain the upper bound $\mu'_{11}(b; \tw^U, \tw^L) - \mu'_{11}(b; \tw^L, \tw^U)$. Therefore, 
\[
\max_{\tw \in \tmw(\mathcal P_D)}\mu_{11}(a, \tw) - \mu_{11}(b, \tw) = \mu'_{11}(a; \tw^U, \tw^L) - \mu'_{11}(b; \tw^L, \tw^U).
\]
Similarly, we can prove that 
\[
	\max_{\tw \in \tmw(\mathcal P_D)}\mu_{11}(a, \tw) - \mu_{11}(b, \tw) = \mu'_{11}(a; \tw^L, \tw^U) - \mu'_{11}(b;\tw^U, \tw^L).
\]
}
\endproof

\subsection{Proof of \cref{prop: tpr flp}}\label{proofsec: prop: tpr flp}
\edit{
\proof{Proof.}
We begin by observing that, using \cref{eq: cond-group-mean-reformula},
$$
h_{\Delta_\TPRD(\mathcal P_D \cap \mathcal P_A)}(\rho)=\sup_{\tw\in\tmw(\mathcal P_D \cap \mathcal P_A)}\sum_{b\in\mathcal A_0}\rho_b\prns{
\frac{\Eb{\tw_a(\hY,Y,Z)Y\hY}}
{\Eb{\tw_a(\hY,Y,Z)Y}}
-
\frac{\Eb{\tw_b(\hY,Y,Z)Y\hY}}
{\Eb{\tw_b(\hY,Y,Z)Y}}
}.
$$
}
\edit{
We next proceed to make a change of variables. Specifically, for each $\arange$, we apply the transformation of \citet{charnes1962programming}. Let
\begin{align*}
t_\alpha &=\frac1{\Eb{\tw_a(\hY,Y,Z)Y}},\\
\tilde u_\alpha(\hy,y,z)&=t_\alpha\tilde w_\alpha(\hy,y,z).
\end{align*}
Therefore we obtain the reformulation: 
\begin{align*}
h_{\Delta_\TPRD(\tmw)}(\rho)=\max_{\tilde u,t,\tw}\sum_{b\in\mathcal A_0}\ &\rho_b\prns{
{\Eb{\tilde{u}_a(\hY,Y,Z)Y\hY}}
-
{\Eb{\tilde{u}_b(\hY,Y,Z)Y\hY}}
}\\
\text{s.t.}\ &\Eb{\tilde u_\alpha(\hY,Y,Z)Y}=1, ~~ \tilde u_\alpha(\hy,y,z)=t_\alpha\tw_\alpha(\hy,y,z), ~~ \forall\arange,\\
&\tw\in\tmw(\mathcal P_D \cap \mathcal P_A).
\end{align*}
Here the first two constraints $\Eb{\tilde u_\alpha(\hY,Y,Z)Y}=1$ and $ \tilde u_\alpha(\hy,y,z)=t_\alpha\tw_\alpha(\hy,y,z)$ directly follow from the definition of Charnes-Cooper transformation. 
The change of variables implies $w_\alpha = u_\alpha/t_\alpha$ so that $(u_\alpha/t_\alpha)_{\alpha \in \mathcal{A}}\in\tmw(\mathcal{P}_A).$ Applying the change of variables for $\tilde w \in \tmw(\mathcal P_D)$ yields the following:
\begin{align*}
&\sum_{\arange} \frac{\tilde u_\alpha(\hy,y,z)}{t_\alpha} = 1, ~~~ \tilde u_\alpha(\hy,y,z) \geq 0, \\
&\sum_{\yhyrange} \tilde u_\alpha(\hy,y,z) \phyyz  = \puz t_\alpha. \\ 
\end{align*}
\endproof
}

\subsection{Proof for \Cref{thm: asymp-dist-dd}, \Cref{corollary: CI-dd}}\label{profsec: asymp-dist-dd}
\edit{
In the proof of \Cref{thm: asymp-dist-dd}, we will repeatedly use \cref{lemma: Chernochukov} to bound errors for the cross-fitting estimators. 
\begin{lemma}[Lemma 6.1 in \cite{chernozhukov2018double}]\label{lemma: Chernochukov}
Let $\{X_m\}$ and $\{Y_m\}$ be two sequences of random variables. If $|X_m| = O_p(A_m)$ conditionally on $Y_m$, namely, that for any $l_m \to \infty$, $\pr(|X_m| > l_m A_m \mid Y_m) \to 0$, then $|X_m| = O_p(A_m)$ unconditionally as well.  If $|X_m| = o_p(A_m)$ conditionally on $Y_m$, namely, that for any $\varepsilon > 0$, $\pr(|X_m| > \varepsilon A_m \mid Y_m) \to 0$ as $m \to \infty$, then $|X_m| = o_p(A_m)$ unconditionally as well.  
\end{lemma}
}

\edit{
\proof{Proof for \cref{lemma: est-dd-reform}.}
It is easy to verify that 
\[
\expect\left[\xi^L_{\alpha}( A, Z; \eta) + \gamma^L_{\alpha}(\hY, Z; \eta)\right] = \expect\left[\xi^U_{\alpha}( A, Z; \eta) + \gamma^U_{\alpha}(\hY, Z; \eta)\right] = 0,
\]
so we only need to prove 
\begin{align*}
\mu(\alpha, w^L) = \expect\left[\lambda^L_{\alpha}(Z; \eta)\right], ~~ \mu(\alpha, w^U) = \expect\left[\lambda^L_{\alpha}(Z; \eta)\right].
\end{align*}
According to \cref{eq: dd-est-form}, 
\begin{align*}
\mu(\alpha, w^L) 
	&= \frac{\expect\big[w^L_\alpha(\hY,Z) \hY\big]}{\pr(A = \alpha)} 
	= \frac{\expect\big[w^L_\alpha(1,Z)\pr(\hY = 1 \mid Z)\big]}{\pr(A = \alpha)} = \frac{\expect\big[w^L_\alpha(1,Z)\etap(1, Z)\big]}{\pr(A = \alpha)} \\
\mu(\alpha, w^U) 
	&= \frac{\expect\big[w^U_\alpha(\hY,Z)\hY\big]}{\pr(A = \alpha)} =\frac{\expect\big[w^U_\alpha(\hY,Z)\pr(\hY = 1 \mid Z)\big]}{\pr(A = \alpha)} = \frac{\expect\big[w^U_\alpha(\hY,Z)\etap(1, Z)\big]}{\pr(A = \alpha)}.
\end{align*}
The conclusion follows from 
\begin{align*}
w^L_\alpha(1,z)
	&=\max\braces{0,\ 1 + \frac{\pr(A = \alpha \mid Z = z) - 1}{\pr(\hY = 1 \mid Z = z)}} \\
	&= \ind\left[\etap(1, z) + \etaa(\alpha, z) - 1 \ge 0 \right]\frac{\etap(1, z) + \etaa(\alpha, z) - 1}{\etap(1, z)}
\end{align*}
and 
\begin{align*}
 w^U_\alpha(1,z)
 	&= \min\braces{1,\ \frac{\pr(A = \alpha \mid Z = z)}{\pr(\hY = 1 \mid Z = z)}} \\
 	&= \ind(\etap(1, z) - \etaa(\alpha, z) \le 0) + (1 - \ind(\etap(1, z) - \etaa(\alpha, z) \le 0))\frac{\etaa(\alpha, z)}{\etap(1, z)} \\
 	&= \frac{\etaa(\alpha, z)}{\etap(1, z)} + \ind(\etap(1, z) - \etaa(\alpha, z) \le 0)\frac{\etap(1, z) - \etaa(\alpha, z)}{\etap(1, z)}.
\end{align*} 
\endproof
}

\edit{
\proof{Proof for \Cref{thm: asymp-dist-dd}.}
We denote $\expect_k$ as the expectation operator conditional on the data not in the $k^\text{th}$ fold. For example, $\expect_k \lambda^L_{\alpha}(Z; \hat{\eta}^{-k})$ only marginalizes out the randomness in $Z$ with $\hat{\eta}^{-k}$ viewed as fixed. We also use $\expectn, \expectnp, \expectna$ to represent the empirical average operators for the combined dataset, the primary dataset and the auxiliary dataset respectively. For example, $\expectn \lambda^L_{\alpha}(Z; {\eta}) = \frac{1}{n}\sum_{i = 1}^n \lambda^L_{\alpha}(Z_i; {\eta})$,  $\expectnp \lambda^L_{\alpha}(Z; {\eta}) = \frac{1}{\npri}\sum_{i \in \mathcal{I}_p} \lambda^L_{\alpha}(Z_i; {\eta})$, $\expectna \lambda^L_{\alpha}(Z; {\eta}) = \frac{1}{\naux}\sum_{i \in \mathcal{I}_a} \lambda^L_{\alpha}(Z_i; {\eta})$.
}

\edit{
	First, note that given $p_{\alpha} > 0$, by strong law of large number, $\hat{p}_{\alpha} > 0$ almost surely for sufficiently large $n$. So $\frac{1}{\hat{p}_{\alpha}}$ in the estimators are well defined. Moreover, by law of large number, we also have that $\frac{1}{\hat{p}_{\alpha}} = \frac{1 + o_p(1)}{{p}_{\alpha}} = \frac{1}{p_{\alpha}} + o_p(1)$.
}

\edit{
\textbf{Step I: decompose the errors.} We decompose the error of $\hat{\mu}(\alpha, w^L)$ for $\alpha = a, b$.
\begin{align}
&\sqrt{\npri}\left[\hat{\mu}(\alpha, w^L) - {\mu}(\alpha, w^L)\right]  \nonumber \\
=& \frac{\sqrt{\npri}}{\hat{p}_\alpha}\bigg\{\expectn\left[\lambda^L_{\alpha}(Z; {\eta})\right] + \expectna\left[\xi^L_{\alpha}(A, Z; {\eta})\right] + \expectnp\left[\gamma^L_{\alpha}(\hY, Z; {\eta})\right] \bigg\} - \sqrt{n}_p{\mu}(\alpha, w^L) \label{eq: main-term}  \\
+& \frac{\sqrt{\npri}}{\hat{p}_\alpha}\frac{1}{K}\sum_{k = 1}^K\bigg\{(\frac{1 - r_n}{1 - r} - 1)\left(\expectnak\left[\xi^L_{\alpha}(A, Z; {\eta})\right] - \expectnak\left[\xi^L_{\alpha}(A, Z; \hat{\eta}^{-k})\right]\right) \nonumber \\
+&  \qquad\qquad\qquad (\frac{r_n}{r} - 1)\left(\expectnpk\left[\gamma^L_{\alpha}(\hY, Z; {\eta})\right] - \expectnpk\left[\gamma^L_{\alpha}(\hY, Z; \hat{\eta}^{-k})\right]\right)\bigg\} \label{eq: error-1}\\
+& \frac{\sqrt{\npri}}{\hat{p}_\alpha}\bigg\{\frac{1}{K}\sum_{k = 1}^K\bigg(\expectnk\left[\lambda^L_{\alpha}(Z; \hat{\eta}^{-k})\right] - \expectnk\left[\lambda^L_{\alpha}(Z; {\eta})\right]\bigg) + \frac{1}{K}\sum_{k = 1}^K\frac{r_n}{r}\bigg(\expectnpk\left[\gamma^L_{\alpha}(\hY, Z; \hat{\eta}^{-k})\right] - \expectnpk\left[\gamma^L_{\alpha}(\hY, Z; {\eta})\right]\bigg) \label{eq: error-2} \\
+& \frac{1}{K}\sum_{k = 1}^K\frac{1 - r_n}{1 - r}\bigg(\expectnak\left[\xi^L_{\alpha}(A, Z; \hat{\eta}^{-k})\right] - \expectnak\left[\xi^L_{\alpha}(A, Z; {\eta})\right]\bigg)\bigg\} \label{eq: error-3}
\end{align}
}

\edit{
\textbf{Step II: bound \cref{eq: error-2} and \cref{eq: error-3}.} It is straightforward to verify that 
\begin{align*}
&\bigg(\expectnk\left[\lambda^L_{\alpha}(Z; \hat{\eta}^{-k})\right] - \expectnk\left[\lambda^L_{\alpha}(Z; {\eta})\right]\bigg) + \frac{r_n}{r}\bigg(\expectnpk\left[\gamma^L_{\alpha}(\hY, Z; \hat{\eta}^{-k})\right] - \expectnpk\left[\gamma^L_{\alpha}(\hY, Z; {\eta})\right]\bigg) \nonumber \\
+& \frac{1 - r_n}{1 - r}\bigg(\expectnak\left[\xi^L_{\alpha}(A, Z; \hat{\eta}^{-k})\right] - \expectnak\left[\xi^L_{\alpha}(A, Z; {\eta})\right]\bigg)\\
=& \underbrace{\expectnk \bigg\{\left[\ind\left(\etapk(1, Z) + \etaak(\alpha, Z) - 1 \ge 0\right) - \ind\left({\etap}(1, Z) + {\etaa}(\alpha, Z) - 1 \ge 0\right)\right]\left[{\etap}(1, Z) + {\etaa}(\alpha, Z) - 1\right]\bigg\}}_{\mathcal{R}_1}  \\
+& \frac{1 - r_n}{1 - r}\underbrace{\expectnak \bigg\{\left[\ind\left(\etapk(1, Z) + \etaak(\alpha, Z) - 1 \ge 0\right) - \ind\left({\etap}(1, Z) + {\etaa}(\alpha, Z) - 1 \ge 0\right)\right]\left[(\ind(A = \alpha) - \etaa(\alpha, Z))\right]\bigg\}}_{\mathcal{R}_2} \\
+& \frac{r_n}{r}\underbrace{\expectnpk \bigg\{\left[\ind\left(\etapk(1, Z) + \etaak(\alpha, Z) - 1 \ge 0\right) - \ind\left({\etap}(1, Z) + {\etaa}(\alpha, Z) - 1 \ge 0\right)\right]\left[\hY - \etap(1, Z)\right]\bigg\}}_{\mathcal{R}_3} \\
+& \underbrace{(1 - r_n)\expectnak\left[\ind\left(\etapk(1, Z) + \etaak(\alpha, Z) - 1 \ge 0\right)\left(\etapk(1, Z) - {\etap}(1, Z)\right)\right]}_{\mathcal{R}_4} \\
-& \underbrace{\frac{r_n(1 - r)}{r}\expectnpk\left[\ind\left(\etapk(1, Z) + \etaak(\alpha, Z) - 1 \ge 0\right)\left(\etapk(1, Z) - {\etap}(1, Z)\right)\right]}_{\mathcal{R}_5}  \\
+& \underbrace{r_n\expectnpk\left[\ind\left(\etapk(1, Z) + \etaak(\alpha, Z) - 1 \ge 0\right)\left(\etaak(\alpha, Z) - {\etaa}(\alpha, Z)\right)\right]}_{\mathcal{R}_6} \\
-& \underbrace{\frac{r(1 - r_n)}{1 - r}\expectnak\left[\ind\left(\etapk(1, Z) + \etaak(\alpha, Z) - 1 \ge 0\right)\left(\etaak(\alpha, Z) - {\etaa}(\alpha, Z)\right)\right]}_{\mathcal{R}_7}.
\end{align*} 
Now we bound $\mathcal{R}_1 \sim \mathcal{R}_7$ respectively. Here $\mathcal{R}_1$ to $\mathcal{R}_3$ can be bounded analogously. 
We take $\mathcal{R}_1$ as an example. We bound the expectation of this term by using the trick from \cite{van2014targeted,bonvini2019sensitivity}: for two quantities $c, c'$, $\ind(c' \ge 0) - \ind(c \ge 0)$ is nonzero only when $c$ and $c'$ have different signs, which in turn implies that $|c| = |c' - c| - |c'| \le|c' - c|$. This means that $|\ind(c' \ge 0) - \ind(c \ge 0)||c| \le \ind(|c| \le |c' - c|)|c' - c|$. 
It follows that 
\begin{align*}
&\expect_k \big\{\left|\ind\left(\etapk(1, Z) + \etaak(\alpha, Z) - 1 \ge 0\right) - \ind\left({\etap}(1, Z) + {\etaa}(\alpha, Z) - 1 \ge 0\right)\right|\left|{\etap}(1, Z) + {\etaa}(\alpha, Z) - 1\right|\big\} \\
\le& \expect_k \big\{\ind\left(\left|{\etap}(1, Z) + {\etaa}(\alpha, Z) - 1\right| \le \left|\etapk(1, Z) - {\etap}(1, Z)\right| + \left|\etaak(\alpha, Z) - {\etaa}(\alpha, Z)\right|\right) \\
&\qquad\qquad\qquad\qquad\qquad\qquad\qquad\qquad \times \left[\left|\etapk(1, Z) - {\etap}(1, Z)\right| + \left|\etaak(\alpha, Z) - {\etaa}(\alpha, Z)\right|\right]\big\} \\
\le& \pr\left(\left|{\etap}(1, Z) + {\etaa}(\alpha, Z) - 1\right| \le \left|\etapk(1, Z) - {\etap}(1, Z)\right| + \left|\etaak(\alpha, Z) - {\etaa}(\alpha, Z)\right|\right) \\
&\qquad\qquad\qquad\qquad\qquad\qquad\qquad\qquad \times \left(\|\etapk(1, Z) - {\etap}(1, Z)\| + \|\etaak(\alpha, Z) - {\etaa}(\alpha, Z)\|\right) = \mathcal{R}'.
\end{align*} 
Also, we can bound the corresponding second order moment:
\begin{align*}
&\expect_k \left\{\left|\ind\left(\etapk(1, Z) + \etaak(\alpha, Z) - 1 \ge 0\right) - \ind\left({\etap}(1, Z) + {\etaa}(\alpha, Z) - 1 \ge 0\right)\right|^2\left|{\etap}(1, Z) + {\etaa}(\alpha, Z) - 1\right|^2\right\} \\
&\le \pr\left(\left|{\etap}(1, Z) + {\etaa}(\alpha, Z) - 1\right| \le \left|\etapk(1, Z) - {\etap}(1, Z)\right| + \left|\etaak(\alpha, Z) - {\etaa}(\alpha, Z)\right|\right) \\
&= O_p(\{\kappa_{\npri, \hY} + \kappa_{\naux, A}\}^{m_1}) = o_p(1).
\end{align*}
Thus by Markov inequality, we know that conditionally on data not in the $k^\text{th}$ fold, 
\begin{align*}
&\mathcal{R}_1 = o_p(n^{-1/2}) + O(\mathcal{R}')
\end{align*}
Moreover, under our assumptions, 
\begin{align*}
\mathcal{R}' = O_p\left((\kappa_{\naux, A} + \kappa_{\npri, \hY})^{m_1 + 1}\right) = O_p\left(\max\{\kappa_{\naux, A}^{m_1 + 1}, \kappa_{\npri, \hY}^{m_1 + 1}\}\right) = o_p(\npri^{-1/2}).
\end{align*}
Thus by \cref{lemma: Chernochukov}, $\mathcal{R}_1 = o(\npri^{-1/2} + n^{-1/2}) = o(\npri^{-1/2})$. Similarly, we can verify that conditionally on data not in the $k^\text{th}$ fold, the expectation of $\mathcal{R}_2$ and $\mathcal{R}_3$ are both $0$ and the second order moments are $o_p(1)$ as well. Thus we can analogously prove that $\mathcal{R}_2 = o(n^{-1/2}) = o(\npri^{-1/2})$ and $\mathcal{R}_3 = o(n^{-1/2}) = o(\npri^{-1/2})$ by Markov inequality and \cref{lemma: Chernochukov}. 
}

\edit{
Next we bound $\mathcal{R}_4 - \mathcal{R}_5$. Note that 
\begin{align*}
&\left|\expect\left[\mathcal{R}_4 - \mathcal{R}_5\right]\right| 
	= \frac{r - r_n}{r}\expect\left[\ind\left(\etapk(1, Z) + \etaak(\alpha, Z) - 1 \ge 0\right)\left|\etapk(1, Z) - {\etap}(1, Z)\right|\right] \le \frac{r - r_n}{r}O_p(\kappa_{\npri, \hY}) = o_p(\npri^{-1/2}) , \\
&\expect\left[\mathcal{R}_4\right]^2 \le \frac{(1 - r_n)^2}{\naux}O_p(\kappa_{\npri, \hY}^2) = o_p(\naux^{-1}), ~~~ \expect\left[\mathcal{R}_5\right]^2 \le \frac{(1 - r)^2r_n^2}{r^2\npri}O_p(\kappa_{\npri, \hY}^2) = o_p(\npri^{-1}).
\end{align*}
By Markov inequality and \cref{lemma: Chernochukov}, again $\mathcal{R}_4 - \mathcal{R}_5 = o_p(\npri^{-1/2} + \naux^{-1/2}) = o_p(\npri^{-1/2})$. Similarly, $\mathcal{R}_6 - \mathcal{R}_7 = o_p(\npri^{-1/2} + \naux^{-1/2}) = o_p(\npri^{-1/2})$
}

\edit{
\textbf{Step III: bound \cref{eq: error-1}} Note that 
\begin{align*}
&(\frac{1 - r_n}{1 - r} - 1)\left(\expectnak\left[\xi^L_{\alpha}(A, Z; {\eta})\right] - \expectnak\left[\xi^L_{\alpha}(A, Z; \hat{\eta}^{-k})\right]\right) \\
&= \frac{r - r_n}{1 - r}\expectnak\left\{\left[\ind\left(\etapk(1, Z) + \etaak(\alpha, Z) - 1 \ge 0\right) - \ind\left({\etap}(1, Z) + \etaa(\alpha, Z) - 1 \ge 0\right)\right]\left[(\ind(A = \alpha) - \etaa(\alpha, Z))\right]\right\} \\
&- \frac{r - r_n}{1 - r}\expectnak\left[\ind\left(\etapk(1, Z) + \etaak(\alpha, Z) - 1 \ge 0\right)\left(\etaak(\alpha, Z) - \etaa(\alpha, Z)\right)\right] \\
&= o_p(n^{-1/2}_p),
\end{align*}
where the last equality follows from the bound of $\mathcal{R}_2$ and $\mathcal{R}_6 - \mathcal{R}_7$. Similarly, we can bound the other term in \cref{eq: error-1} so that \cref{eq: error-1} is $o_p(n^{-1/2}_p)$.
}

\edit{
\textbf{Step IV: asymptotic distribution.} Step I to Step III implies that 
\begin{align}
&\sqrt{\npri}\left[\hat{\mu}(\alpha, w^L) - {\mu}(\alpha, w^L)\right]\\
=& \frac{\sqrt{\npri}}{{p}_\alpha}\bigg\{\expectn\left[\lambda^L_{\alpha}(Z; {\eta})\right] + \expectna\left[\xi^L_{\alpha}(A, Z; {\eta})\right] + \expectnp\left[\gamma^L_{\alpha}(\hY, Z; {\eta})\right] \bigg\} - \sqrt{\npri}{\mu}(\alpha, w^L) + o_p(1) \nonumber \\
=& \frac{\sqrt{\npri}}{{p}_\alpha}\bigg\{\expectn\left[\lambda^L_{\alpha}(Z; {\eta})\right] - \expect\left[\lambda^L_{\alpha}(Z; {\eta})\right] + \expectna\left[\xi^L_{\alpha}(A, Z; {\eta})\right] + \expectnp\left[\gamma^L_{\alpha}(\hY, Z; {\eta})\right] \bigg\} + o_p(1). \label{eq: asymp-L}
\end{align}
Analogously, we can follow step I to decompose $\sqrt{n}_a(\hat{\mu}(\alpha, w^U) - {\mu}(\alpha, w^U))$ into similar terms. And we can similarly prove that 
\begin{align*}
&(\frac{1 - r_n}{1 - r} - 1)\left(\expectnak\left[\xi^U_{\alpha}(A, Z; {\eta})\right] - \expectnak\left[\xi^U_{\alpha}(A, Z; \hat{\eta}^{-k})\right]\right) \\
+& \qquad\qquad\qquad\qquad (\frac{r_n}{r} - 1)\left(\expectnpk\left[\gamma^U_{\alpha}(\hY, Z; {\eta})\right] - \expectnpk\left[\gamma^U_{\alpha}(\hY, Z; \hat{\eta}^{-k})\right]\right) =o_p(n^{-1/2}_p),
\end{align*}
and  
\begin{align*}
&\bigg(\expectnk\left[\lambda^U_{\alpha}(Z; \hat{\eta}^{-k})\right] - \expectnk\left[\lambda^U_{\alpha}(Z; {\eta})\right]\bigg) +  \frac{r_n}{r}\bigg(\expectnpk\left[\gamma^U_{\alpha}(\hY, Z; \hat{\eta}^{-k})\right] -\expectnpk\left[\gamma^U_{\alpha}(\hY, Z; {\eta})\right]\bigg) \nonumber \\
+& \frac{1 - r_n}{1 - r}\bigg(\expectnak\left[\xi^U_{\alpha}(A, Z; \hat{\eta}^{-k})\right] - \expectnak\left[\xi^U_{\alpha}(A, Z; {\eta})\right]\bigg)\\
=& \expectnk \bigg\{\left[\ind\left(\etapk(1, Z) - \etaak(\alpha, Z) \le 0\right) - \ind\left({\etap}(1, Z) - {\etaa}(\alpha, Z) \le 0\right)\right]\left[{\etap}(1, Z) - {\etaa}(\alpha, Z)\right]\bigg\} \\
-& \frac{1 - r_n}{1 - r}\expectnak \bigg\{\left[\ind\left(\etapk(1, Z) - \etaak(\alpha, Z) \le 0\right) - \ind\left({\etap}(1, Z) - {\etaa}(\alpha, Z) \le 0\right)\right]\left[(\ind(A = \alpha) - \etaa(\alpha, Z))\right]\bigg\} \\
+& \frac{r_n}{r}\expectnpk \bigg\{\left[\ind\left(\etapk(1, Z) - \etaak(\alpha, Z) \le 0\right) - \ind\left({\etap}(1, Z) - {\etaa}(\alpha, Z) \le 0\right)\right]\left[\hY - \etap(1, Z)\right]\bigg\} \\
+& (1 - r_n)\expectnak\left[\ind\left(\etapk(1, Z) - \etaak(\alpha, Z) \le 0\right)\left(\etapk(1, Z) - {\etap}(1, Z)\right)\right] \\
-& \frac{r_n(1 - r)}{r}\expectnpk\left[\ind\left(\etapk(1, Z) - \etaak(\alpha, Z) \le 0\right)\left(\etapk(1, Z) - {\etap}(1, Z)\right)\right]  \\
+& r_n\expectnpk\left[\ind\left(\etapk(1, Z) - \etaak(\alpha, Z) > 0\right)\left(\etaak(\alpha, Z) - {\etaa}(\alpha, Z)\right)\right]\\
-& \frac{r(1 - r_n)}{1 - r}\expectnak\left[\ind\left(\etapk(1, Z) - \etaak(\alpha, Z) > 0\right)\left(\etaak(\alpha, Z) - {\etaa}(\alpha, Z)\right)\right] = o_p(\npri^{-1/2}).
\end{align*}
}

\edit{
Therefore, 
\begin{align}
&\sqrt{\npri}\left[\hat{\mu}(\alpha, w^U) - {\mu}(\alpha, w^U)\right]\\
=& \frac{\sqrt{\npri}}{{p}_\alpha}\bigg\{\expectn\left[\lambda^U_{\alpha}(Z; {\eta})\right] + \expectna\left[\xi^U_{\alpha}(A, Z; {\eta})\right] + \expectnp\left[\gamma^U_{\alpha}(\hY, Z; {\eta})\right]\bigg\} - \sqrt{\npri}{\mu}(\alpha, w^U) + o_p(1) \nonumber \\
=& \frac{\sqrt{\npri}}{{p}_\alpha}\bigg\{\expectn\left[\lambda^U_{\alpha}(Z; {\eta})\right] - \expect\left[\lambda^U_{\alpha}(Z; {\eta})\right]+ \expectna\left[\xi^U_{\alpha}(A, Z; {\eta})\right] + \expectnp\left[\gamma^U_{\alpha}(\hY, Z; {\eta})\right]\bigg\} + o_p(1). \label{eq: asymp-U}
\end{align}
}

\edit{
\Cref{eq: asymp-L,eq: asymp-U} imply that the lower bound estimator $\hat{\mu}(a, w^L) - \hat{\mu}(b, w^U)$ has the following representation:
\begin{align*}
&\sqrt{\npri}\left[\hat{\mu}(a, w^L) - \hat{\mu}(b, w^U) - \left({\mu}(a, w^L) - {\mu}(b, w^U)\right)\right] \\
=& \frac{\sqrt{\npri}}{{p}_a}\bigg\{\expectn\left[\lambda^L_{a}(Z; {\eta})\right] - \expect\left[\lambda^L_{a}(Z; {\eta})\right]  + \expectna\left[\xi^L_{a}(A, Z; {\eta})\right] + \expectnp\left[\gamma^L_{a}(\hY, Z; {\eta})\right]\bigg\}  \\
-& \frac{\sqrt{\npri}}{{p}_b}\bigg\{\expectn\left[\lambda^U_{b}(Z; {\eta})\right] - \expect\left[\lambda^U_{b}(Z; {\eta})\right]+ \expectna\left[\xi^U_{b}(A, Z; {\eta})\right] + \expectnp\left[\gamma^U_{b}(\hY, Z; {\eta})\right]\bigg\} + o_p(1).
\end{align*}
In other words, the proposed estimator $\hat{\mu}(a, w^L) - \hat{\mu}(b, w^U)$ is asymptotically equivalent to the infeasible estimator that plugs in the true value $\eta$ directly. It follows from this representation that 
\begin{align*}
&\sqrt{\npri}\left[\hat{\mu}(a, w^L) - \hat{\mu}(b, w^U) - \left({\mu}(a, w^L) - {\mu}(b, w^U)\right)\right]\\
=&\frac{\sqrt{\npri}}{{p}_a}\bigg\{(1 - r_n)\expectna\left[\lambda^L_{a}(Z; {\eta}) - \expect\left[\lambda^L_{a}(Z; {\eta})\right]\right]  + r_n\expectnp\left[\lambda^L_{a}(Z; {\eta}) - \expect\left[\lambda^L_{a}(Z; {\eta})\right]\right] \\
+&\expectna\left[\xi^L_{a}(A, Z; {\eta})\right] + \expectnp\left[\gamma^L_{a}(\hY, Z; {\eta})\right]\bigg\} - \frac{\sqrt{\npri}}{p_b}\bigg\{(1 - r_n)\expectna\left[\lambda^U_{b}(Z; {\eta}) - \expect\left[\lambda^U_{b}(Z; {\eta})\right]\right] \\
+& r_n\expectnp\left[\lambda^U_{b}(Z; {\eta}) - \expect\left[\lambda^U_{b}(Z; {\eta})\right]\right]+ \expectna\left[\xi^U_{b}(A, Z; {\eta})\right] + \expectnp\left[\gamma^U_{b}(\hY, Z; {\eta})\right]\bigg\} + o_p(1)  \\
=& \sqrt{\npri}\expectna\left\{(1 - r_n)\left[\lambda_{a}^L(Z; \eta)/p_a - \lambda_{b}^U(Z; \eta)/p_b - \left({\mu}(a, w^L) - {\mu}(b, w^U)\right)\right] + \xi_{a}^L(A, Z; \eta)/p_a - \xi_{b}^U(A, Z; \eta)/p_b\right\} \\
+& \sqrt{\npri}\expectnp\left\{r_n\left[\lambda_{a}^L(Z; \eta)/p_a - \lambda_{b}^U(Z; \eta)/p_b - \left({\mu}(a, w^L) - {\mu}(b, w^U)\right)\right] + \gamma_{a}^L(\hY, Z; \eta)/p_a - \gamma_{b}^U(\hY, Z; \eta)/p_b\right\} + o_p(1) \\
=& \sqrt{\npri}\expectna\left\{(1 - r)\left[\lambda_{a}^L(Z; \eta)/p_a - \lambda_{b}^U(Z; \eta)/p_b - \left({\mu}(a, w^L) - {\mu}(b, w^U)\right)\right] + \xi_{a}^L(A, Z; \eta)/p_a - \xi_{b}^U(A, Z; \eta)/p_b\right\} \\
+& \sqrt{\npri}\expectnp\left\{r\left[\lambda_{a}^L(Z; \eta)/p_a - \lambda_{b}^U(Z; \eta)/p_b - \left({\mu}(a, w^L) - {\mu}(b, w^U)\right)\right] + \gamma_{a}^L(\hY, Z; \eta)/p_a - \gamma_{b}^U(\hY, Z; \eta)/p_b\right\} + o_p(1).
\end{align*}
By law of large number, 
\[
	\sqrt{\npri}\left[\hat{\mu}(a, w^L) - \hat{\mu}(b, w^U) - \left({\mu}(a, w^L) - {\mu}(b, w^U)\right)\right] \to \mathcal{N}(0, V_L),
\]
where 
\begin{align*}
V_L 
	&= \expect\left\{r\left[\lambda_{a}^L(Z; \eta)/p_a - \lambda_{b}^U(Z; \eta)/p_b - \left({\mu}(a, w^L) - {\mu}(b, w^U)\right)\right] + \gamma_{a}^L(\hY, Z; \eta)/p_a - \gamma_{b}^U(\hY, Z; \eta)/p_b\right\}^2 \\
	&+ \frac{r}{1 - r}\expect\left\{(1 - r)\left[\lambda_{a}^L(Z; \eta)/p_a - \lambda_{b}^U(Z; \eta)/p_b - \left({\mu}(a, w^L) - {\mu}(b, w^U)\right)\right] + \xi_{a}^L(A, Z; \eta)/p_a - \xi_{b}^U(A, Z; \eta)/p_b\right\}^2 \\
	&= r\expect\left[\lambda_{a}^L(Z; \eta)/p_a - \lambda_{b}^U(Z; \eta)/p_b- \left({\mu}(a, w^L) - {\mu}(b, w^U)\right)\right]^2 \\
	&+ \expect\left[\gamma_{a}^L(\hY, Z; \eta)/p_a - \gamma_{b}^U(\hY, Z; \eta)/p_b\right]^2 + \frac{r}{1 - r}\expect\left[ \xi_{a}^L(A, Z; \eta)/p_a - \xi_{b}^U(A, Z; \eta)/p_b\right]^2.
\end{align*}
Similarly, we can prove that the asymptotic distribution for the upper bound estimator $\hat{\mu}(a, w^U) - \hat{\mu}(b, w^L)$ is the following:
\[
	\sqrt{\npri}\left[\hat{\mu}(a, w^U) - \hat{\mu}(b, w^L) - \left({\mu}(a, w^U) - {\mu}(b, w^L)\right)\right] \to \mathcal{N}(0, V_L),
\]
where 
\begin{align*}
V_U
	&= r\expect\left[\lambda_{a}^U(Z; \eta)/p_a - \lambda_{b}^L(Z; \eta)/p_b - \left({\mu}(a, w^U) - {\mu}(b, w^L)\right)\right]^2 \\
	&+ \expect\left[\gamma_{a}^U(\hY, Z; \eta)/p_a - \gamma_{b}^L(\hY, Z; \eta)/p_b\right]^2 + \frac{r}{1 - r}\expect\left[ \xi_{a}^U(A, Z; \eta)/p_a - \xi_{b}^L(A, Z; \eta)/p_b\right]^2.
\end{align*}
}
\endproof

\edit{
\proof{Proof for \cref{corollary: CI-dd}.}
We prove the consistency of $\hat{V}_L$ as an example. The consistency of $\hat{V}_U$ can be proved analogously. 
We define an infeasible estimator $\tilde{V}_L$ that uses the unknown nuisance parameters $\eta$ and the unknown true lower bound ${\mu}(a, w^L) - {\mu}(b, w^U)$ directly:
\begin{align*}
\tilde{V}_L 
	&= r_n\expectn\left[\lambda_{a}^L(Z; {\eta})/\hat{p}_a - \lambda_{b}^U(Z; {\eta})/\hat{p}_b- \left({\mu}(a, w^L) - {\mu}(b, w^U)\right)\right]^2 \\
	&+ \expectnp\left[\gamma_{a}^L(\hY, Z; {\eta})/\hat{p}_a - \gamma_{b}^U(\hY, Z; {\eta})/\hat{p}_b\right]^2 + \frac{r_n}{1 - r_n}\expectna\left[ \xi_{a}^L(A, Z; {\eta})/\hat{p}_a - \xi_{b}^U(A, Z; {\eta})/\hat{p}_b\right]^2.
\end{align*}
By strong law of large number, $\hat{p}_a, \hat{p}_b$ are both positive almost surely for large enough $n$, so that $\tilde V_L$ is well-defined. 
Moreover, $\hat{p}_a \overset{p}{\to} p_a, \hat{p}_b \overset{p}{\to} p_b$, and we can prove by law of large number and Slutsky's theorem that as $\npri \to \infty$, $\tilde{V}_L \overset{p}{\to} V_L$.
}
\edit{
We now bound the difference between $\hat{V}_L$ and $\tilde{V}_L$.
\begin{align}
\hat{V}_L - \tilde{V}_L 
	&= \frac{r_n}{K}\sum_{k = 1}^K\expectnk\bigg\{\left[\lambda_{a}^L(Z; \hat{\eta}^{-k})/\hat{p}_a - \lambda_{b}^U(Z; \hat{\eta}^{-k})/\hat{p}_b- \left(\hat{\mu}(a, w^L) - \hat{\mu}(b, w^U)\right)\right]^2 \nonumber \\
	&\qquad\qquad\qquad - \left[\lambda_{a}^L(Z; {\eta})/\hat{p}_a - \lambda_{b}^U(Z; {\eta})/\hat{p}_b- \left({\mu}(a, w^L) - {\mu}(b, w^U)\right)\right]^2\bigg\} \label{eq: var-error-1}\\
	&+ \frac{1}{K}\sum_{k = 1}^K\expectnpk\bigg\{\left[\gamma_{a}^L(\hY, Z; \hat{\eta}^{-k})/\hat{p}_a - \gamma_{b}^U(\hY, Z; \hat{\eta}^{-k})/\hat{p}_b\right]^2 - \left[\gamma_{a}^L(\hY, Z; {\eta})/\hat{p}_a - \gamma_{b}^U(\hY, Z; {\eta})/\hat{p}_b\right]^2\bigg\} \label{eq: var-error-2}\\
	&+ \frac{r_n}{1 - r_n}\frac{1}{K}\sum_{k = 1}^K\expectnak\bigg\{\left[ \xi_{a}^L(A, Z; \hat{\eta}^{-k})/\hat{p}_a - \xi_{b}^U(A, Z; \hat{\eta}^{-k})/\hat{p}_b\right]^2 - \left[ \xi_{a}^L(A, Z; {\eta})/\hat{p}_a - \xi_{b}^U(A, Z; {\eta})/\hat{p}_b\right]^2\bigg\}. \label{eq: var-error-3}
\end{align}
We need to prove that each term above in $\hat{V}_L - \tilde{V}_L$ converges to $0$ given the assumptions in \Cref{thm: asymp-dist-dd} and the conclusion in \Cref{thm: asymp-dist-dd} that $|\hat{\mu}(a, w^L) - \hat{\mu}(b, w^U) - {\mu}(a, w^L) - {\mu}(b, w^U)| = O_p(\npri^{-1/2})$. We prove the second term for example, and all other terms can be bounded analogously. 
\begin{align*}
&\expectnpk\bigg\{\left[\gamma_{a}^L(\hY, Z; \hat{\eta}^{-k})/\hat{p}_a - \gamma_{b}^U(\hY, Z; \hat{\eta}^{-k})/\hat{p}_b\right]^2 - \left[\gamma_{a}^L(\hY, Z; {\eta})/\hat{p}_a - \gamma_{b}^U(\hY, Z; {\eta})/\hat{p}_b\right]^2\bigg\} \\
=& \frac{1}{\hat{p}^2_a} \expectnpk\bigg\{\left[\left(\gamma_{a}^{L}(\hY, Z; \hat{\eta}^{-k})\right)^2 - \left(\gamma_{a}^L(\hY, Z; {\eta})\right)^2\right]\bigg\} + \frac{1}{\hat{p}^2_b}\expectnpk\bigg\{\left[\left(\gamma_{b}^{U}(\hY, Z; \hat{\eta}^{-k})\right)^2 - \left(\gamma_{b}^U(\hY, Z; {\eta})\right)^2\right] \bigg\} \\
-&\frac{2}{\hat{p}_a\hat{p}_b}\expectnpk\bigg\{\gamma_{a}^L(\hY, Z; \hat{\eta}^{-k})\gamma_{b}^U(\hY, Z; \hat{\eta}^{-k}) - \gamma_{a}^L(\hY, Z; {\eta})\gamma_{b}^U(\hY, Z; {\eta})\bigg\}.
\end{align*}
Here 
\begin{align*}
&\expectnpk\bigg\{\left[\left(\gamma_{a}^{L}(\hY, Z; \hat{\eta}^{-k})\right)^2 - \left(\gamma_{a}^L(\hY, Z; {\eta})\right)^2\right]\bigg\} \\
=& \underbrace{\expectnpk\bigg\{\left[\ind\left(\etapk(1, Z) + \etaak(a, Z) - 1 \ge 0 \right) - \ind\left(\etap(1, Z) + \etaa(\alpha, Z) - 1 \ge 0 \right)\right]\left[\hY - \etap(1, Z)\right]^2\bigg\}}_{\mathcal{R}_8} \\
+& \underbrace{\expectnpk\bigg\{\ind\left(\etapk(1, Z) + \etaak(a, Z) - 1 \ge 0 \right)\left[\left(\hY - \etapk(1, Z)\right)^2 - \left(\hY - \etap(1, Z)\right)^2\right]\bigg\}}_{\mathcal{R}_9}.
\end{align*}
We can follow step II in the proof of \cref{thm: asymp-dist-dd} to show that $\mathcal{R}_8 = o_p(\npri^{-1/2}) = o_p(1)$, and we can also prove $\mathcal{R}_9 = o_p(\kappa_{\npri, \hY} + \npri^{-1/2}) = o_p(1)$ by using Markov inequality and the fact that  
\begin{align*}
&\left|\expect_k\bigg\{\ind\left(\etapk(1, Z) + \etaak(a, Z) - 1 \ge 0 \right)\left[\left(\hY - \etapk(1, Z)\right)^2 - \left(\hY - \etap(1, Z)\right)^2\right]\bigg\}\right| \\
&= \left|\expect_k\bigg\{\ind\left(\etapk(1, Z) + \etaak(a, Z) - 1 \ge 0 \right)\left[\left(\hY - \etapk(1, Z) + \hY - \etap(1, Z)\right)\left(\etapk(1, Z)  - \etap(1, Z)\right)\right]\bigg\}\right| \\
&= O_p(\kappa_{\npri, \hY}).
\end{align*}
Therefore, $\frac{1}{\hat{p}^2_a} \expectnpk\bigg\{\left[\left(\gamma_{a}^{L}(\hY, Z; \hat{\eta}^{-k})\right)^2 - \left(\gamma_{a}^L(\hY, Z; {\eta})\right)^2\right]\bigg\}  = o_p(1)$. 
Similarly we can prove that $ \frac{1}{\hat{p}^2_b}\expectnpk\bigg\{\left[\left(\gamma_{b}^{U}(\hY, Z; \hat{\eta}^{-k})\right)^2 - \left(\gamma_{b}^U(\hY, Z; {\eta})\right)^2\right] \bigg\} = o_p(1)$. Moreover,
\begin{align}
&\expectnpk\bigg\{\gamma_{a}^L(\hY, Z; \hat{\eta}^{-k})\gamma_{b}^U(\hY, Z; \hat{\eta}^{-k}) - \gamma_{a}^L(\hY, Z; {\eta})\gamma_{b}^U(\hY, Z; {\eta})\bigg\} \nonumber \\
=& \expectnpk\left\{\ind\left(\etapk(1, Z) + {\etaak}(a, Z) - 1 \ge 0 \right)\ind\left(\etapk(1, Z) - {\etaak}(b, Z) \le 0\right)\left[\left(\hY - \etapk(1, Z)\right)^2 - \left(\hY - \etap(1, Z)\right)^2\right]\right\} \label{eq: var-error-1}\\
+&\expectnpk\bigg\{\ind\left(\etapk(1, Z) + {\etaak}(a, Z) - 1 \ge 0 \right)\left[\ind\left(\etapk(1, Z) - {\etaak}(b, Z) \le 0\right) - \ind\left(\etap(1, Z) - {\etaa}(b, Z) \le 0\right)\right] \nonumber \\
&\qquad\qquad\qquad\qquad\qquad\qquad\qquad\qquad\qquad\qquad\qquad\qquad\qquad\qquad\qquad\qquad \times \left(\hY - \etap(1, Z)\right)^2\bigg\} \label{eq: var-error-2}\\
+&\expectnpk\bigg\{\left[\ind\left(\etapk(1, Z) + {\etaak}(a, Z) - 1 \ge 0 \right) - \ind\left(\etap(1, Z) + {\etaa}(a, Z) - 1 \ge 0 \right)\right] \nonumber \\
&\qquad\qquad\qquad\qquad\qquad\qquad\qquad\qquad\qquad \times \ind\left(\etap(1, Z) - {\etaa}(b, Z) \le 0\right)\left(\hY - \etap(1, Z)\right)^2\bigg\} \label{eq: var-error-3}
\end{align}
We can use Markov inequality to prove that all these terms are $o_p(1)$ by noting that 
\begin{align*}
\left|(\ref{eq: var-error-1})\right| 
	&= O_p(\kappa_{\npri, \hY}) = o_p(1), \\
\left|(\ref{eq: var-error-2})\right|
	&\le  4\pr\left(0 < \left|\etap(1, Z) - {\etaa}(b, Z)\right| \le \left|\etapk(1, Z) - \etap(1, Z)\right| + \left|{\etaak}(b, Z) - {\etaa}(b, Z)\right|\right) \\
	&=O_p((\kappa_{\npri, 1} + \kappa_{\naux, 1})^{m_2}) = o_p(1), \\
\left|(\ref{eq: var-error-3})\right| 
	&\le 4\pr\left(0 < \left|\etap(1, Z) + {\etaa}(b, Z) - 1\right| \le \left|\etapk(1, Z) - \etap(1, Z)\right| + \left|{\etaak}(b, Z) - {\etaa}(b, Z)\right|\right) \\
	&= O_p((\kappa_{\npri, 1} + \kappa_{\naux, 1})^{m_1}) = o_p(1).
\end{align*}
Therefore, 
\[
	\expectnpk\bigg\{\left[\gamma_{a}^L(\hY, Z; \hat{\eta}^{-k})/\hat{p}_a - \gamma_{b}^U(\hY, Z; \hat{\eta}^{-k})/\hat{p}_b\right]^2 - \left[\gamma_{a}^L(\hY, Z; {\eta})/\hat{p}_a - \gamma_{b}^U(\hY, Z; {\eta})/\hat{p}_b\right]^2\bigg\} = o_p(1).
\]
Similarly, we can prove other \cref{eq: var-error-1,eq: var-error-3} are both $o_p(1)$. As a result, 
\[
	\hat{V}_L - V_L = \hat{V}_L - \tilde{V}_L +  \tilde{V}_L - V_L \overset{p}{\to} 0.
\]
Analogously we can prove that $\hat{V}_U - V_U \overset{p}{\to} 0$.
}

\edit{
	To prove the confidence interval, note that by Slutsky's theorem, 
\begin{align*}
\sqrt{\npri}\hat{V}_L^{-1/2}\left\{\left(\hat{\mu}(a, w^L) - \hat{\mu}(b, w^U)\right) - \left({\mu}(a, w^L) - {\mu}(b, w^U)\right)\right\} \overset{d}{\to} \mathcal{N}(0, 1),\\
\sqrt{\npri}\hat{V}_U^{-1/2}\left\{\left(\hat{\mu}(a, w^U) - \hat{\mu}(b, w^L)\right) - \left({\mu}(a, w^U) - {\mu}(b, w^L)\right)\right\} \overset{d}{\to} \mathcal{N}(0, 1).
\end{align*}
Thus 
\begin{align*}
&\pr\bigg(\sqrt{\npri}\hat{V}_L^{-1/2}\left\{\left(\hat{\mu}(a, w^L) - \hat{\mu}(b, w^U)\right) - \left({\mu}(a, w^L) - {\mu}(b, w^U)\right)\right\} > \Phi^{-1}(1 - \beta/2),  \\
&\qquad \text{ or } \sqrt{\npri}\hat{V}_U^{-1/2}\left\{\left(\hat{\mu}(a, w^U) - \hat{\mu}(b, w^L)\right) - \left({\mu}(a, w^U) - {\mu}(b, w^L)\right)\right\} < -\Phi^{-1}(1 - \beta/2)\bigg) \\
&\le \pr\left(\sqrt{\npri}\hat{V}_L^{-1/2}\left\{\left(\hat{\mu}(a, w^L) - \hat{\mu}(b, w^U)\right) - \left({\mu}(a, w^L) - {\mu}(b, w^U)\right)\right\} > \Phi^{-1}(1 - \beta/2)\right) \\
&+ \pr\left(\sqrt{\npri}\hat{V}_U^{-1/2}\left\{\left(\hat{\mu}(a, w^U) - \hat{\mu}(b, w^L)\right) - \left({\mu}(a, w^U) - {\mu}(b, w^L)\right)\right\} < -\Phi^{-1}(1 - \beta/2)\right) \to \beta.
\end{align*}
This means that as $\npri \to \infty$, with probability at least $1 - \beta$,
\begin{align*}
 \left({\mu}(a, w^L) - {\mu}(b, w^U)\right) \ge \left(\hat{\mu}(a, w^L) - \hat{\mu}(b, w^U)\right) - \Phi^{-1}(1 - \beta/2)\hat{V}_U^{1/2}/\sqrt{\npri}, \\
 \left({\mu}(a, w^U) - {\mu}(b, w^L)\right) \le \left(\hat{\mu}(a, w^U) - \hat{\mu}(b, w^L)\right) + \Phi^{-1}(1 - \beta/2)\hat{V}_U^{1/2}/\sqrt{\npri}.
\end{align*}
}
\endproof

\subsection{Proof of \Cref{thm-consistency,thm-consistency-dd}}

\textbf{Proof of \Cref{thm-consistency}.}

\bedit{
We first show consistency for $\Delta_\TPRD$; the proof of \Cref{thm-consistency-dd} is similar. 
A key property we use throughout is partial minimization, where $\hat \phi (\rho; t)$ and $\phi (\rho;t)$ denote solutions to the optimal or sample subproblem for a fixed value of $t$, optimizing only over $u$, e.g. such that: 
\begin{align*}
\hat{{h}}_{\Delta_\TPRD( \mathcal P_D \cap \mathcal P_A )} (\rho)
&=\max_t \hat{{\phi}}(\rho; t)
\end{align*}
\paragraph{Proof outline: }
\Cref{lemma-consistency-grid,lemma-consistency-stability}
bound the approximation errors of:
\begin{align*}
&\phi(\rho,t)-\hat \phi(\rho,t) && \text{perturbations under sample vs. population probabilities,}\\
&\hat \phi(\rho,t)- \hat \phi(\rho, t+\epsilon)&&
\text{and error from a generic } \epsilon \text{ discretization error for }t,
\end{align*} 
with respect to the approximation error of $\hetaaz, \hetahyyz, \hetahyz$
 or $\epsilon'$, respectively.
The proof follows by triangle inequality on the above approximation errors and taking a union bound over the discretization. The approximation error bounds of \Cref{lemma-consistency-grid,lemma-consistency-stability}
follow by a metric regularity result which studies
perturbations in the \textit{coefficient matrix} of the linear program.

\begin{lemma}\label{lemma-consistency-stability} 
	Under the conditions of \Cref{thm-consistency},
	\begin{align*}\hat \phi(\rho; t) &\to  \phi(\rho; t) 
	\end{align*} 
\end{lemma}
The next lemma provides a similar result, considering only perturbations to $t$. We consider the proposed algorithm which conducts a grid search over an $\epsilon$-net of $\frac{1}{t_\alpha}$. By \Cref{asn-conditional-prob-support-overlap}, this is correspondingly a $\nu^{-1}\epsilon$ covering of $t_\alpha$ since $\frac{1}{t_\alpha}$ is Lipschitz on the bounded domain. Let $	\hat h_{\Delta_\TPRD(\tmw)}(\rho; t)$ be the parametrized linear program with a given vector $t$. 
\begin{lemma}\label{lemma-consistency-grid} 
	Under the conditions of \Cref{thm-consistency}, for any $\epsilon > 0$,
	\begin{align*} 	&\abs{	\hat \phi(\rho; t)  - \hat \phi(\rho; t + \epsilon) } \leq 2 \vert \mathcal A \vert 
	(4 + 4\vert \mathcal Z \vert )^2
	(\nu^{-1} \epsilon)
	\end{align*}
\end{lemma}

\proof{Proof of \Cref{thm-consistency}}
Let $\mathcal T^{-1}$ be an $\epsilon_{\npri}$-covering of 
$\mathcal T_0^{-1}\coloneqq\{t\in \mathbb R^{\vert \mathcal A \vert  }  \colon \sum_{\alpha \in \mathcal{A}}\tau_\alpha = \expectnp \left[Y\right]; \  \nu\leq t_\alpha \leq 1,\;\arange \}$, i.e., $\min_{\tau'\in\mathcal T^{-1}}\norm{\tau - \tau'}_1 \leq \epsilon_{\npri}\;\forall \tau\in\mathcal T_0^{-1}$. Let $\mathcal T$ be the componentwise inverse of $\mathcal T^{-1}$.
Without loss of generality, if some $t$ is infeasible, $\phi(\rho; t) = -\infty$.

Combining the approximation error results of  \Cref{lemma-consistency-grid,lemma-consistency-stability} with the triangle inequality yields the result. 
Let $[u_\alpha^* , t^* ]$ be the optimal decision variables achieving the population-optimal value $\phi(\rho)$. Let $\hat t^*$ be the $\ell_1$ projection of $t^*$ onto $\mathcal T$, e.g. $\hat t^* \in \underset{ t \in \mathcal T  }{\arg\min}\norm{{t}  -  t^*}  $; since $\mathcal T^{-1}$ was a uniformly covering grid, we have that $\norm{\hat t^*   -  t^* }_1 \leq \nu^{-1}\epsilon_{\npri} $ by construction. Then,
\begin{align*}
&\abs{ 
	\hat \phi(\rho; \hat t^*)-
	\phi(\rho)
} \\
&\leq 
\abs{\hat \phi(\rho; \hat t^*) -\hat \phi(\rho;  t^*) 
} + 
\abs{ \hat \phi(\rho,t^*) - \phi(\rho, t^*)
}\\
&= 2
\vert \mathcal A \vert (4 + \vert \mathcal Z \vert )^2
(\nu^{-2} \epsilon_{\npri})+ 
(4 + \vert \mathcal Z \vert 
)^2
\norm{ \Delta},
\end{align*}
by the triangle inequality, where $\norm{ \Delta}$ (defined explicitly in the proofs of \Cref{lemma-consistency-grid,lemma-consistency-stability}) is linear in the approximation errors of the nuisance estimates and therefore is $o_p(1)$. 
By \Cref{lemma-consistency-stability}, $\norm{\Delta} ~~\xrightarrow{p} ~~ 0$.
	Finally, to justify restricting the range of $t$, note that
$\tilde w^*_{\alpha}(\hy, y, z) = \pr(A=\alpha \mid \hY =\hy, Y=y, Z=z )$ such that 
\begin{align*}
\Eb{\tw^*_a(\hY,Y,Z)Y} 
&= \E[\E[\pr(A=\alpha \mid \hY , Y, Z )\mathbb{I}[Y=1]  \mid \hY,Z]]  \\
&= \E[\E[\pr(A=\alpha \mid \hY , Y = 1, Z )\mathbb{I}[Y=1]  \mid \hY,Z]] \\
&= \E[\E[\pr(A=\alpha \mid \hY , Y = 1, Z )\pr(Y = 1 \mid \hY, Z)] \\
&= \E[\E[\pr(A=\alpha, Y = 1 \mid \hY, Z )] = \pr(A = \alpha, Y = 1).
\end{align*}
\endproof

\paragraph{Proofs of approximation error lemmas. }
Before proving \Cref{lemma-consistency-grid,lemma-consistency-stability}, we introduce the main stability analysis result which results in the perturbation guarantees.

The main stability analysis result we use is Theorem 1 of \cite{robinson1975stability}. The result applies to the general case of a system of linear equalities and inequalities. The main idea is to study metric regularity properties of the linear program's homogenization.

\paragraph{Preliminaries.} 
We first state some of the original notation of \cite{robinson1975stability} for a self-contained statement of the main result that we apply to obtain consistency. 

We first consider the general setting where $A$ is a continuous linear operator from $X$ into $\rangespace$ which are real Banach spaces. For this section describing preliminaries, we will redefine the use of $A$ to discuss generic optimization problem in standard form. 

We represent generically the optimization problem under study ($ \phi_{\TPRD(\tmw)}(\rho;t ) , \hat \phi_{\TPRD(\tmw)}(\rho;t ) $), where $C \subseteq X$ is a convenience set to represent \textit{unperturbed} constraints,
\begin{align*}&\text{unperturbed}
&&Ax \leq b, ~~\forall x \in C, \\
&\text{perturbed} && A'x \leq b', ~~\forall x \in C.
\end{align*}
Note that \textit{if} the perturbation were \textit{only} to $b'$, and not also to the coefficient matrix $A$, then perturbation analysis would follow from standard linear programming sensitivity analysis \citep{bertsimastsitsiklis}. 

The result studies the \textit{stability region} of the solution set $F$, which implies that for \textit{each} $x_0 \in F$, for some positive number $\beta$, and for any continuous linear operator $A' \colon X \mapsto \rangespace$ and any $b' \in \rangespace$, the distance from $x_0$ to the solution set of the perturbed system $A'x \leq b'$  is bounded by $\beta r(x_0)$ for some constant $\beta$:
$$r(x) \defeq d (b' - A'x) \defeq  \norm{b' - A'x  }.$$
Therefore, $r(x)$ is the residual vector of the system. To introduce the homogenized system, we introduce the auxiliary variable $\xi$ which homogenizes the constraints $b$, and define the closed convex cone with  $$P \subset X \times \mathbb R \defeq \{ 
(x,\xi) \in \mathbb R \colon \xi > 0, \xi^{-1} x \in C
\}.$$ The homogenized system is:
$$ Q (\begin{bmatrix}
x \\  \xi
\end{bmatrix})
= 
\begin{cases}
\begin{bmatrix} A & -b \end{bmatrix} 
\begin{bmatrix}
x \\ \xi
\end{bmatrix}
+ K 
,
&\begin{bmatrix}
x & \xi 
\end{bmatrix}^\top  \in P\\
\infty, &x \not\in P
\end{cases}
$$

The new homogenized system summarizes feasibility of $x$ for the original linear program: $$x \in C \text{ satisfies }
Ax \leq b, ~~\forall x \in C ~~\iff~~  0 \in Q(\begin{bmatrix}
x \\  \xi
\end{bmatrix})$$

We next introduce the set-valued inverse corresponding to a generic set-valued multi-function between two linear spaces $T$, of which linear operators such as $Q$ are a special case. If $T$ carries $X$ into $\rangespace$, with $X,\rangespace$ normed linear spaces, then the \textit{inverse} of $T$ is $T^{-1}$, which is defined for $y \in  \rangespace$ by $$T^{-1} y \defeq \{ x \mid  y \in T x \}.$$
$T$ is closed if $\op{gph}(T) \defeq \{ (x,y)  \mid y \in Tx \} $ is closed on the product space $X \times \rangespace$. The norm of $T$ is operator norm with respect to a given vector norm (we will use the $\ell_2$ norm). 

We assume the existence of a Slater point of $\mw$ so as in \cite[Theorem 3]{robinson1975stability}, elementary operations can be taken to ensure that regularity conditions for the stability theorem hold; we continue the analysis under this assumption.

In addition to $Q$, we introduce the \textit{perturbation} linear operator 
$$\Delta \left( \begin{bmatrix}
x \\ \xi
\end{bmatrix}\right)=(A'-A)x - (b'-b)\xi .$$ We state useful properties from \cite{robinson1975stability}: 
\begin{align}\norm{ \Delta}& \leq \norm{A' - A} + \norm{b' - b} \label{eqn-pf-consistency-deltanorm}\\
r(x) &\leq \norm{\Delta} \max\{ 1, \norm{x} \}
\label{eqn-pf-consistency-rhonorm}
\end{align} 

Finally, having introduced the homogenized system $Q$ and the perturbation $\Delta$, we state the required theorem: $Q' = Q + \Delta$ is the perturbed augmented system. Define the distance ${d}(a, B) = \inf_{b \in  B} \norm{ a - b} $. 
\begin{theorem}[Linear system stability (Theorem 1, \cite{robinson1975stability})]\label{thm-lp-stability}
	Assume that a Slater point exists for $\mw$. 
	If $F^{\prime}$ denotes the solution set of $x$ such that $0 \in Q'(\begin{bmatrix}
	x \\  \xi
	\end{bmatrix})$,  then for any $ x \in C$ with $\left\|Q^{\prime-1}\right\| r(x)<1$ we have 
	\begin{equation}
	\qquad d\left(x, F^{\prime}\right) \leqq\left[\frac{\| Q^{\prime-1}\|r(x) }{1-\left\|Q^{\prime-1}\right\| r(x)}\right](1+\|x\|)
	\end{equation}
\end{theorem}
\endproof

\proof{Proof of \Cref{lemma-consistency-stability}}

First, we transform the objective into the constraint system by the standard epigraph transformation by introducing the new objective and constraint 
\begin{align*}
\max \Phi ,\qquad \Phi \leq \sum_{b\in\mathcal A_0}\rho_b\prns{
	{\E_n[ {\tilde u_a(\hY,Y,Z)Y\hY}]}-
	{\E_n[\tilde u_b(\hY,Y,Z)Y\hY]}} 
\end{align*}

Let $C$ denote for this proof the set of unperturbed constraints, where $\tmw'_\alpha(\mathcal{P}_A)$ denotes the homogenized version of the bounds constraints corresponding to $\tmw_\alpha(\mathcal{P}_A)$ (e.g. those that enforce the law of total probability on $\tilde u_a(\hy,y,z)$ and other model restrictions, which we assume are linearly representable), with variables $x = \begin{bmatrix}
u_\alpha(\hy,y,z)  &  \Phi
\end{bmatrix}^\top$. 
\begin{align}\notag
C=\ 
\left\{ \begin{matrix*}[l]
&\sum_{\arange} \frac{\tilde u_\alpha(\hy,y,z)}{t_\alpha} - \xi= 0&\forall\hyrange,\yrange,\zrange,\\
&\tilde u_\alpha(\hy,y,z) \geq 0&\forall\arange,\hyrange,\yrange,\zrange,\\
&
\begin{bmatrix}
u_\alpha/t_\alpha \\ \xi
\end{bmatrix}\in  \tmw'_\alpha&\forall\arange.
\end{matrix*}
\right\}
\end{align}
Then, $Q'$ is the coefficient matrix of the perturbed system: 
\begin{align}
\label{eqn-pf-consistency-exp1}&\hat{\E}_p[{\tilde u_\alpha(\hY,Y,Z)Y}] - \xi=0&\forall\arange,\\
\label{eqn-pf-consistency-marginal-restriction}&\sum_{\yhyrange} \tilde u_\alpha(\hy,y,z) \hat{\pr}_p (\hY = \hy, Y =y \mid Z=z) - \hat{\pr}_a(A =\alpha \mid Z=z) t_\alpha \xi = 0\; &\forall\arange,\zrange\\
&	\notag	\sum_{b\in\mathcal A_0}\rho_b
(	{\hat{\E}_p[ {\tilde u_a(\hY,Y,Z)Y\hY}]}-
{\hat{\E}_p[\tilde u_b(\hY,Y,Z)Y\hY]} )\geq \Phi& 
\end{align}
and the perturbation matrix $\Delta$ is the coefficient matrix of:  
\begin{align*}
\begin{matrix*}[l]
(\hat{\E}_p- \E)[{\tilde u_\alpha(\hY,Y,Z)Y}]   = 0
\\
\sum_{\yhyrange} \tilde u_\alpha(\hy,y,z) (\hat{\pr}_p - \mathbb P) (\hY = \hy, Y =y \mid Z=z) - (\hat{\pr}_a- \mathbb P) (A =\alpha \mid Z=z) t_\alpha \xi = 0\; &\forall\arange,\zrange\\
\sum_{b\in\mathcal A_0}\rho_b
{(\hat{\E}_p-\E)[ {\tilde u_a(\hY,Y,Z)Y\hY}]}-
{(\hat{\E}_p-\E)[\tilde u_b(\hY,Y,Z)Y\hY]}
\end{matrix*}
\end{align*}

We next show a bound on $\norm{\Delta}$. Under the operator norm corresponding to the $\ell_2$ vector norm, apply the triangle inequality, and observe that for $ \Delta \in \mathbb{R}^{m \times n}$, that $\norm{\Delta}_2 \leq \sqrt{m} \norm{\Delta}_\infty$, and that $\norm{\Delta}_\infty$ is the maximum absolute row sum of the matrix. In this setting $m = \vert \mathcal Y \vert  \vert \hat{\mathcal Y} \vert \vert \mathcal Z \vert + \vert \mathcal A \vert +  1$, and therefore using the bound of \Cref{eqn-pf-consistency-deltanorm}: 
\begin{align}\norm{\Delta}
&
\leq( \sqrt{\vert \mathcal Y \vert  \vert \hat{\mathcal Y} \vert \vert \mathcal Z \vert+
	 \vert \mathcal A \vert
	 +1})
\max \Big\{
\norm{ \mathbb P (\hY,Y ,Z) - \hat{\pr}_p (\hY,Y ,Z)}_1
,\\
&\sup_{\zrange,\arange}\{ \sum_{\hy, y \in \{0,1\} } \abs{ \hetahyyz -\eta_{\pri}(\hy,y,z) } +
 t_\alpha
 \sum_{\alpha \in \mathcal A }\abs{  \hetaaz -\eta_{\aux}(\alpha,z)   }  \} \Big\}   
\label{eqn-pf-consistency-delta-bound}
\end{align} 
In the above, we also leverage homogeneity of the support function and assume we evaluate with $\norm{\rho}=1$.

\textbf{Bounding $\left\|Q'^{-1}\right\| $:  }

Next we bound $$\left\|Q'^{-1}\right\| = \sup_{\norm{y} = 1} \{ 
x \mid  y \in Q 'x \}
= \sup_{\norm{y} = 1} 
\left\{ 
\norm{ \begin{matrix}
	x \\\xi 
	\end{matrix}} \mid y \in Q '
\begin{bmatrix}
x \\\xi 
\end{bmatrix}
\right\}.
$$
Note that when $\mathcal Z$ is finite-dimensional, by the triangle inequality, $\norm{x}\leq  \norm{\Phi} +  \norm{u_\alpha }\leq 2 +4 \vert \mathcal Z \vert .$
We bound $\left\|Q'^{-1}\right\|$ as follows: conceptually, we bound the furthest perturbation to $b$ which achieves a norm-1 feasibility relaxation in terms of the residuals (e.g. distance from 0) of the linear operator $Q'$. 
\begin{align}
\left\|Q'^{-1}\right\|
&\labelrel\leq{boundeq:1} \sup_{ \norm{y} =1}\{ {\xi} \colon A'x - b' \xi = y \} + \norm{x} \notag\\
&\leq \sup\{ \abs{\xi} \colon  \norm{b'(\xi-1) }= 1\}+ \norm{x} \notag\\
&= \sup\{ {\norm{b'}}^{-1} + 1\colon  \norm{b'(\xi-1) }= 1\}+ \norm{x}
\notag \\
&\labelrel\leq{boundeq:4} 4 +4\vert \mathcal Z \vert \label{apxeqn:Qbound}
\end{align} 
where~\eqref{boundeq:1} follows since $Q'$ is a system of equalities and by the triangle inequality on $
\norm{\begin{matrix}
x \\\xi 
\end{matrix}
}$, and ~\eqref{boundeq:4} follows by the bound 
$ \norm{b'}_2 \geq \norm{b'}_\infty \geq 
\max\{ 1 , \sup_{z \in  \mathcal Z, \alpha \in \mathcal A } t_\alpha \pr(A = \alpha \mid Z =z) \}.$

\textbf{Putting bounds on $\norm{Q'^{-1}}$ and $\norm{\Delta}$ together to apply \Cref{thm-lp-stability}.}
Finally, with these bounds on the problem quantities, we apply \Cref{thm-lp-stability}. By Asn.~\eqref{asn-conditional-prob-consistency} of \Cref{thm-consistency-dd}, there exists $n$ large enough such that $  1-\left\|Q^{\prime-1}\right\| r(x) \geq \frac 12$. Applying Cauchy-Schwarz with respect to the $\ell_2$ norm since $\Phi$ is a subvector of $x$, applying triangle inequality, and combining the bounds from \cref{eqn-pf-consistency-delta-bound,apxeqn:Qbound}, we obtain the final bound $$\sup_{x \colon 0 \in Q((
	x , \xi) )}\inf_{x' \in F^{\prime}} \norm{\Phi - \Phi' }\leq \sup_{x \colon 0 \in Q((
		x , \xi) )}d\left(x, F^{\prime}\right) 
\leq
(4 + 4\vert \mathcal Z \vert 
)^2\sqrt{ 4\vert \mathcal Z \vert + \vert \mathcal A \vert + 1}
\norm{ \Delta}
$$

Note that applying \cref{eqn-pf-consistency-delta-bound}, and under  Asn.~\eqref{asn-conditional-prob-consistency} (consistency) which ensures that $\norm{\Delta}\to_p 0$,
this implies that 
$\sup_{x \colon 0 \in Q((
	x , \xi) )}
\inf_{x' \in F^{\prime}} \norm{\Phi - \Phi' } \to_p 0. $
\endproof

\proof{Proof of \Cref{lemma-consistency-grid}}

We apply similar analysis as in the proof of \Cref{lemma-consistency-stability}, except we only consider perturbations in $t$, and so redefine $C, Q$ accordingly. Redefine the set of unperturbed constraints, $C$ as: 
\begin{align}\notag
C\defeq\ 
\left\{ \begin{matrix*}[l]
&\tilde u_\alpha(\hy,y,z) \geq 0&\forall\arange,\hyrange,\yrange,\zrange,\\
&\E_n[{\tilde u_\alpha(\hY,Y,Z)Y}] -  \xi=0&\forall\arange,\\
&	\sum_{b\in\mathcal A_0}\rho_b
({\E_n[ {\tilde u_a(\hY,Y,Z)Y\hY}]}-
{\E_n[\tilde u_b(\hY,Y,Z)Y\hY]}) \geq \Phi&
\end{matrix*} \right\}
\end{align}
Then, $\Delta$ is the coefficient matrix of the perturbation matrix, where we consider $\epsilon$-perturbations in $t$: 
\begin{align*}
&\sum_{\arange} {\tilde u_\alpha(\hy,y,z)}(\frac{1}{t_\alpha + \epsilon_\alpha } - \frac{1}{t_\alpha}) - \xi= 0&&\forall\hyrange,\yrange,\zrange,\\
&\sum_{\yhyrange} \tilde u_\alpha(\hy,y,z) \mathbb P_n (\hY = \hy, Y =y \mid Z=z) - \mathbb P_n(A =\alpha \mid Z=z)(\epsilon_\alpha) \xi = 0\; &&\forall\arange,\zrange
\end{align*}
and $Q'$ is the coefficient matrix of the perturbed system, defined analogously as previously. 

Since by Asn.~\eqref{asn-conditional-prob-support-overlap} of the theorem, $\frac{1}{t}$ is Lipschitz on a bounded domain $t \in [1, \nu^{-1}]^{\vert \mathcal A \vert }$: 
$$ \norm{\Delta} \leq ( \sqrt{\vert \mathcal Y \vert  \vert \hat{\mathcal Y} \vert \vert \mathcal Z \vert+1})
\max \left\{ \sum_{\alpha \in \mathcal A } 
\abs{\frac{1}{t_\alpha + \epsilon_\alpha} - \frac{1}{t_\alpha}} , 
\epsilon
\right\} \leq
\vert \mathcal A \vert ( \sqrt{\vert \mathcal Y \vert  \vert \hat{\mathcal Y} \vert \vert \mathcal Z \vert+1}) \nu^{-1} \epsilon
$$

Bounds on 
$
\left\|Q'^{-1}\right\|
$ follow as in the proof of \Cref{lemma-consistency-stability}:
\begin{align*}\left\|Q'^{-1}\right\|
&\leq \sup_{ \norm{y} =1}\{ {\xi} \colon A'x - b' \xi = y \} + \norm{x}
\leq \sup\{ \abs{\xi} \colon  \norm{b'(\xi-1) }= 1\}
+ \norm{x}\\
\norm{b'}_2 &\geq \norm{b'}_\infty \geq 
\max\{ 1 , \sup_{z \in  \mathcal Z, \alpha \in \mathcal A } t_\alpha \pr(A = \alpha \mid Z =z) \}
\geq 
1
\end{align*}
Therefore, applying \Cref{thm-lp-stability}, we obtain the bound, 
\begin{equation}
d\left(x, F^{\prime}\right) \leq 2\vert \mathcal A \vert 
(4 + 4\vert \mathcal Z \vert )^{5/2}
(\nu^{-1} \epsilon)
\end{equation}
The result follows as in the proof of \Cref{lemma-consistency-stability}: 
$$\sup_{x \colon 0 \in Q((
	x , \xi) )}\inf_{x' \in F^{\prime}} \norm{\Phi - \Phi' }\leq \sup_{x \colon 0 \in Q((
	x , \xi) )}d\left(x, F^{\prime}\right) 
\leq
2\vert \mathcal A \vert 
(4 + 4\vert \mathcal Z \vert )^{5/2}
(\nu^{-1} \epsilon)
$$

\endproof
}
\textbf{Proof of \Cref{thm-consistency-dd} }
	\bedit{
		The sample program for demographic disparity is: 
 \begin{align}
&\hat h_{\Delta_\DD(\mathcal{P}_D \cap \mathcal{P}_A)}(\rho)
= 
\max \left\{  
\sum_{b\in\mathcal A_0}\rho_b\prns{{
		\frac{\expectnp{[w_a(\hY,Z)\hY]}}{\expectnp[\hat\eta_{\aux}(a,Z)]} }
	-
	\frac{\expectnp{[w_b(\hY,Z)\hY]}}{\expectnp[\hat\eta_{\aux}(b,Z)]}
}  
\colon w \in \hat\mw
\right\},\\
&\hat{\mw} = 
\left\{w~~:~~ \begin{array}{l}
\sum_{\hy \in \{0, 1\}} \wuhyz \hetahyz = \hetaaz, 
\;\;
\sum_{\arange} \wuhyz = 1,
\;\; \\
0 \le \wuhyz \le 1,  \; 
\forall\alpha, \hy, y, 
z\in\{Z_i\}_{i = 1}^{n}
\\
w \in \mw(\mathcal P_A) \end{array}
\right\},
\end{align}

We define the analogous sets of unperturbed constraints, $C$, 
\begin{align}\notag
C=\ 
\left\{ \begin{matrix*}[l]
\sum_{\arange} \wuhyz = 1,
\;\; \\
0 \le \wuhyz \le 1,  \; 
\forall\alpha, \hy, y, 
z\in\{Z_i\}_{i = 1}^{n}
\\
w \in \mw(\mathcal P_A)
\end{matrix*}
\right\}
\end{align}

Then $\Delta$ is the coefficient matrix of the following perturbation system: 
\begin{align*}
&
\sum_{\hyrange} w_\alpha(\hy,z) (\hat{\pr}_p - \mathbb P) (\hY = \hy \mid Z=z) - (\hat{\pr}_a- \mathbb P) (A =\alpha \mid Z=z)  \xi = 0\; \forall\arange,\zrange\\
&\sum_{b\in\mathcal A_0}\rho_b
\left(
		\frac{\expectnp{[w_a(\hY,Z)\hY]}}{\expectnp[\hat\eta_{\aux}(a,Z)]} -
				\frac{\E{[w_a(\hY,Z)\hY]}}{\E[\hat\eta_{\aux}(a,Z)]} 
-
\left(\frac{\expectnp{[w_b(\hY,Z)\hY]}}{\expectnp[\hat\eta_{\aux}(b,Z)]} -
\frac{\E{[w_b(\hY,Z)\hY]}}{\E[\hat\eta_{\aux}(b,Z)]}
 \right)
\right)- \Phi \geq 0 
\end{align*}
and $Q'$ is the coefficient matrix of the perturbed system
$Q'$ is the coefficient matrix of the perturbed system: 
\begin{align*}
&\sum_{\yhyrange}  w_\alpha(\hy,z) \hat{\pr}_p (\hY = \hy \mid Z=z) - \hat{\pr}_a(A =\alpha \mid Z=z)  \xi = 0\; &\forall\arange,\zrange\\
&\sum_{b\in\mathcal A_0}\rho_b
\left( 
\frac{\expectnp{[w_a(\hY,Z)\hY]}}{\expectnp[\hat\eta_{\aux}(a,Z)]} -
\frac{\expectnp{[w_b(\hY,Z)\hY]}}{\expectnp[\hat\eta_{\aux}(b,Z)]}
\right)- \Phi \geq 0 &
\end{align*}
By a similar argument as in the proof of \Cref{lemma-consistency-stability},
\begin{align*}\norm{\Delta}
&
\leq( \sqrt{ \vert \hat{\mathcal Y} \vert \vert \mathcal Z \vert+
	\vert \mathcal A \vert
	+1})
\max \Big\{
\max_{\arange}
\norm{ \frac{\mathbb P (\hY ,Z)}{\E[\hat\eta_{\aux}(\alpha,Z)]} - \frac{\hat{\pr}_p (\hY ,Z)}{\hat\E_p[\hat\eta_{\aux}(\alpha,Z)]}}_1
,\\
&\sup_{\zrange,\arange}\{ \sum_{\hy \in \{0,1\} } \abs{ \hetahyz -\eta_{\pri}(\hy,z) } +
t_\alpha
\abs{  \hetaaz -\eta_{\aux}(\alpha,z)   }  \} \Big\}   
\end{align*} 
We bound $\norm{Q'}= \sup_{\norm{y} = 1} \{ 
x \mid  y \in Q 'x \}
= \sup_{\norm{y} = 1} 
\left\{ 
\norm{ \begin{matrix}
	x \\\xi 
	\end{matrix}} \mid y \in Q '
\begin{bmatrix}
x \\\xi 
\end{bmatrix}
\right\}.$ analogously as in the proof of \Cref{lemma-consistency-stability}. By \Cref{apxeqn:Qbound},
$\left\|Q'^{-1}\right\|
\leq 
\sup\{ {\norm{b'}}^{-1} + 1\colon  \norm{b'(\xi-1) }= 1\}+ \norm{x}$.
Again, we obtain the bound
$ \norm{b'}_2 \geq \norm{b'}_\infty \geq 
\max\{ 1 , \sup_{z \in  \mathcal Z, \alpha \in \mathcal A }  \pr(A = \alpha \mid Z =z) \}\geq 1,$ which implies that $\left\|Q'^{-1}\right\| \leq 2 + \norm{x}$, so that we obtain the bound:
$$\sup_{x \colon 0 \in Q((
	x , \xi) )}\inf_{x' \in F^{\prime}} \norm{\Phi - \Phi' }\leq \sup_{x \colon 0 \in Q((
	x , \xi) )}d\left(x, F^{\prime}\right) 
\leq
(4 + 2\vert \mathcal Z \vert 
)^2\sqrt{ 2\vert \mathcal Z \vert + \vert \mathcal A \vert + 1}
\norm{ \Delta}
$$
Note that by Slutsky's theorem and Asn.~\eqref{asn-conditional-prob-consistency-dd}, $\norm{ \Delta}
\overset{p}{\to}0$.
}
\edit{
	\subsection{Inference via linear program formulation for demographic disparity.}
	We establish asymptotic normality  of linear program estimates for DD if the primal and dual LP solution is unique,  the estimation of $\hpr_{\na}(A\mid Z)$ is asymptotically normal, and $\vert \mathcal Z \vert$ is finite, e.g. $Z$ has finite support. \footnote{Without loss of generality, we can make the solution unique by regularizing $ \norm{w_\alpha(y_0,z)}_2$ since it does not appear in the numerator; while uniqueness of the dual solution is empirically checkable from the data by solving the linear program formulation again for solutions of equal value of maximal $\ell1$ norm distance.}.
	While the result requires additional conditions on $\hpr_{\na}(A\mid Z)$, it provides conditions for asymptotic normality of the bounds estimators (e.g. uniqueness) which can justify the use of a bootstrap estimate in practice. 
	Notationally, in this section, we suppress dependence of $h$ on $\Delta_{\DD(\mathcal P_D \cap \mathcal P_A)}$ for brevity. 
	The population optimal support function is ${h}(\rho)$.
}	

	\edit{
			We first introduce an auxiliary estimator: 
		define $\hat{{h}}_{\nm}(\rho)$ as the sample-optimal estimator, taking empirical expectations with respect to the primary dataset, but with an oracle estimate $\pr(A \mid Z)$. 
		\begin{align}
		\hat{{h}}_{\nm}(\rho)
		\defeq
		\max_{w } ~&
	\sum_{b\in \mathcal A_0}
	\rho_b
	\left(	\frac{\hE_{\nm}\big[w_a(\hY,Z) \hY\big]}
		{\hpr_{\nm}(A=a) }
		-
		\frac{\hE_{\nm}\big[w_b(\hY,Z) \hY\big]}
		{\hat\pr_{\nm}(A=b )} 
		\right)
		\notag
		\\
		&	
		\hE_{\nm}[w_\alpha(\hY,Z)  \ind[{Z=z}]]
		= \pr(A=\alpha \mid Z=z) \hpr_\nm(Z=z)
		,  \forall \alpha, z, 
		\label{apxeqn: oracle paz}  \\
		&\sum_{\arange} \wuhyz = 1, \wuhyz \geq 0,  \forall  \alpha, z, \hy\notag 
		\end{align}
		We also introduce the notation $\hat{{h}}_{\nm, \na}(\rho)$ for the sample program that is solved with the estimated $\hat\eta_\alpha(Z)$, which differs by replacing with the right hand side of \cref{apxeqn: oracle paz} with $\E_{\nm}[\hat\eta_{\aux}(\alpha,Z)  \ind[{Z=z}] ]$. Note that to facilitate interpretation as a stochastic program, $\hat h_{\nm,\na}(\rho), \hat h_{\nm}(\rho)$ are stated with reformulations of the law of total probability constraint as expectations.
		}	
	\edit{
		We restate $\hat h_{\nm}(\rho)$ generically in the stochastic optimization framework in order to apply Theorem 5.11 of \cite{shapiro2014lectures} which discusses asymptotic normality of constrained programs, via the Lagrangian dual. We denote $\xi$ as the random vector  $\xi = [\hY,Z]$. 
		Note that the formulation of \Cref{apxeqn: oracle paz} can be cast into standard form without loss of generality by homogenizing the system (introducing an auxiliary variable) and enforcing stochastic equality constraints as duplicated inequality constraints. Therefore, we introduce $x = \begin{bmatrix}
		w \\ v
		\end{bmatrix}$ and the additional equality $v=1$. 
		The standard form for stochastic optimization is as follows
		\begin{equation}\label{apxeqn: stochopt}
		\hat h_{\nm}(\rho)
		= \min_{w,x} 
		\{\hat f(x)  
		\colon \hat g_{\alpha \mid z}(x) = 0 , \arange,\zrange ; \;\; w \in \mw(\mathcal P_A ), v =1
		\}
		\end{equation}
		where $\hat f(x), \hat g_{\alpha\mid z}(x)$ are the sample average analogues of the respective integrands $F(x, \xi), G_{\alpha\mid z}(x, \xi)$. For example, $F(x, \xi) = 
		\sum_{b \in \mathcal A_0}	\rho_b
		\left(	\frac{w_a(\hY,Z) \hY}
		{\hpr_{\nm}(A=a) }
		-
		\frac{w_b(\hY,Z) \hY}
		{\hat\pr_{\nm}(A=b )} 
		\right)$, and 
		${G_{\alpha\mid z}(x, \xi) = w_\alpha(\hY,Z)  \ind[{Z=z}]
		-\pr(A=\alpha \mid Z=z) \hpr_\nm(Z=z)v}$.
	}
	
	\edit{Let $Y(w)$ and $ Y_{\alpha\mid z}^{\op{LTP}}(w), ~i=1,\dots, p$ denote zero-mean normally distributed random variables with the same covariance structure as $F(x,\xi)$ and $ G_{\alpha\mid z}(x,\xi),\arange,\zrange$, respectively (since $v=1$). 
		Binding, non-stochastic constraints, such as binding constraints (nonzero Lagrange multipliers) associated with $w \in \mw(\mathcal P_A)$, have degenerate distributional limits and do not contribute to the asymptotic variance. We then show that under certain regularity conditions, the value of the sample problem $\hat h_{\nm,\na}(\rho)$ is  asymptotically normal. 
	}

\edit{
	\begin{proposition}[Inference for demographic disparity linear program estimates.]\label{prop-nuisance-az}
		Under the following conditions: 
		\begin{enumerate}
			\item $\frac{n_{\nm}}{n_{\na}+n_{\nm}} \to r$ (Fixed limiting proportion of auxiliary and primary dataset.)
			\item 
			Asymptotic normality of $\pr(A=\alpha \mid Z=z)$:\label{asn-slow-rate-az} 
			$ n_{\na}^{-\frac 12}(\hpr_{\na}(A =\alpha\mid Z=z) - \pr(A =\alpha\mid Z=z)) \xrightarrow{D} Y_{\alpha \mid z} \text{, where }Y_{\alpha \mid z}\sim N(0, V_{\alpha \mid z}).$
			\item Uniqueness of the dual solution $\overline{\lambda}$.\label{asn-uniqueness} 
			\item $\vert \mathcal Z \vert$ is finite, e.g. $Z$ has finite support.
		\end{enumerate}
The sample estimator converges as: $$
{n_\nm}^{\frac 12} (
\hat{{h}}_{\nm, \na}
(\rho) - {h}(\rho) 
) \xrightarrow{D}
		 Y(w^*) + \sum_{\arange,\zrange} \overline\lambda_{\alpha\mid z}
	(	 Y_{\alpha, z}^{\op{LTP}}(w^*) + Y_{\alpha,z})
		$$
	\end{proposition}
}
\edit{
	\proof{{Proof of \Cref{prop-nuisance-az}}}
	The result follows by the estimation error decomposition, 
	\begin{align*}
	n_{\nm}^{-\frac 12}(	\hat{{h}}_{\nm, \na}
	(\rho)-{{h}}(\rho)
	) = 
	n_{\nm}^{-\frac 12}( 
	\hat{{h}}_{\nm}
(\rho)
	-{{h}}(\rho) )  
	+	n_{\nm}^{-\frac 12}( 
	\hat{{h}}_{\nm, \na}
(\rho)
	-\hat{{h}}_{\nm}
(\rho) ),
	\end{align*} 
	showing each of the above terms converges in distribution, and using that the empirical processes are computed on different datasets (e.g. are independent).
}

	\paragraph{Convergence of $n_{\nm}^{-\frac 12}( \hat{{h}}_{\nm}(\rho)) -{{h}}(\rho) )  $ when $\pr(A\mid Z)$ is known:}

\edit{
	We first state the main theorem, Theorem 5.11 of \cite{shapiro2014lectures}, that we use to obtain asymptotic normality with the sample stochastic program. 
	\begin{theorem}[Theorem 5.11 of \cite{shapiro2014lectures}]\label{thm-constrained-lp-inf}
		Suppose that: the sample is iid, the problem is convex, the set of optimal solutions $\mathcal S $ is nonempty and bounded, $f(w), g_i(w), i = 1, \dots, p$ are finite on a neighborhood of $\mathcal S $, the Slater condition for the true problem holds. For the integrands $F, G_i$, assume that second moments are finite, and that $F, G_i$ are Lipschitz continuous with respect to $w$. 
		Then $n_{\nm}^{\frac 12} (
		\hat{{h}}_{\nm}(\rho) - {{h}(\rho)}
		) 
		\xrightarrow{D} 
		\inf_{w \in \mathcal S} \sup_{\lambda \in \Lambda} 
		[ Y(w) + \sum_{i=1}^p \lambda_k Y_i(w) ]
		$.
		If $\mathcal S = \{ \overline x\}, \Lambda = \{ \overline{\lambda} \}$ are singletons, then 
		$ n_{\nm}^{\frac 12} (
		\hat{{h}}_{\nm}(\rho) - {{h}(\rho)}
		) 
		\xrightarrow{D} 
		N(0, \sigma^2)$
		\text{ where } $\sigma^2 \defeq \op{Var} \left[
		F(\overline x, \xi) + \sum_{i=1}^p \overline{\lambda_i} G_i(\overline{x}, \xi)
		\right]
		$.
	\end{theorem}
}
\edit{
	A direct application of \Cref{thm-constrained-lp-inf} to the form of \Cref{apxeqn: stochopt} yields that
	\begin{equation}\label{eqn-intermed-asyn-norm}
	n_{\nm}^{\frac 12} (
	\hat{{h}}_{\nm}(\rho) - {{h}}(\rho)
	) 
	~~\xrightarrow{D} ~~
	\inf_{W \in \mathcal S} \sup_{\lambda \in \Lambda} 
	[ Y(w) + \sum_{i=1}^p \lambda_k Y_{\alpha\mid z}^{\op{LTP}}(w) ]
	\end{equation}
	with limiting variance $$ \op{Var} \left[
	w^*(\hY, Z)\mathbb{I}[\hY=\hy] + \sum_{\arange,\zrange} 
	\overline{\lambda_{\alpha,z}} (\pr(A=\alpha,Z=z) -\E[w^*_\alpha(\hY,Z) \ind[{Z=z}]] )
	\right]$$
	The sample-optimal value $\hat{{h}}_{\nm}(\mw)$ is asymptotically normal
if the solution is unique. 
}
\paragraph{Convergence of $n_{\nm}^{-\frac 12}( 
	\hat{{h}}_{\nm, \na}
	(\rho) 
	-\hat{{h}}_{\nm}
	(\rho)  )$.}
\edit{
	Since the error from nuisance estimation is a right-hand side perturbation to a linear program, 
	standard sensitivity analysis tools from linear programming admit a first-order expansion of the optimal linear program value with respect to the approximation error. 
	By \cite[Theorem 5.2]{bertsimastsitsiklis}, and in particular under Asn. \eqref{asn-uniqueness}  of this proposition:
	$$  	n_{\nm}^{-\frac 12}(\hat{{h}}_{\nm, \na}
	(\rho) - \hat{{h}}_{\nm} (\rho))=n_{\nm}^{-\frac 12} \sum_{\alpha,z} 
	\overline\lambda_{\alpha, z} (\hpr_{\na}(A=\alpha \mid Z=z) - \mathbb P(A=\alpha \mid Z=z) ). $$
Asn. \eqref{asn-uniqueness} grants that the linear program solution is \textit{unique}. Then, the sensitivity analysis result which shows that the subgradient of the value function (e.g. a piecewise linear function) is given by optimal dual mulitpliers implies that the dual multipliers are indeed a \textit{gradient} of the value function evaluated at $\hat{{h}}_{\nm, \na}$. Asn. \eqref{asn-slow-rate-az} required asymptotic normality of $\hat \eta_{\na}- \eta_{\na}$ such that the approximation error, as a linear combination of the residual terms, satisfies: 
	  $$	\sum_{\arange,\zrange}
	\overline\lambda_{\alpha, z} (\hpr_{\na}(A=\alpha \mid Z=z) - \mathbb P(A=\alpha \mid Z=z) ) ~\xrightarrow{D}  
	\sum_{\arange,\zrange} \overline{\lambda}_{\alpha, z} Y_{\alpha,z}.$$ 
	Lastly, since $\hpr_{\na}(A=\alpha \mid Z=z) - \mathbb P(A=\alpha \mid Z=z) $ is an independent empirical process: it is evaluated with respect to a different dataset, so that the sum of two random variables converging in distribution to normal random variables converges to the sum of their limits. 
}
\endproof

\subsection{Proof for \cref{thm: asymp-dist-tprd} and \cref{corollary: CI-tprd}}
\edit{
\proof{Proof for \cref{thm: asymp-dist-tprd}.}
We can verify that 
\begin{align*}
&\sqrt{\npri}(\hat{\mu}'_{\hy y}(\alpha, \tilde{w}^L, \tilde{w}^U) - {\mu}'_{\hy y}(\alpha, \tilde{w}^L, \tilde{w}^U)) \\
=& \frac{{\mu}'_{1 - \hy, y}(\alpha; \tw^U,\tw^L)}{\hat{\overline{w}}^L_{\alpha}(\hy, y)  + \hat{\overline{w}}^U_{\alpha}(1 - \hy, y)}
\sqrt{\npri}\left(\hat{\overline{w}}^L_{\alpha}(\hy, y) - {\overline{w}}^L_{\alpha}(\hy, y)\right)
- \frac{{\mu}'_{\hy y}(\alpha; \tw^L,\tw^U)}{\hat{\overline{w}}^L_{\alpha}(\hy, y)  + \hat{\overline{w}}^U_{\alpha}(1 - \hy, y)}
\sqrt{\npri}\left(\hat{\overline{w}}^U_{\alpha}(1 - \hy, y) - {\overline{w}}^U_{\alpha}(1 - \hy, y)\right)
\end{align*}
By following the proof of \Cref{thm: asymp-dist-dd}, we can prove analogously that under conditions (i)-(iv),
\begin{align*}
&\sqrt{\npri}\left(\hat{\overline{w}}^L_{\alpha}(\hy, y) - {\overline{w}}^L_{\alpha}(\hy, y)\right) \\
=& \sqrt{\npri}\left\{\expectn\left[\tilde{\lambda}^L_{\alpha, \hy y}(Z; \tilde\eta) - {\overline{w}}^L_{\alpha}(\hy, y)\right] + \expectna\left[\tilde{\xi}^L_{\alpha, \hy y}(A, Z; \tilde\eta)\right] + \expectnp\left[\tilde{\gamma}^L_{\alpha, \hy y}(\hY, Y, Z; \tilde\eta)\right]\right\} + o_p(1), \\
&\sqrt{\npri}\left(\hat{\overline{w}}^U_{\alpha}(1 - \hy, y) - {\overline{w}}^U_{\alpha}(1 - \hy, y)\right) \\
=& \sqrt{\npri}\left\{\expectn\left[\tilde{\lambda}^U_{\alpha, 1 - \hy, y}(Z; \tilde\eta) - {\overline{w}}^U_{\alpha}(1 - \hy, y)\right] + \expectna\left[\tilde{\xi}^U_{\alpha, 1 - \hy, y}(A, Z; \tilde\eta)\right] + \expectnp\left[\tilde{\gamma}^U_{\alpha, 1 - \hy, y}(\hY, Y, Z; \tilde\eta)\right]\right\} + o_p(1).
\end{align*}
This means that 
\begin{align*}
&\sqrt{\npri}(\hat{\mu}'_{\hy y}(\alpha, \tilde{w}^L, \tilde{w}^U) - {\mu}'_{\hy y}(\alpha, \tilde{w}^L, \tilde{w}^U)) \\
=& \frac{{\mu}'_{1 - \hy, y}(\alpha; \tw^U,\tw^L)}{{\overline{w}}^L_{\alpha}(\hy, y)  + {\overline{w}}^U_{\alpha}(1 - \hy, y)} \sqrt{\npri}\left\{\expectn\left[\tilde{\lambda}^L_{\alpha, \hy y}(Z; \tilde\eta) - {\overline{w}}^L_{\alpha}(\hy, y)\right] + \expectna\left[\tilde{\xi}^L_{\alpha, \hy y}(A, Z; \tilde\eta)\right] + \expectnp\left[\tilde{\gamma}^L_{\alpha, \hy y}(\hY, Y, Z; \tilde\eta)\right]\right\} \\
-& \frac{{\mu}'_{\hy y}(\alpha; \tw^L,\tw^U)}{{\overline{w}}^L_{\alpha}(\hy, y)  + {\overline{w}}^U_{\alpha}(1 - \hy, y)}\sqrt{\npri}\bigg\{\expectn\left[\tilde{\lambda}^U_{\alpha, 1 - \hy, y}(Z; \tilde\eta) - {\overline{w}}^U_{\alpha}(1 - \hy, y)\right] + \expectna\left[\tilde{\xi}^U_{\alpha, 1 - \hy, y}(A, Z; \tilde\eta)\right] \\
&\qquad\qquad\qquad\qquad\qquad\qquad\qquad\qquad\qquad\qquad\qquad\qquad\qquad\qquad + \expectnp\left[\tilde{\gamma}^U_{\alpha, 1 - \hy, y}(\hY, Y, Z; \tilde\eta)\right]\bigg\} + o_p(1).
\end{align*}
It follows that 
\begin{align*}
&\sqrt{\npri}(\hat{\mu}'_{\hy y}(\alpha, \tilde{w}^L, \tilde{w}^U) - {\mu}'_{\hy y}(\alpha, \tilde{w}^L, \tilde{w}^U)) \\
=&\sqrt{\npri}\expectnp\bigg\{r_n\left[\frac{{\mu}'_{1 - \hy, y}(\alpha; \tw^U,\tw^L)}{{\overline{w}}^L_{\alpha}(\hy, y)  + {\overline{w}}^U_{\alpha}(1 - \hy, y)}\tilde{\lambda}^L_{\alpha, \hy y}(Z; \tilde\eta) - \frac{{\mu}'_{\hy y}(\alpha; \tw^L,\tw^U)}{{\overline{w}}^L_{\alpha}(\hy, y)  + {\overline{w}}^U_{\alpha}(1 - \hy, y)}\tilde{\lambda}^U_{\alpha, 1 - \hy, y}(Z; \tilde\eta)\right]\bigg\} \\
+&\sqrt{\npri}\expectnp\bigg\{\frac{{\mu}'_{1 - \hy, y}(\alpha; \tw^U,\tw^L)}{{\overline{w}}^L_{\alpha}(\hy, y)  + {\overline{w}}^U_{\alpha}(1 - \hy, y)}\tilde{\gamma}^L_{\alpha, \hy y}(\hY, Y, Z; \tilde\eta) - \frac{{\mu}'_{\hy y}(\alpha; \tw^L,\tw^U)}{{\overline{w}}^L_{\alpha}(\hy, y)  + {\overline{w}}^U_{\alpha}(1 - \hy, y)}\tilde{\gamma}^L_{\alpha, 1 -\hy, y}(\hY, Y, Z; \tilde\eta)\bigg\} \\
+&\sqrt{\npri}\expectna\bigg\{(1 - r_n)\left[\frac{{\mu}'_{1 - \hy, y}(\alpha; \tw^U,\tw^L)}{{\overline{w}}^L_{\alpha}(\hy, y)  + {\overline{w}}^U_{\alpha}(1 - \hy, y)}\tilde{\lambda}^L_{\alpha, \hy y}(Z; \tilde\eta) - \frac{{\mu}'_{\hy y}(\alpha; \tw^L,\tw^U)}{{\overline{w}}^L_{\alpha}(\hy, y)  + {\overline{w}}^U_{\alpha}(1 - \hy, y)}\tilde{\lambda}^U_{\alpha, 1 - \hy, y}(Z; \tilde\eta)\right]\bigg\} \\
+& \sqrt{\npri}\expectna\bigg\{\frac{{\mu}'_{1 - \hy, y}(\alpha; \tw^U,\tw^L)}{{\overline{w}}^L_{\alpha}(\hy, y)  + {\overline{w}}^U_{\alpha}(1 - \hy, y)}\tilde{\xi}^L_{\alpha, \hy y}(A, Z; \tilde\eta) - \frac{{\mu}'_{\hy y}(\alpha; \tw^L,\tw^U)}{{\overline{w}}^L_{\alpha}(\hy, y)  + {\overline{w}}^U_{\alpha}(1 - \hy, y)}\tilde{\xi}^L_{\alpha, 1 -\hy, y}(A, Z; \tilde\eta)\bigg\} + o_p(1).
\end{align*}
Similarly 
\begin{align*}
&\sqrt{\npri}(\hat{\mu}'_{\hy y}(\alpha, \tilde{w}^U, \tilde{w}^L) - {\mu}'_{\hy y}(\alpha, \tilde{w}^U, \tilde{w}^L)) \\
=&\sqrt{\npri}\expectnp\bigg\{r_n\left[\frac{{\mu}'_{1 - \hy, y}(\alpha; \tw^L,\tw^U)}{{\overline{w}}^U_{\alpha}(\hy, y)  + {\overline{w}}^L_{\alpha}(1 - \hy, y)}\tilde{\lambda}^U_{\alpha, \hy y}(Z; \tilde\eta) - \frac{{\mu}'_{\hy y}(\alpha; \tw^U,\tw^L)}{{\overline{w}}^U_{\alpha}(\hy, y)  + {\overline{w}}^L_{\alpha}(1 - \hy, y)}\tilde{\lambda}^L_{\alpha, 1 - \hy, y}(Z; \tilde\eta)\right]\bigg\} \\
+&\sqrt{\npri}\expectnp\bigg\{\frac{{\mu}'_{1 - \hy, y}(\alpha; \tw^L,\tw^U)}{{\overline{w}}^U_{\alpha}(\hy, y)  + {\overline{w}}^L_{\alpha}(1 - \hy, y)}\tilde{\gamma}^U_{\alpha, \hy y}(\hY, Y, Z; \tilde\eta) - \frac{{\mu}'_{\hy y}(\alpha; \tw^U,\tw^L)}{{\overline{w}}^U_{\alpha}(\hy, y)  + {\overline{w}}^L_{\alpha}(1 - \hy, y)}\tilde{\gamma}^U_{\alpha, 1 -\hy, y}(\hY, Y, Z; \tilde\eta)\bigg\} \\
+&\sqrt{\npri}\expectna\bigg\{(1 - r_n)\left[\frac{{\mu}'_{1 - \hy, y}(\alpha; \tw^L,\tw^U)}{{\overline{w}}^U_{\alpha}(\hy, y)  + {\overline{w}}^L_{\alpha}(1 - \hy, y)}\tilde{\lambda}^U_{\alpha, \hy y}(Z; \tilde\eta) - \frac{{\mu}'_{\hy y}(\alpha; \tw^U,\tw^L)}{{\overline{w}}^U_{\alpha}(\hy, y)  + {\overline{w}}^L_{\alpha}(1 - \hy, y)}\tilde{\lambda}^L_{\alpha, 1 - \hy, y}(Z; \tilde\eta)\right]\bigg\} \\
+& \sqrt{\npri}\expectna\bigg\{\frac{{\mu}'_{1 - \hy, y}(\alpha; \tw^L,\tw^U)}{{\overline{w}}^U_{\alpha}(\hy, y)  + {\overline{w}}^L_{\alpha}(1 - \hy, y)}\tilde{\xi}^U_{\alpha, \hy y}(A, Z; \tilde\eta) - \frac{{\mu}'_{\hy y}(\alpha; \tw^U,\tw^L)}{{\overline{w}}^U_{\alpha}(\hy, y)  + {\overline{w}}^L_{\alpha}(1 - \hy, y)}\tilde{\xi}^U_{\alpha, 1 -\hy, y}(A, Z; \tilde\eta)\bigg\} + o_p(1).
\end{align*}
Therefore, the lower bound estimator $\hat{\mu}'_{\hy y}(a, \tilde{w}^L, \tilde{w}^U)  - \hat{\mu}'_{\hy y}(b, \tilde{w}^U, \tilde{w}^L)$ satisfies the following:
\begin{align*}
&\sqrt{\npri}\left[\hat{\mu}'_{\hy y}(a, \tilde{w}^L, \tilde{w}^U)  - \hat{\mu}'_{\hy y}(b, \tilde{w}^U, \tilde{w}^L) - \left({\mu}'_{\hy y}(a, \tilde{w}^L, \tilde{w}^U) - {\mu}'_{\hy y}(b, \tilde{w}^U, \tilde{w}^L\right)\right] \\
=&\sqrt{\npri}\expectnp\bigg\{r_n\left[\frac{{\mu}'_{1 - \hy, y}(a; \tw^U,\tw^L)}{{\overline{w}}^L_{a}(\hy, y)  + {\overline{w}}^U_{a}(1 - \hy, y)}\tilde{\lambda}^L_{a, \hy y}(Z; \tilde\eta) - \frac{{\mu}'_{\hy y}(a; \tw^L,\tw^U)}{{\overline{w}}^L_{a}(\hy, y)  + {\overline{w}}^U_{a}(1 - \hy, y)}\tilde{\lambda}^U_{a, 1 - \hy, y}(Z; \tilde\eta)\right]\\
-&\qquad\qquad r_n\left[\frac{{\mu}'_{1 - \hy, y}(b; \tw^L,\tw^U)}{{\overline{w}}^U_{b}(\hy, y)  + {\overline{w}}^L_{b}(1 - \hy, y)}\tilde{\lambda}^U_{b, \hy y}(Z; \tilde\eta) - \frac{{\mu}'_{\hy y}(b; \tw^U,\tw^L)}{{\overline{w}}^U_{b}(\hy, y)  + {\overline{w}}^L_{b}(1 - \hy, y)}\tilde{\lambda}^L_{b, 1 - \hy, y}(Z; \tilde\eta)\right]\bigg\} \\
+&\sqrt{\npri}\expectnp\bigg\{\frac{{\mu}'_{1 - \hy, y}(a; \tw^U,\tw^L)}{{\overline{w}}^L_{a}(\hy, y)  + {\overline{w}}^U_{a}(1 - \hy, y)}\tilde{\gamma}^L_{a, \hy y}(\hY, Y, Z; \tilde\eta) - \frac{{\mu}'_{\hy y}(a; \tw^L,\tw^U)}{{\overline{w}}^L_{a}(\hy, y)  + {\overline{w}}^U_{a}(1 - \hy, y)}\tilde{\gamma}^L_{a, 1 -\hy, y}(\hY, Y, Z; \tilde\eta)\\
-&\qquad\qquad \left[\frac{{\mu}'_{1 - \hy, y}(b; \tw^L,\tw^U)}{{\overline{w}}^U_{b}(\hy, y)  + {\overline{w}}^L_{b}(1 - \hy, y)}\tilde{\gamma}^U_{b, \hy y}(\hY, Y, Z; \tilde\eta) - \frac{{\mu}'_{\hy y}(b; \tw^U,\tw^L)}{{\overline{w}}^U_{b}(\hy, y)  + {\overline{w}}^L_{b}(1 - \hy, y)}\tilde{\gamma}^U_{b, 1 -\hy, y}(\hY, Y, Z; \tilde\eta)\right]\bigg\} \\
+&\sqrt{\npri}\expectna\bigg\{(1 - r_n)\left[\frac{{\mu}'_{1 - \hy, y}(a; \tw^U,\tw^L)}{{\overline{w}}^L_{a}(\hy, y)  + {\overline{w}}^U_{a}(1 - \hy, y)}\tilde{\lambda}^L_{a, \hy y}(Z; \tilde\eta) - \frac{{\mu}'_{\hy y}(a; \tw^L,\tw^U)}{{\overline{w}}^L_{a}(\hy, y)  + {\overline{w}}^U_{a}(1 - \hy, y)}\tilde{\lambda}^U_{a, 1 - \hy, y}(Z; \tilde\eta)\right] \\
-&\qquad\qquad (1 - r_n)\left[\frac{{\mu}'_{1 - \hy, y}(b; \tw^L,\tw^U)}{{\overline{w}}^U_{b}(\hy, y)  + {\overline{w}}^L_{b}(1 - \hy, y)}\tilde{\lambda}^U_{b, \hy y}(Z; \tilde\eta) - \frac{{\mu}'_{\hy y}(b; \tw^U,\tw^L)}{{\overline{w}}^U_{b}(\hy, y)  + {\overline{w}}^L_{b}(1 - \hy, y)}\tilde{\lambda}^L_{b, 1 - \hy, y}(Z; \tilde\eta)\right]\bigg\} \\
+& \sqrt{\npri}\expectna\bigg\{\frac{{\mu}'_{1 - \hy, y}(a; \tw^U,\tw^L)}{{\overline{w}}^L_{a}(\hy, y)  + {\overline{w}}^U_{a}(1 - \hy, y)}\tilde{\xi}^L_{a, \hy y}(A, Z; \tilde\eta) - \frac{{\mu}'_{\hy y}(a; \tw^L,\tw^U)}{{\overline{w}}^L_{a}(\hy, y)  + {\overline{w}}^U_{a}(1 - \hy, y)}\tilde{\xi}^L_{a, 1 -\hy, y}(A, Z; \tilde\eta) \\
-&\qquad\qquad \left[\frac{{\mu}'_{1 - \hy, y}(b; \tw^L,\tw^U)}{{\overline{w}}^U_{b}(\hy, y)  + {\overline{w}}^L_{b}(1 - \hy, y)}\tilde{\xi}^U_{b, \hy y}(A, Z; \tilde\eta) - \frac{{\mu}'_{\hy y}(b; \tw^U,\tw^L)}{{\overline{w}}^U_{b}(\hy, y)  + {\overline{w}}^L_{b}(1 - \hy, y)}\tilde{\xi}^U_{b, 1 -\hy, y}(A, Z; \tilde\eta)\right]\bigg\} + o_p(1).
\end{align*}
By Central Limit Theorem, 
\[
	\sqrt{\npri}\left[\hat{\mu}'_{\hy y}(a, \tilde{w}^L, \tilde{w}^U)  - \hat{\mu}'_{\hy y}(b, \tilde{w}^U, \tilde{w}^L) - \left({\mu}'_{\hy y}(a, \tilde{w}^L, \tilde{w}^U) - {\mu}'_{\hy y}(b, \tilde{w}^U, \tilde{w}^L\right)\right] \overset{d}{\to} \mathcal{N}(0, \tilde{V}_L)
\]
where 
\begin{align*}
\tilde{V}_L 
	&= r\expect\bigg\{\left[\frac{{\mu}'_{1 - \hy, y}(a; \tw^U,\tw^L)}{{\overline{w}}^L_{a}(\hy, y)  + {\overline{w}}^U_{a}(1 - \hy, y)}\tilde{\lambda}^L_{a, \hy y}(Z; \tilde\eta) - \frac{{\mu}'_{\hy y}(a; \tw^L,\tw^U)}{{\overline{w}}^L_{a}(\hy, y)  + {\overline{w}}^U_{a}(1 - \hy, y)}\tilde{\lambda}^U_{a, 1 - \hy, y}(Z; \tilde\eta)\right]\\
	&\qquad\qquad - \left[\frac{{\mu}'_{1 - \hy, y}(b; \tw^L,\tw^U)}{{\overline{w}}^U_{b}(\hy, y)  + {\overline{w}}^L_{b}(1 - \hy, y)}\tilde{\lambda}^U_{b, \hy y}(Z; \tilde\eta) - \frac{{\mu}'_{\hy y}(b; \tw^U,\tw^L)}{{\overline{w}}^U_{b}(\hy, y)  + {\overline{w}}^L_{b}(1 - \hy, y)}\tilde{\lambda}^L_{b, 1 - \hy, y}(Z; \tilde\eta)\right]\bigg\}^2 \\
	&+ \expect\bigg\{\frac{{\mu}'_{1 - \hy, y}(a; \tw^U,\tw^L)}{{\overline{w}}^L_{a}(\hy, y)  + {\overline{w}}^U_{a}(1 - \hy, y)}\tilde{\gamma}^L_{a, \hy y}(\hY, Y, Z; \tilde\eta) - \frac{{\mu}'_{\hy y}(a; \tw^L,\tw^U)}{{\overline{w}}^L_{a}(\hy, y)  + {\overline{w}}^U_{a}(1 - \hy, y)}\tilde{\gamma}^L_{a, 1 -\hy, y}(\hY, Y, Z; \tilde\eta)\\
	&\qquad\qquad - \left[\frac{{\mu}'_{1 - \hy, y}(b; \tw^L,\tw^U)}{{\overline{w}}^U_{b}(\hy, y)  + {\overline{w}}^L_{b}(1 - \hy, y)}\tilde{\gamma}^U_{b, \hy y}(\hY, Y, Z; \tilde\eta) - \frac{{\mu}'_{\hy y}(b; \tw^U,\tw^L)}{{\overline{w}}^U_{b}(\hy, y)  + {\overline{w}}^L_{b}(1 - \hy, y)}\tilde{\gamma}^U_{b, 1 -\hy, y}(\hY, Y, Z; \tilde\eta)\right]\bigg\}^2 \\ 
	&+ \frac{r}{1 - r}\expect\bigg\{\frac{{\mu}'_{1 - \hy, y}(a; \tw^U,\tw^L)}{{\overline{w}}^L_{a}(\hy, y)  + {\overline{w}}^U_{a}(1 - \hy, y)}\tilde{\xi}^L_{a, \hy y}(A, Z; \tilde\eta) - \frac{{\mu}'_{\hy y}(a; \tw^L,\tw^U)}{{\overline{w}}^L_{a}(\hy, y)  + {\overline{w}}^U_{a}(1 - \hy, y)}\tilde{\xi}^L_{a, 1 -\hy, y}(A, Z; \tilde\eta) \\
	&\qquad\qquad -\left[\frac{{\mu}'_{1 - \hy, y}(b; \tw^L,\tw^U)}{{\overline{w}}^U_{b}(\hy, y)  + {\overline{w}}^L_{b}(1 - \hy, y)}\tilde{\xi}^U_{b, \hy y}(A, Z; \tilde\eta) - \frac{{\mu}'_{\hy y}(b; \tw^U,\tw^L)}{{\overline{w}}^U_{b}(\hy, y)  + {\overline{w}}^L_{b}(1 - \hy, y)}\tilde{\xi}^U_{b, 1 -\hy, y}(A, Z; \tilde\eta)\right]\bigg\}^2.
\end{align*}
Similarly, we can prove that the upper bound estimator $\hat{\mu}'_{\hy y}(a, \tilde{w}^U, \tilde{w}^L)  - \hat{\mu}'_{\hy y}(b, \tilde{w}^L, \tilde{w}^U)$ satisfies that 
\[
	\sqrt{\npri}\left[\hat{\mu}'_{\hy y}(a, \tilde{w}^U, \tilde{w}^L)  - \hat{\mu}'_{\hy y}(b, \tilde{w}^L, \tilde{w}^U) - \left({\mu}'_{\hy y}(a, \tilde{w}^U, \tilde{w}^L)  - {\mu}'_{\hy y}(b, \tilde{w}^L, \tilde{w}^U)\right)\right] \overset{d}{\to} \mathcal{N}(0, \tilde{V}_U)
\]
where 
\begin{align*}
\tilde{V}_U 
	&= r\expect\bigg\{\left[\frac{{\mu}'_{1 - \hy, y}(b; \tw^L,\tw^U)}{{\overline{w}}^U_{b}(\hy, y)  + {\overline{w}}^L_{b}(1 - \hy, y)}\tilde{\lambda}^U_{b, \hy y}(Z; \tilde\eta) - \frac{{\mu}'_{\hy y}(b; \tw^U,\tw^L)}{{\overline{w}}^U_{b}(\hy, y)  + {\overline{w}}^L_{b}(1 - \hy, y)}\tilde{\lambda}^L_{b, 1 - \hy, y}(Z; \tilde\eta)\right]\\
	&\qquad\qquad - \left[\frac{{\mu}'_{1 - \hy, y}(a; \tw^U,\tw^L)}{{\overline{w}}^L_{a}(\hy, y)  + {\overline{w}}^U_{a}(1 - \hy, y)}\tilde{\lambda}^L_{a, \hy y}(Z; \tilde\eta) - \frac{{\mu}'_{\hy y}(a; \tw^L,\tw^U)}{{\overline{w}}^L_{a}(\hy, y)  + {\overline{w}}^U_{a}(1 - \hy, y)}\tilde{\lambda}^U_{a, 1 - \hy, y}(Z; \tilde\eta)\right]\bigg\}^2 \\
	&+ \expect\bigg\{\frac{{\mu}'_{1 - \hy, y}(b; \tw^L,\tw^U)}{{\overline{w}}^U_{b}(\hy, y)  + {\overline{w}}^L_{b}(1 - \hy, y)}\tilde{\gamma}^U_{b, \hy y}(\hY, Y, Z; \tilde\eta) - \frac{{\mu}'_{\hy y}(b; \tw^U,\tw^L)}{{\overline{w}}^U_{b}(\hy, y)  + {\overline{w}}^L_{b}(1 - \hy, y)}\tilde{\gamma}^U_{b, 1 -\hy, y}(\hY, Y, Z; \tilde\eta)\\
	&\qquad\qquad - \left[\frac{{\mu}'_{1 - \hy, y}(a; \tw^U,\tw^L)}{{\overline{w}}^L_{a}(\hy, y)  + {\overline{w}}^U_{a}(1 - \hy, y)}\tilde{\gamma}^L_{a, \hy y}(\hY, Y, Z; \tilde\eta) - \frac{{\mu}'_{\hy y}(a; \tw^L,\tw^U)}{{\overline{w}}^L_{a}(\hy, y)  + {\overline{w}}^U_{a}(1 - \hy, y)}\tilde{\gamma}^L_{a, 1 -\hy, y}(\hY, Y, Z; \tilde\eta)\right]\bigg\}^2 \\ 
	&+ \frac{r}{1 - r}\expect\bigg\{\frac{{\mu}'_{1 - \hy, y}(b; \tw^L,\tw^U)}{{\overline{w}}^U_{b}(\hy, y)  + {\overline{w}}^L_{b}(1 - \hy, y)}\tilde{\xi}^U_{b, \hy y}(A, Z; \tilde\eta) - \frac{{\mu}'_{\hy y}(b; \tw^U,\tw^L)}{{\overline{w}}^U_{b}(\hy, y)  + {\overline{w}}^L_{b}(1 - \hy, y)}\tilde{\xi}^U_{b, 1 -\hy, y}(A, Z; \tilde\eta) \\
	&\qquad\qquad -\left[\frac{{\mu}'_{1 - \hy, y}(a; \tw^U,\tw^L)}{{\overline{w}}^L_{a}(\hy, y)  + {\overline{w}}^U_{a}(1 - \hy, y)}\tilde{\xi}^L_{a, \hy y}(A, Z; \tilde\eta) - \frac{{\mu}'_{\hy y}(a; \tw^L,\tw^U)}{{\overline{w}}^L_{a}(\hy, y)  + {\overline{w}}^U_{a}(1 - \hy, y)}\tilde{\xi}^L_{a, 1 -\hy, y}(A, Z; \tilde\eta)\right]\bigg\}^2.
\end{align*}
\endproof
}

\edit{
\proof{Proof for \cref{corollary: CI-tprd}}
We can prove this corollary by following procedures in the proof of \cref{corollary: CI-dd}.
\endproof
}

\subsection{Proof for \cref{thm: tprd-dd-known-dist}}
\proof{Proof for \cref{thm: tprd-dd-known-dist}.}
\begin{align*}
&\sqrt{\npri}(\hat{\mu}(\alpha, w^L) - {\mu}(\alpha, w^L)) \\
=& \frac{1}{\expect\left[\etaa(\alpha, Z)\right]}\sqrt{\npri}\bigg(\expectnp\left[\ind\left({\etaph}(1, Z) + \etaa(\alpha, Z) -1  \ge 0 \right)\left(\hY + \etaa(\alpha, Z) - 1\right)\right] \\
& \qquad\qquad\qquad\qquad\qquad -\expect\left[\ind\left({\etap}(1, Z) + \etaa(\alpha, Z) -1  \ge 0 \right)\left(\etap(1, Z) + \etaa(\alpha, Z) - 1\right)\right]\bigg) + o_p(1) \\
=& \frac{1}{\expect\left[\etaa(\alpha, Z)\right]}\sqrt{\npri}\bigg(\expectnp\big[\ind\left({\etaph}(1, Z) + \etaa(\alpha, Z) -1  \ge 0 \right)\left(\hY + \etaa(\alpha, Z) - 1\right) \\
	&\qquad\qquad\qquad\qquad\qquad - \ind\left({\etap}(1, Z) + \etaa(\alpha, Z) -1  \ge 0 \right)\left(\etap(1, Z) + \etaa(\alpha, Z) - 1\right)\big]\bigg) + o_p(1) \\
+& \frac{1}{\expect\left[\etaa(\alpha, Z)\right]}\sqrt{\npri}\bigg(\expectnp\left[\ind\left({\etap}(1, Z) + \etaa(\alpha, Z) -1  \ge 0 \right)\left(\etap(1, Z) + \etaa(\alpha, Z) - 1\right)\right] \\
&\qquad\qquad\qquad\qquad\qquad  -\expect\left[\ind\left({\etap}(1, Z) + \etaa(\alpha, Z) -1  \ge 0 \right)\left(\etap(1, Z) + \etaa(\alpha, Z) - 1\right)\right]\bigg).
\end{align*}
Note that 
\begin{align*}
&\sqrt{\npri}\bigg(\expectnp\big[\ind\left({\etaph}(1, Z) + \etaa(\alpha, Z) -1  \ge 0 \right)\left(\hY + \etaa(\alpha, Z) - 1\right) \\
&\qquad\qquad \qquad  - \ind\left({\etap}(1, Z) + \etaa(\alpha, Z) -1  \ge 0 \right)\left(\etap(1, Z) + \etaa(\alpha, Z) - 1\right)\big]\bigg) \\
=& \sqrt{\npri}\bigg(\expectnp\left[\left(\ind\left({\etaph}(1, Z) + \etaa(\alpha, Z) -1  \ge 0 \right) - \ind\left({\etap}(1, Z) + \etaa(\alpha, Z) -1  \ge 0 \right)\right)\left(\etap(1, Z) + \etaa(\alpha, Z) - 1\right)\right]\bigg) \\
+&\sqrt{\npri}\bigg(\expectnp\left[\left(\ind\left({\etaph}(1, Z) + \etaa(\alpha, Z) -1  \ge 0 \right) - \ind\left({\etap}(1, Z) + \etaa(\alpha, Z) -1  \ge 0 \right)\right) \left(\hY - {\etap}(1, Z)\right)\right]\bigg) \\
+& \sqrt{\npri}\expectnp \left[\ind\left({\etap}(1, Z) + \etaa(\alpha, Z) -1  \ge 0 \right)\left(\hY - {\etap}(1, Z)\right)\right] \\
=&\sqrt{\npri}\expectnp \left[\ind\left({\etap}(1, Z) + \etaa(\alpha, Z) -1  \ge 0 \right)\left(\hY - {\etap}(1, Z)\right)\right] + o_p(1),
\end{align*}
where the last equation can be verified by following the step II in the proof of \Cref{thm: asymp-dist-dd} and the conditions 2 to 4 in \cref{thm: tprd-dd-known-dist}.

Therefore, 
\begin{align*}
&\sqrt{\npri}(\hat{\mu}(\alpha, w^L) - {\mu}(\alpha, w^L)) \\
=& \sqrt{\npri}\expectnp\bigg[\ind\left({\etap}(1, Z) + \etaa(\alpha, Z) -1  \ge 0 \right)\left(\eta_1(Z) + \etaa(\alpha, Z) - 1\right)/\expect\left[\etaa(\alpha, Z)\right] - {\mu}(\alpha, w^L)\bigg] \\
+&\qquad\qquad\qquad\qquad\qquad\qquad \sqrt{n}_p\expectnp \left[\ind\left({\etap}(1, Z) + \etaa(\alpha, Z) -1  \ge 0 \right)\left(\hY - {\etap}(1, Z)\right)/\expect\left[\etaa(\alpha, Z)\right]\right] + o_p(1) \\
=& \sqrt{\npri}\expectnp\bigg[\ind\left({\etap}(1, Z) + \etaa(\alpha, Z) -1  \ge 0 \right)\left(\hat{Y} + \etaa(\alpha, Z) - 1\right)/\expect\left[\etaa(\alpha, Z)\right] - {\mu}(\alpha, w^L)\bigg].
\end{align*}
Similarly, we can prove that 
\begin{align*}
&\sqrt{\npri}(\hat{\mu}(\alpha, w^U) - {\mu}(\alpha, w^U)) \\
=& \sqrt{\npri}\expectnp\bigg[\left[\ind\left({\etap}(1, Z) - \etaa(\alpha, Z)  \le 0 \right)\left(\etap(1, Z) - \etaa(\alpha, Z)\right) + \etaa(\alpha, Z)\right]/\expect\left[\etaa(\alpha, Z)\right] - {\mu}(\alpha, w^U)\bigg] \\
+&\qquad\qquad\qquad\qquad\qquad\qquad\qquad\qquad  \sqrt{n}_p\expectnp \left[\ind\left({\etap}(1, Z) - \etaa(\alpha, Z)  \le 0 \right)\left(\hY - {\etap}(1, Z)\right)/\expect\left[\etaa(\alpha, Z)\right]\right] + o_p(1) \\
=& \sqrt{\npri}\expectnp\bigg[\left[\ind\left({\etap}(1, Z) - \etaa(\alpha, Z)  \le 0 \right)\left(\hY - \etaa(\alpha, Z)\right) + \etaa(\alpha, Z)\right]/\expect\left[\etaa(\alpha, Z)\right] - {\mu}(\alpha, w^U)\bigg] + o_p(1).
\end{align*}
Therefore, 
\begin{align*}
&\sqrt{\npri}\left[\hat{\mu}(\alpha, w^L) - \hat{\mu}(\alpha, w^U) - \left({\mu}(\alpha, w^L) - {\mu}(\alpha, w^U)\right)\right] \\
=& \sqrt{\npri}\expectnp\bigg[\ind\left({\etap}(1, Z) + \etaa(a, Z) -1  \ge 0 \right)\left(\hY + \etaa(a, Z) - 1\right)/\expect\left[\etaa(a, Z)\right] \\
-&\qquad\qquad \left[\ind\left({\etap}(1, Z) - \etaa(b, Z)  \le 0 \right)\left(\hY - \etaa(b, Z)\right) + \etaa(b, Z)\right]/\expect\left[\etaa(b, Z)\right] - \left({\mu}(a, w^L)  - {\mu}(b, w^U)\right)\bigg] + o_p(1)\\
\to&   \mathcal{N}(0, V_L),
\end{align*}
where 
\begin{align*}
V_L 
=& \expect\bigg[\ind\left({\etap}(1, Z) + \etaa(a,Z) -1  \ge 0 \right)\left(\hY + \etaa(a,Z) - 1\right)/\expect\left[\etaa(a,Z)\right] \\
-&\qquad\qquad \left[\ind\left({\etap}(1, Z) - \etaa(b, Z)  \le 0 \right)\left(\hY - \etaa(b, Z)\right) + \etaa(b, Z)\right]/\expect\left[\etaa(b, Z)\right] - \left({\mu}(a, w^L)  - {\mu}(b, w^U)\right)\bigg]^2. \\
\end{align*}

Similarly, 
\begin{align*}
&\sqrt{\npri}\left[\hat{\mu}(\alpha, w^U) - \hat{\mu}(\alpha, w^L) - \left({\mu}(\alpha, w^U) - {\mu}(\alpha, w^L)\right)\right] \\
=& \sqrt{\npri}\expectnp\bigg[\left[\ind\left({\etap}(1, Z) - \etaa(a,Z)  \le 0 \right)\left(\hY - \etaa(a,Z) \right) + \etaa(a, Z)\right]/\expect\left[\etaa(a,Z)\right] \\
-&\qquad\qquad\qquad \ind\left({\etap}(1, Z) + \etaa(b, Z) - 1  \ge 0 \right)\left(\hY + \etaa(b, Z) - 1\right)/\expect\left[\etaa(b, Z)\right] - \left({\mu}(a, w^U)  - {\mu}(b, w^L)\right)\bigg] \\
+& o_p(1) \to \mathcal{N}(0, V_U),
\end{align*}
where 
\begin{align*}
V_U 
	=& \expect\bigg[\left[\ind\left({\etap}(1, Z) - \etaa(a,Z)  \le 0 \right)\left(\hY - \etaa(a,Z) \right) + \etaa(a, Z)\right]/\expect\left[\etaa(a,Z)\right] \\
	-&\qquad\qquad\qquad \ind\left({\etap}(1, Z) + \etaa(b, Z) - 1  \ge 0 \right)\left(\hY + \etaa(b, Z) - 1\right)/\expect\left[\etaa(b, Z)\right] - \left({\mu}(a, w^U)  - {\mu}(b, w^L)\right)\bigg]^2. \\
\end{align*}
\endproof

\subsection{Proof for \cref{sec: cal-CI}}
\proof{Proof for \cref{corollary: asymp-dist-dd-extended}}
According to the proof of \cref{thm: asymp-dist-dd}, 
\begin{align*}
&\sqrt{\npri}\left[\hat{\mu}(a, w^L) - \hat{\mu}(b, w^U) - \left({\mu}(a, w^L) - {\mu}(b, w^U)\right)\right]\\
=& \sqrt{\npri}\expectna\left\{(1 - r)\left[\lambda_{a}^L(Z; \eta)/p_a - \lambda_{b}^U(Z; \eta)/p_b - \left({\mu}(a, w^L) - {\mu}(b, w^U)\right)\right] + \xi_{a}^L(A, Z; \eta)/p_a - \xi_{b}^U(A, Z; \eta)/p_b\right\} \\
+& \sqrt{\npri}\expectnp\left\{r\left[\lambda_{a}^L(Z; \eta)/p_a - \lambda_{b}^U(Z; \eta)/p_b - \left({\mu}(a, w^L) - {\mu}(b, w^U)\right)\right] + \gamma_{a}^L(\hY, Z; \eta)/p_a - \gamma_{b}^U(\hY, Z; \eta)/p_b\right\} + o_p(1), \\
&\sqrt{\npri}\left[\hat{\mu}(a, w^U) - \hat{\mu}(b, w^L) - \left({\mu}(a, w^U) - {\mu}(b, w^L)\right)\right]\\
=& \sqrt{\npri}\expectna\left\{(1 - r)\left[\lambda_{a}^U(Z; \eta)/p_a - \lambda_{b}^L(Z; \eta)/p_b - \left({\mu}(a, w^U) - {\mu}(b, w^L)\right)\right] + \xi_{a}^U(A, Z; \eta)/p_a - \xi_{b}^L(A, Z; \eta)/p_b\right\} \\
+& \sqrt{\npri}\expectnp\left\{r\left[\lambda_{a}^U(Z; \eta)/p_a - \lambda_{b}^L(Z; \eta)/p_b - \left({\mu}(a, w^U) - {\mu}(b, w^L)\right)\right] + \gamma_{a}^U(\hY, Z; \eta)/p_a - \gamma_{b}^L(\hY, Z; \eta)/p_b\right\} + o_p(1).
\end{align*}
The conclusion then follows directly from central limit theorem.
\endproof

\proof{Proof for \cref{corollary: cal-CI}.}
\cref{corollary: CI-dd} proves that $\hat{V}_L \overset{p}{\to} V_L$ and $\hat{V}_U \overset{p}{\to} V_U$, and we can analogously prove that $\hat{\op{CV}}_{LU} \overset{p}{\to} \op{CV}_{LU}$.
By \cref{corollary: asymp-dist-dd-extended} and  Slutsky's theorem,
\begin{align*}
\sqrt{n}_p\begin{bmatrix}
\hat{V}_L & \hat{\op{CV}}_{LU} \\
\hat{\op{CV}}_{LU} & \hat{V}_U 
\end{bmatrix}^{-1/2}\begin{bmatrix}\hat{\mu}(a, w^L) - \hat{\mu}(b, w^U) - \left({\mu}(a, w^L) - {\mu}(b, w^U)\right) \\ \hat{\mu}(a, w^U) - \hat{\mu}(b, w^L) - \left({\mu}(a, w^U) - {\mu}(b, w^L)\right) \end{bmatrix}\overset{d}{\to} \mathcal{N}
\left(
\begin{bmatrix}
0 \\ 0
\end{bmatrix},
\begin{bmatrix}
1 & 0 \\
0 & 1
\end{bmatrix}
\right).
\end{align*}
We introduce the following shorthand notations:
\[
 \begin{bmatrix}
\hat{x} - x\\
\hat{y} - y
\end{bmatrix} = \begin{bmatrix}\hat{\mu}(a, w^L) - \hat{\mu}(b, w^U) - \left({\mu}(a, w^L) - {\mu}(b, w^U)\right) \\ \hat{\mu}(a, w^U) - \hat{\mu}(b, w^L) - \left({\mu}(a, w^U) - {\mu}(b, w^L)\right) \end{bmatrix}.
\]
and
\begin{align*}
\begin{bmatrix}
 p & q \\
q & r
\end{bmatrix}
\coloneqq  
\begin{bmatrix}
\hat{V}_L & \hat{\op{CV}}_{LU} \\
\hat{\op{CV}}_{LU} & \hat{V}_U 
\end{bmatrix}^{-1/2}
= \frac{1}{s\sqrt{\hat{V}_L  + \hat{V}_U + 2{s}}}
\begin{bmatrix}
\hat{V}_U + {s} & -\hat{\op{CV}}_{LU} \\
-\hat{\op{CV}}_{LU} & \hat{V}_L + {s} 
\end{bmatrix}  
\end{align*}
where $s = \op{det}(\hat{V}^{1/2})$, and $p, r > 0$. Note that 
\begin{align*}
pr - q^2 = \frac{(\hat{V}_U + {s})(\hat{V}_L + {s}) - \hat{\op{CV}}^2_{LU}}{s^2({\hat{V}_L  + \hat{V}_U + 2{s}})} = \frac{2s^2 + s(\hat{V}_L  + \hat{V}_U )}{s^2({\hat{V}_L  + \hat{V}_U + 2{s}})} = \frac{1}{s} > 0.
\end{align*}

 \begin{align*}
\sqrt{n}_p
\begin{bmatrix}
 p & q \\
q & r
\end{bmatrix} 
 \begin{bmatrix}
\hat{x} - x\\
\hat{y} - y
\end{bmatrix}
=
\sqrt{n}_p\begin{bmatrix}
p(\hat{x} - x) + q(\hat{y} - y)\\
q(\hat{x} - x) + r(\hat{y} - y)
\end{bmatrix} \overset{d}{\to} \mathcal{N}
\left(
\begin{bmatrix}
0 \\ 0
\end{bmatrix},
\begin{bmatrix}
1 & 0 \\
0 & 1
\end{bmatrix}
\right), 
\end{align*}
Thus asymptotically $p(\hat{x} - x) + q(\hat{y} - y)$ and $q(\hat{x} - x) + r(\hat{y} - y)$ are both asymptotically normal and they are asymptotically independent. This means that for $t = \sqrt{1 - \beta}$,
\begin{align*}
&\pr\left(p(\hat{x} - x) + q(\hat{y} - y) \le \Phi^{-1}(t)/\npri^{1/2}, ~~ q(\hat{x} - x) + r(\hat{y} - y) \ge -\Phi^{-1}(t)/\npri^{1/2}\right) \to t^2 = 1 - \beta
\end{align*}
Note that 
\begin{align}
p(\hat{x} - x) + q(\hat{y} - y) \le \Phi^{-1}(t)/\npri^{1/2} \label{eq: CI-eq-1}\\
q(\hat{x} - x) + r(\hat{y} - y) \ge -\Phi^{-1}(t)/\npri^{1/2}. \label{eq: CI-eq-2}
\end{align}

\cref{eq: CI-eq-1,eq: CI-eq-2} imply that 
\begin{align*}
pq(\hat{x} - x) + {q^2}(\hat{y} - y) \le \frac{q\Phi^{-1}(t)}{\npri^{1/2}} \\
pq(\hat{x} - x) + pr(\hat{y} - y) \ge -\frac{p\Phi^{-1}(t)}{\npri^{1/2}} 
\end{align*}
which in turn implies that 
\begin{align*}
&(pr - q^2)(\hy - y) \ge - (q + p)\frac{\Phi^{-1}(t)}{\npri^{1/2}} \\
\implies & y \le \hy + \frac{q + p}{pr - q^2}\frac{\Phi^{-1}(t)}{\npri^{1/2}}  \\
\implies & {\mu}(a, w^U) - {\mu}(b, w^L) \le \hat{\mu}(a, w^U) - \hat{\mu}(b, w^L) + \frac{q + p}{pr - q^2}\frac{\Phi^{-1}(t)}{\npri^{1/2}}.
\end{align*}

\cref{eq: CI-eq-1,eq: CI-eq-2} also imply that
\begin{align*}
pr(\hat{x} - x) + {qr}(\hat{y} - y) \le \frac{r\Phi^{-1}(t)}{\npri^{1/2}} \\
q^2(\hat{x} - x) + qr(\hat{y} - y) \ge -\frac{q\Phi^{-1}(t)}{\npri^{1/2}} 
\end{align*}
which in turn implies that 
\begin{align*}
&(pr - q^2)(\hat{x} - x) \le (r + q)\frac{\Phi^{-1}(t)}{\npri^{1/2}} \\
\implies& x \ge \hat{x} - \frac{q + r}{pr - q^2}\frac{\Phi^{-1}(t)}{\npri^{1/2}}  \\
\implies& {\mu}(a, w^L) - {\mu}(b, w^U) \ge \hat{\mu}(a, w^L) - \hat{\mu}(b, w^U) - \frac{r + q}{pr - q^2}\frac{\Phi^{-1}(t)}{\npri^{1/2}}.
\end{align*}
Note that 
\begin{align*}
\frac{q + r}{pr - q^2} = \frac{s + \hat V_L - \hat{\op{CV}}_{LU}}{\sqrt{\hat V_L + \hat V_U + 2s}} , ~~~ \frac{q + p}{pr - q^2} = \frac{s + \hat V_U - \hat{\op{CV}}_{LU}}{\sqrt{\hat V_L + \hat V_U + 2s}}.
\end{align*}
Thus the following confidence interval achives $(1 - \beta)$ coverage asymptotically:
\begin{align*}
&\bigg[\hat{\mu}(a, w^L) - \hat{\mu}(b, w^U) - \frac{\op{det}(\hat V^{1/2}) + \hat V_L - \hat{\op{CV}}_{LU}}{\sqrt{\hat V_L + \hat V_U + 2\op{det}(\hat V^{1/2})}}\frac{\Phi^{-1}(t)}{\npri^{1/2}}, \\
&\qquad\qquad\qquad \hat{\mu}(a, w^U) - \hat{\mu}(b, w^L) + \frac{\op{det}(\hat V^{1/2}) + \hat V_U - \hat{\op{CV}}_{LU}}{\sqrt{\hat V_L + \hat V_U + 2\op{det}(\hat V^{1/2})}}\frac{\Phi^{-1}(t)}{\npri^{1/2}}\bigg].
\end{align*}
\endproof

\proof{Proof for \cref{corollary: asymp-dist-tprd-extended}}
According to the proof of \cref{thm: asymp-dist-tprd},
\begin{align*}
&\sqrt{\npri}\left[\hat{\mu}'_{\hy y}(a, \tilde{w}^L, \tilde{w}^U)  - \hat{\mu}'_{\hy y}(b, \tilde{w}^U, \tilde{w}^L) - \left({\mu}'_{\hy y}(a, \tilde{w}^L, \tilde{w}^U) - {\mu}'_{\hy y}(b, \tilde{w}^U, \tilde{w}^L\right)\right] \\
=&\sqrt{\npri}\expectnp\bigg\{r_n\left[\frac{{\mu}'_{1 - \hy, y}(a; \tw^U,\tw^L)}{{\overline{w}}^L_{a}(\hy, y)  + {\overline{w}}^U_{a}(1 - \hy, y)}\tilde{\lambda}^L_{a, \hy y}(Z; \tilde\eta) - \frac{{\mu}'_{\hy y}(a; \tw^L,\tw^U)}{{\overline{w}}^L_{a}(\hy, y)  + {\overline{w}}^U_{a}(1 - \hy, y)}\tilde{\lambda}^U_{a, 1 - \hy, y}(Z; \tilde\eta)\right]\\
-&\qquad\qquad r_n\left[\frac{{\mu}'_{1 - \hy, y}(b; \tw^L,\tw^U)}{{\overline{w}}^U_{b}(\hy, y)  + {\overline{w}}^L_{b}(1 - \hy, y)}\tilde{\lambda}^U_{b, \hy y}(Z; \tilde\eta) - \frac{{\mu}'_{\hy y}(b; \tw^U,\tw^L)}{{\overline{w}}^U_{b}(\hy, y)  + {\overline{w}}^L_{b}(1 - \hy, y)}\tilde{\lambda}^L_{b, 1 - \hy, y}(Z; \tilde\eta)\right]\bigg\} \\
+&\sqrt{\npri}\expectnp\bigg\{\frac{{\mu}'_{1 - \hy, y}(a; \tw^U,\tw^L)}{{\overline{w}}^L_{a}(\hy, y)  + {\overline{w}}^U_{a}(1 - \hy, y)}\tilde{\gamma}^L_{a, \hy y}(\hY, Y, Z; \tilde\eta) - \frac{{\mu}'_{\hy y}(a; \tw^L,\tw^U)}{{\overline{w}}^L_{a}(\hy, y)  + {\overline{w}}^U_{a}(1 - \hy, y)}\tilde{\gamma}^L_{a, 1 -\hy, y}(\hY, Y, Z; \tilde\eta)\\
-&\qquad\qquad \left[\frac{{\mu}'_{1 - \hy, y}(b; \tw^L,\tw^U)}{{\overline{w}}^U_{b}(\hy, y)  + {\overline{w}}^L_{b}(1 - \hy, y)}\tilde{\gamma}^U_{b, \hy y}(\hY, Y, Z; \tilde\eta) - \frac{{\mu}'_{\hy y}(b; \tw^U,\tw^L)}{{\overline{w}}^U_{b}(\hy, y)  + {\overline{w}}^L_{b}(1 - \hy, y)}\tilde{\gamma}^U_{b, 1 -\hy, y}(\hY, Y, Z; \tilde\eta)\right]\bigg\} \\
+&\sqrt{\npri}\expectna\bigg\{(1 - r_n)\left[\frac{{\mu}'_{1 - \hy, y}(a; \tw^U,\tw^L)}{{\overline{w}}^L_{a}(\hy, y)  + {\overline{w}}^U_{a}(1 - \hy, y)}\tilde{\lambda}^L_{a, \hy y}(Z; \tilde\eta) - \frac{{\mu}'_{\hy y}(a; \tw^L,\tw^U)}{{\overline{w}}^L_{a}(\hy, y)  + {\overline{w}}^U_{a}(1 - \hy, y)}\tilde{\lambda}^U_{a, 1 - \hy, y}(Z; \tilde\eta)\right] \\
-&\qquad\qquad (1 - r_n)\left[\frac{{\mu}'_{1 - \hy, y}(b; \tw^L,\tw^U)}{{\overline{w}}^U_{b}(\hy, y)  + {\overline{w}}^L_{b}(1 - \hy, y)}\tilde{\lambda}^U_{b, \hy y}(Z; \tilde\eta) - \frac{{\mu}'_{\hy y}(b; \tw^U,\tw^L)}{{\overline{w}}^U_{b}(\hy, y)  + {\overline{w}}^L_{b}(1 - \hy, y)}\tilde{\lambda}^L_{b, 1 - \hy, y}(Z; \tilde\eta)\right]\bigg\} \\
+& \sqrt{\npri}\expectna\bigg\{\frac{{\mu}'_{1 - \hy, y}(a; \tw^U,\tw^L)}{{\overline{w}}^L_{a}(\hy, y)  + {\overline{w}}^U_{a}(1 - \hy, y)}\tilde{\xi}^L_{a, \hy y}(A, Z; \tilde\eta) - \frac{{\mu}'_{\hy y}(a; \tw^L,\tw^U)}{{\overline{w}}^L_{a}(\hy, y)  + {\overline{w}}^U_{a}(1 - \hy, y)}\tilde{\xi}^L_{a, 1 -\hy, y}(A, Z; \tilde\eta) \\
-&\qquad\qquad \left[\frac{{\mu}'_{1 - \hy, y}(b; \tw^L,\tw^U)}{{\overline{w}}^U_{b}(\hy, y)  + {\overline{w}}^L_{b}(1 - \hy, y)}\tilde{\xi}^U_{b, \hy y}(A, Z; \tilde\eta) - \frac{{\mu}'_{\hy y}(b; \tw^U,\tw^L)}{{\overline{w}}^U_{b}(\hy, y)  + {\overline{w}}^L_{b}(1 - \hy, y)}\tilde{\xi}^U_{b, 1 -\hy, y}(A, Z; \tilde\eta)\right]\bigg\} + o_p(1),
\end{align*} 
and symmetrically the decomposition of $\sqrt{\npri}\left[\hat{\mu}'_{\hy y}(a, \tilde{w}^U, \tilde{w}^L)  - \hat{\mu}'_{\hy y}(b, \tilde{w}^L, \tilde{w}^U) - \left({\mu}'_{\hy y}(a, \tilde{w}^U, \tilde{w}^L) - {\mu}'_{\hy y}(b, \tilde{w}^L, \tilde{w}^U\right)\right]$ can be obtained by switching $L$ and $U$.

The conclusion then follows from central limit theorem.
\endproof

\proof{Proof for \cref{corollary: cal-CI-tprd}.}
By following the proof of \cref{corollary: CI-dd}, we can analogously prove that $\hat{\tilde{V}}_L \overset{p}{\to} \tilde{V}_L$, $\hat{\tilde{V}}_U \overset{p}{\to} \tilde{V}_U$, and $\hat{\tilde{\op{CV}}}_{LU}(\hy, y) \overset{p}{\to} {\tilde{\op{CV}}}_{LU}(\hy, y)$. The confidence interval result directly follows from the proof of \cref{corollary: cal-CI}.
\endproof

\section{Supplementary Information for BISG}\label{apx-casestudies}

\paragraph{Literature on BISG and Other Proxy Methods.}
\citet{fremont2005use} provide a comprehensive review on methods that use only geolocation or surname to impute unobserved race information and comment on their relative strengths for different groups in a US context. 
As surname and geolocation proxies complement each other, hybrid approaches like BISG were proposed to combine both \citep{elliott2008new, elliott2009using} and extended to further include first name \citep{voicu2018using}. 
In terms of the accuracy of race imputation,
BISG has been shown to outperform surname-only and geolocation-only analysis 
in many datasets, including medicare administration data \citep{dembosky2019indirect}, mortgage data \citep{CFPBproxy2014}, and voter registration records \citep{imai2016improving}. 

\paragraph{Background on BISG.}
The original BISG proxy method
\citep{CFPBproxy2014} uses an individual's surname $Z_s$ and residence geolocation $Z_g$ (census tract, ZIP code, county, etc.) as proxy variables, and estimates the conditional probability of race labels, $\pr(A=\alpha\mid Z_s,Z_g)$ from the auxiliary dataset (e.g. decennial census data).
Specifically, BISG uses a na\"ive Bayes classifier \citep[\S6.6.3]{friedman2001elements}: it assumes surname and geolocation are independent given race and uses Bayes's law to combine two separate estimates of the conditional probability of races labels given surname and geolocation, i.e., $\pr(A=\alpha\mid Z_s)$ and $\pr(A=\alpha\mid Z_g)$ respectively. 
$\pr(A=\alpha\mid Z_s)$ is typically estimated from a census surname list that includes the fraction of different races for surnames occurring at least 100 times \citep{censussurname}. And, $\pr(A=\alpha\mid Z_g)$ is typically estimated from census Summary File I \citep{us20102010}. See \cite{Baines} for more implicit assumptions in constructing the BISG proxy probabilities besides the naive bayes assumption.

\end{APPENDICES}

\end{document}